\documentclass{ecai}
\usepackage{times}
\usepackage{graphicx}
\usepackage{latexsym}

\usepackage{xcolor}
\usepackage{amssymb}
\usepackage{amsmath}
\usepackage{subcaption}
\usepackage{colortbl}
\usepackage{tikz}
\usepackage{url}
\usepackage[normalem]{ulem}
\usepackage[linesnumbered,algoruled,boxed,lined]{algorithm2e}

\newcommand*\circledl[1]{\tikz[baseline=(char.base)]{\node[shape=circle,draw,inner sep=1.5pt] (char) {#1};}}
\newcommand*\circledm[1]{\tikz[baseline=(char.base)]{\node[shape=circle,draw,inner sep=0.9pt] (char) {#1};}}
\newcommand*\circleds[1]{\tikz[baseline=(char.base)]{\node[shape=circle,draw,inner sep=0pt] (char) {#1};}}
\newcommand{\figref}[1]{Figure~\ref{fig:#1}}
\newcommand{\tabref}[1]{Table~\ref{tab:#1}}

\newcommand{\secref}[1]{Section~\ref{sec:#1}}

\newcommand\Tstrut{\rule{0pt}{2.4ex}}
\newcommand{\vect}{\textrm{vec}}

\newcommand{\add}[1]{#1}
\newcommand{\remove}[1]{}

\definecolor{rc}{rgb}{0.93,0.93,1}

\usepackage{pifont}
\usepackage[symbol*,bottom,flushmargin,hang,multiple]{footmisc}
\DefineFNsymbols*{smb}{{\ding{66}}{\ding{71}}{\ding{71}}{4}{5}{6}{7}{8}{9}{10}{11}{12}{13}}
\DefineFNsymbols*{numbers}{{1}{2}{3}{4}{5}{6}{7}{8}{9}{10}{11}{12}{13}}
\setfnsymbol{smb}

\begin{document}
	
\title{Classifying the classifier: dissecting the\\ weight space of neural networks}

\author{Gabriel Eilertsen\footnotemark[1]\,, \;
	Daniel J\"onsson\footnotemark[1]\,, \;
	Timo Ropinski\footnotemark[2]\,, \;
	Jonas Unger\footnotemark[1]\,, \!\and\!\!\!\!
	Anders Ynnerman\footnotemark[1]} 


\maketitle
\bibliographystyle{ecai}
\footnotetext[1]{Department of Science and Technology, Link\"oping University, Sweden}
\footnotetext[2]{Institute of Media Informatics, Ulm University, Germany}
\setfnsymbol{numbers}

\begin{abstract}
	This paper presents an empirical study on the weights of neural networks, where we interpret each model as a point in a high-dimensional space -- the \emph{neural weight space}. To explore the complex structure of this space, we sample from a diverse selection of training variations (dataset, optimization procedure, architecture, etc.) of neural network classifiers, and train a large number of models to represent the weight space. Then, we use a machine learning approach for analyzing and extracting information from this space. Most centrally, we train a number of novel \emph{deep meta-classifiers} with the objective of classifying different properties of the training setup by identifying their footprints in the weight space. Thus, the meta-classifiers probe for patterns induced by hyper-parameters, so that we can quantify how much, where, and when these are encoded through the optimization process. 
	\add{This provides a novel and complementary view for explainable AI, and}
	we show how meta-classifiers can reveal a great deal of information about the training setup and optimization, by only considering a small subset of randomly selected consecutive weights.
	\add{To promote further research on the weight space, we release the \emph{neural weight space} (NWS) dataset -- a collection of 320K weight snapshots from 16K individually trained deep neural networks.}
	
\end{abstract}

\section{Introduction}
The complex and non-linear nature of deep neural networks (DNNs) makes it difficult to understand how they operate, what features are used to form decisions, and how different selections of hyper-parameters influence the final optimized weights. This has led to the development of methods \add{in explainable AI (XAI)} for visualizing and understanding neural networks, and in particular for convolutional neural networks (CNNs). Thus, many methods are focused on the input image space, for example by deriving images that maximize class probability or individual neuron activations~\cite{Simonyan2013,Zeiler2014}. There are also methods which directly investigate neuron statistics of different layers~\cite{Liu2017}, or use layer activations for information retrieval~\cite{Li2015,Alain2016,Raghu2017}. 
However, these methods primarily focus on local properties, such as individual neurons and layers, and an in-depth analysis of the full set of model weights and the weight space statistics has largely been left unexplored. 


In this paper, we present a dissection and exploration of the \emph{neural weight space} (NWS) -- the space spanned by the weights of a large number of trained neural networks. 
We represent the space by training a total of 16K CNNs, where the training setup is randomly sampled from a diverse set of hyper-parameter combinations.
The performance of the trained models alone can give valuable information when related to the training setups, and suggest optimal combinations of hyper-parameters. However, given its complexity, it is difficult to directly reason about the sampled neural weight space, e.g. in terms of Euclidean or other distance measures -- there is a large number of symmetries in this space, and many possible permutations represent models from the same equivalence class. 
To address this challenge, we use a machine learning approach for discovering patterns in the weight space, by utilizing a set of meta-classifiers. These are trained with the objective of predicting the hyper-parameters used in optimization. Since each sample in the hyper-parameter space corresponds to points (models) in the weight space, the meta-classifiers seek to learn the inverse mapping from weights to hyper-parameters. This gives us a tool to directly reason about hyper-parameters in the weight space.
To enable comparison between heterogeneous architectures and to probe for local information, we introduce the concept of local meta-classifiers operating on only small subsets of the weights in a model. The accuracy of a local meta-classifier enables us to quantify where differences due to hyper-parameter selection are encoded within a model. 

We demonstrate how we can find detailed information on how optimization shapes the weights of neural networks.
For example, we can quantify how the particular dataset used for training influences the convolutional layers stronger than the deepest layers, and how initialization is the most distinguishing characteristic of the weight space. 
\add{We also see how weights closer to the input and output of a network faster diverge from the starting point as compared to the ``more hidden'' weights.}
Moreover, we can measure how properties in earlier layers, e.g. the filter size of convolutional layers, influence the weights in deeper layers. It is also possible to pinpoint how individual features of the weights influence a meta-classifier, e.g. how a majority of the differences imposed on the weight space by the optimizer are located in the bias weights of the convolutional layers. All such findings could aid in understanding DNNs and help future research on neural network optimization. Also, since we show that a large amount of information about the training setup could be revealed by meta-classifiers, the techniques are important in privacy related problem formulations, providing a tool for leaking information on a black-box model without any knowledge of its architecture.
\vspace{-0.2cm}\\
In summary, we present the following set of contributions:
\begin{itemize}
	\vspace{-0.1cm}
	\item We use the neural weight space as a general setting for exploring properties of neural networks, by representing it using a large number of trained CNNs.
	\item \add{We release the neural weight space (NWS) dataset, comprising 320K weight snapshots from 16K individually trained nets, together with scripts for training more samples and for training meta-classifiers\footnote{\url{https://github.com/gabrieleilertsen/nws}}.}
	\item We introduce the concept of neural network meta-classification for performing a dissection of the weight space, quantifying how much, where and when the training hyper-parameters are encoded in the trained weights.
	\item We demonstrate how a large amount of information about the training setup of a network can be revealed by only considering a small subset of consecutive weights, and how hyper-parameters can be practically compared across different architectures.
\end{itemize}

\add{We see our study as a first step towards understanding and visualizing neural networks from a direct inspection of the weight space. This opens up new possibilities in XAI, for understanding and explaining learning in neural networks in a way that has previously not been explored.}
Also, there are many other potential applications of the sampled weight space, such as learning-based initialization and regularization, learning measures for estimating distances between networks, learning to prevent privacy leakage of the training setup, learning model compression and pruning, learning-based hyper-parameter selection, and many more.

Throughout the paper we use the term \emph{weights} denoted $\theta$ to describe the trainable parameters of neural nets (including bias and batch normalization parameters), and the term \emph{hyper-parameters} denoted $\phi$ to describe the non-trainable parameters. For full generality we also include dataset choice and architectural specifications under the term hyper-parameter.

\section{Related work}\label{sec:background}

\noindent \textbf{Visualization and analysis:} 
A significant body of work has been directed towards \add{explainable AI (XAI) in deep learning, for} visualization and understanding of different aspects of neural networks~\cite{Hohman2018}, e.g. using methods for neural feature visualization. These aim at estimating the input images to CNNs that maximize the activation of certain channels or neurons~\cite{Simonyan2013,Zeiler2014,Mahendran2015,Yosinski2015,Selvaraju2017}, providing information on what are the salient features picked up by a CNN.
Another interesting viewpoint is how information is structured by neural networks. It is, for example, possible to define interpretability of the learned representations, with individual units corresponding to unique and interpretative concepts~\cite{Zhou2015,Bau2017,Zhou2018}. 
It has also been demonstrated how neural networks learn hierarchical representations~\cite{Bilal2018}. 

Many methods rely on comparing and embedding DNN layer activations. For example, activations generated by multiple images can be concatenated to represent a single trained model~\cite{Erhan2010}, and activations of a large number of images can be visualized using dimensionality reduction~\cite{Rauber2017}.
The activations can also be used to regress the training objective, measuring the level of abstraction and separation of different layers~\cite{Alain2016},
and for measuring similarity between different trainings and layers~\cite{Li2015,Raghu2017}, e.g. to show how different layers of CNNs converge.
In contrast to these methods we operate directly on the weights, and we explore a very large number of trainings from heterogeneous architectures. 

There are also previous methods which consider the model weights, e.g. in order to visualize the evolution of weights during training~\cite{Gallagher1997a,Gallagher1997b,Lipton2016,Lorch2016,Antognini2018,Gabella2019}. Another common objective is to monitor the statistics of the weights during training, e.g. using tools such as Tensorboard. While these consider one or few models, our goal is to compare a large number of different models and learn how optimization encodes the properties of the training setup within the model weights.



By training a very large number of fully connected networks, Novak et al. study how the sensitivity to model input correlates with generalization performance~\cite{Novak2018}. For different combinations of base-training and fine-tuning objectives, Zamir et al. quantify the transfer-learning capability between tasks~\cite{Zamir2018}. Yosinski et al. explore different configurations of pre-training and fine-tuning to evaluate and predict performance~\cite{Yosinski2014}. However, while monitoring the accuracy of many trainings is conceptually similar to our analysis in \secref{analysis}, our main focus is to search for information in the weights of all the trained models (\secref{dmc}). 




\vspace{5pt}
\noindent \textbf{Privacy and security:} 
Related to the meta-classification described in \secref{dmc} 
are the previous works aiming at detecting privacy leakage in machine learning models. Many such methods focus on member inference attacks, attempting to estimate if a sample was in the training data of a trained model, or generating plausible estimates of training samples~\cite{Fredrikson2015,Hitaj2017,Nasr2018}. Ateniese et al. train a set of models using different training data statistics~\cite{Ateniese2015}. The learned features are used to train a meta-classifier to classify the characteristics of the original training data. The meta-classifier is then used to infer properties of the non-disclosed training data of a targeted model. The method was tested on Hidden Markov Models and Support Vector Machines. A similar concept was used by Shokri et al. for training an attack model on output activations in order to determine whether a certain record is used as part of the model’s training dataset~\cite{Shokri2017}.
Our dissection using meta-classifiers can reveal information about training setup of an unknown DNN, provided that the trained weights are available; potentially with better accuracy than previous methods since we learn from a very large set of trained models. 






\vspace{5pt}
\noindent \textbf{Meta-learning:} 
There are many examples of methods for learning optimal architectures and hyper-parameters~\cite{Hutter2011,Mockus2012,Bergstra2012,Zoph2017,Real2017,Zoph2018,Liu2019}. However, these methods are mostly focused on evolutionary strategies to search for the optimal network design. In contrast, we are interested in comparing different hyper-parameters and architectural choices, not only to gain knowledge on advantageous setups, but primarily to explore how training setup affects the optimized weights of neural networks. To our knowledge there has not been any previous large-scale samplings of the weight space for the purpose of learning-based understanding of deep learning.


\section{Weight space representation}\label{sec:nws}
We consider all weights of a model as being represented by a single point in the high-dimensional neural weight space (NWS). To create such a representation from the mixture of convolutional filters, biases, and weight matrices of fully-connected (FC) layers, we use a vectorization operation. The vectorization is performed layer by layer, followed by concatenation. For the $i$th convolutional layer, all the filter weights $\mathbf{h}_{i,j}$ from filters $j = 1, ..., K$ are concatenated, followed by bias weights $\mathbf{b}_i$,
\begin{equation}
\theta_i= \theta_{i-1} \shortparallel \vect(\mathbf{h}_{i,1}) \shortparallel ... \shortparallel \vect(\mathbf{h}_{i,K}) \shortparallel \mathbf{b}_i,
\label{eqn:vec_conv}
\end{equation}
where $\vect(\cdot)$ denotes vectorization and $\mathbf{v}_1 \shortparallel \mathbf{v}_2$ concatenates vectors $\mathbf{v}_1$ and $\mathbf{v}_2$. 
For the FC layers, the weight matrices $\mathbf{W}_i$ are added using the same scheme,
\begin{equation}
\theta_i = \theta_{i-1} \shortparallel \vect(\mathbf{W}_i) \shortparallel \mathbf{b}_i. 
\end{equation}
\begin{table}
	\centering
	\setlength{\tabcolsep}{6pt}
	\def\arraystretch{1.3}
	\small
	\caption{Hyper-parameter collection used for random selection of training setup. The convolutional layer width specifies the number of channels of the last convolutional layer, while the FC layer width is the size of the first FC layer. All dataset samples are resized to 32$\times$32$\times$3 pixels. Note that we use the term hyper-parameters to also describe more fundamental choices, such as dataset and architecture. The circled options are used to train the fixed architecture models $C_{fixed}$ in \tabref{datasets}, where the remaining hyper-parameters are randomly selected.}
	\begin{tabular}{l|p{5.5cm}}
		\textbf{Parameter} & \textbf{Values}\\
		\hline
		\rowcolor{rc}
		\Tstrut Dataset & MNIST~\cite{Lecun1998}, CIFAR-10~\cite{Krizhevsky2009}, SVHN~\cite{Netzer2011}, STL-10~\cite{Coates2011}, Fashion-MNIST~\cite{Xiao2017}\\
		Learning rate & $0.0002-0.005$\\
		\rowcolor{rc}
		Batch size & $32, 64, 128, 256$\\
		Augmentation & Off, On\\
		\rowcolor{rc}
		Optimizer & ADAM~\cite{Kingma2014}, RMSProp~\cite{Hinton2012}, Momentum SGD\\
		Activation & ReLU~\cite{Nair2010}, ELU~\cite{Clevert2015}, Sigmoid, TanH\\
		\rowcolor{rc}
		Initialization & Constant, Random normal, Glorot uniform, Glorot normal~\cite{Glorot2010}\\
		\hline
		\hline
		Conv. filter size & $3, \circledl{5}, 7$ \Tstrut\\
		\rowcolor{rc}
		\# of conv. layers & $\circledl{3}, 4, 5$\\
		\# of FC layer & $\circledl{3}, 4, 5$\\
		\rowcolor{rc}
		Conv. layer width & $16, \circledm{32}, 48$\\
		FC layer width & $64, \circleds{128}, 192$\\
	\end{tabular}
	\label{tab:hyperparams}\\
\end{table}
Starting with the empty set $\theta_0 = \emptyset$, and repeating the vectorization operations for all $L$ layers, we arrive at the final weight vector $\theta = \theta_L$. For simplicity, we have not included indices over the 2D convolutional filters $\mathbf{h}$ and weight matrices $\mathbf{W}$ in the notation. Moreover, additional learnable weights, e.g. batch normalization parameters, can simply be added after the biases of each layer. Although the vectorization may rearrange the 2D spatial information e.g. in convolutional filters, there is still spatial structure in the 1D vector. 
We recognize that there is ample room for an improved NWS representation, e.g. accounting for weight space permutations and the 2D nature of filters. 
\remove{However, our main focus is to provide a simple representation,}
\add{However, we focus on starting the exploration of the weight space with a representation that is as simple as possible,}
and from the results in \secref{dmc} we will see that it is possible to extract a great deal of useful information from the vectorized weights.

\section{The NWS dataset}\label{sec:setup}
In this section, we describe the sampling of the NWS dataset. Then, we show how we can correlate training setup with model performance by regressing the test accuracy from hyper-parameters.

\subsection{Sampling}
To generate points in the NWS, we train CNNs by sampling a range of different hyper-parameters, as specified in \tabref{hyperparams}. For each training, the hyper-parameters are selected randomly, similarly to previous techniques for hyper-parameter search~\cite{Bergstra2012}. It is difficult to estimate the number of SGD steps needed for optimization, since this varies with most of the hyper-parameters. Instead, we rely on early stopping to define convergence. For each training we export weights at 20 uniformly sampled points along the optimization trajectory. In order to train and manage the large number of models, we use relatively small CNNs automatically generated based on the architectural hyper-parameters shown in \tabref{hyperparams}.
\begin{table}
	\centering
	\setlength{\tabcolsep}{5.2pt}
	\def\arraystretch{1.3}
	\small
	\caption{The \emph{neural weight space} (NWS) dataset used throughout the paper. We refer to the text and the supplementary material for a description of the capturing process.}
	\begin{tabular}{l|p{3.5cm}|l}
		\textbf{Name} & \textbf{Description} & \textbf{Quantity} \\
		\hline
		\rowcolor{rc}
		$C_{main}$ & Random hyper-parameters (including architecture) & 13K (10K/3K train/test) \Tstrut\\
		$C_{fixed}$ & Random hyper-parameters (fixed architecture) & 3K (2K/1K train/test) \\
	\end{tabular}
	\label{tab:datasets}\\
\end{table} 
The architectures are defined by 6-10 layers, and in total between $\sim$20K and $\sim$390K weights each. They follow a standard design, with a number of convolutional layers and 3 max-pooling layers, followed by a set of fully-connected (FC) layers. The loss is the same for all trainings, specified by cross-entropy. 
For regularization, all models use dropout \cite{Srivastava2014} with 50\% keep probability after each fully connected layer.
Moreover, we use batch normalization~\cite{Ioffe2015} for all trainings and layers. Without the normalization, many of the more difficult hyper-parameter settings do not converge (the supplementary material contains a discussion around this). 

We conduct 13K separate trainings with random hyper-parameters, 
and 3K trainings with fixed architecture for the purpose of global meta-classification (\secref{dmc_global}). \tabref{datasets} lists the resulting NWS datasets used throughout the paper. For detailed explanations of the training setup, and extensive training statistics (convergence, distribution of test accuracy, training time, model size, etc.), we refer to the supplementary material. 


\subsection{Regressing the test accuracy}\label{sec:analysis}

To gain an understanding on the influence of the different hyper-parameters in \tabref{hyperparams}, we regress the test accuracy of the sampled set of networks. Given the hyper-parameters $\phi$, we model the test accuracy with a linear relationship,
%
\begin{equation}
\label{eqn:linear_model}
\hat{a}(\phi) = \tau_0 + \sum_{o \in \Omega_O} \tau_o \phi_o + \sum_{c \in \Omega_C} \sum_{i=1}^{K_c} \tau_{c,i} [\phi_c = i],
\end{equation}
where $\tau$ are the model coefficients and $K_c$ is the number of categories for hyper-parameter $\phi_c$. $\Omega_O$ is the set of ordered hyper-parameters, and $\Omega_C$ is the set of categorical hyper-parameters. The categorical hyper-parameters are split into one binary for each category, as denoted by the Iverson bracket, $[\phi_c = i]$. We fit one linear model for each dataset, which means that we have in total 10 categorical and 1 ordered (learning rate) parameters for each model (see \tabref{hyperparams}). Although some of the categorical parameters actually are ordered (batch size, filter size, etc.), we split these to fit one descriptive correlation for each of the categories. In total we have 32 categorical, 1 ordered and 1 constant coefficient, so that the size of the set $\{\tau_0,\tau_o,\tau_{c,i}\}$ is 34.

\begin{figure}[t!]
	\centering
	\includegraphics[width=\linewidth, trim={0pt 0pt 0pt 0pt}, clip]{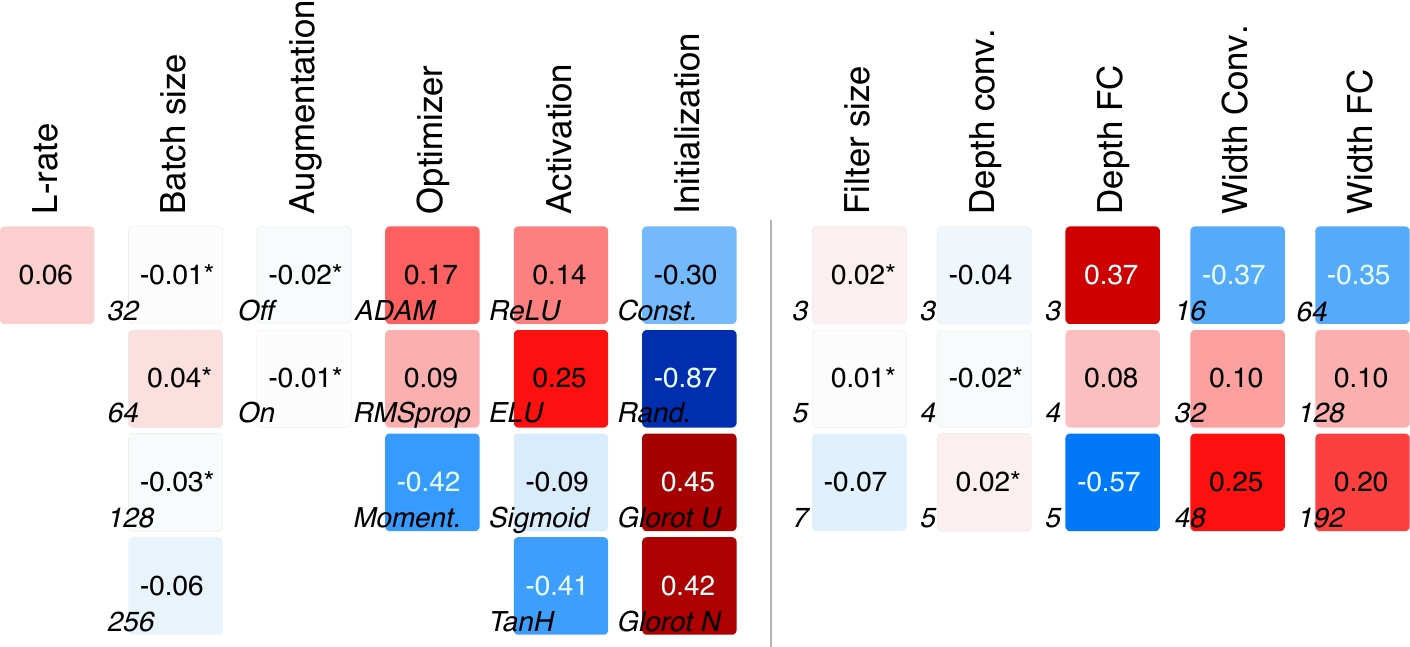}
	\caption{\label{fig:linear_regression} Regression of test accuracy from hyper-parameters on CIFAR-10. Columns represent different hyper-parameters, and the rows categories, as listed in \tabref{hyperparams}. * denotes a p-value that is larger than 5\%. For results from all datasets, we refer to the supplementary material.}
\end{figure}

The distribution of test accuracies on a particular dataset shows two modes; one with successful trainings and one with models that do not learn well. This makes it difficult to fit a linear model to the test accuracy. Instead, we focus only on the mode of models that have learned something useful, by rejecting all trainings with test accuracy lower than a certain threshold in-between the two modes. This sampling reduces the set $C_{main}$ from 13K to around 10.5K. 
Further, for each dataset, we normalize the test accuracy to have zero mean and unit variance. Thus, a positive model coefficient explains a positive effect on the test accuracy and vice versa, and the magnitudes are similar between different datasets. The results of fitting the model to one dataset, CIFAR-10, is displayed in \figref{linear_regression}. 

On average, the single most influential parameter is the initialization, followed by activation function and optimizer, and it is clear how large impact modern techniques have had on optimization (such as ADAM optimizer, ReLU/ELU activation and Glorot initialization).
For architecture specific parameters, a general trend is to promote wider models. However, the number of FC layers has an overall negative correlation. This can be explained by many layers being more difficult to optimize, so that performance suffers when less effective optimizer and initialization is used.
Finally, we recognize that a linear model is only explaining some of the correlations, but it helps in forming an overall understanding of the hyper-parameters.

\section{Meta-classification}\label{sec:dmc}
The objective of a meta-classifier is to learn the mapping $g: \theta \rightarrow \phi_c$, i.e. to estimate a specific hyper-parameter $\phi_c$ from a weight vector $\theta$. The performance of the $g(\theta)$ prediction of $\phi_c$ gives us a notion for comparing different $\theta$ in terms of hyper-parameters. We first give a motivation and definition, followed by examples of global and local meta-classification using meta-classifiers of different complexities.

\subsection{Motivation and definition}\label{sec:dmc_motivation}
For a model $f(\theta,x)$, trained using hyper-parameters $\phi$, what is specific about the learned weights $\theta$ comparing different $\phi$? 
For example, given two sets of weights $\Theta_a = \{\theta_{a,1},...,\theta_{a,N}\}$ and $\Theta_b = \{\theta_{b,1},...,\theta_{b,M}\}$, trained using different hyper-parameters $\phi_a$ and $\phi_b$, respectively, we expect the weights to converge to different locations in the weight space. However, it is difficult to relate to or reason about these locations based on direct inspection of the weights. For example, the Euclidean inter-distance $||\theta_{a,i}-\theta_{b,j}||$ may very well be smaller than the intra-distance $||\theta_{a,i}-\theta_{a,j}||$, due to the complicated and permutable structure of the weight space. In order to find the decision boundary between $\Theta_a$ and $\Theta_b$, we can instead learn it from a large number of samples $N$ and $M$, using a model $g(\theta)$. The model can thus be used to determine in which region (related to this decision boundary) a new weight sample $\theta$ is located. This gives us a notion of quantifying how much of $\phi_a$ or $\phi_b$ is encoded in $\theta$. 

Given a CNN classifier $f(\theta, x)$, parameterized by the trainable weights $\theta$, and operating on image samples $x$, a global meta-classifier is described by $g_c(\tau, \theta)$, where $\theta$ are static samples of the weight space and $\tau$ is the model parameterization. The objective of $g_c$ is to perform classification of the hyper-parameters $\phi_c$ as shown in \tabref{hyperparams}, to determine e.g. which dataset was used in training, if augmentation was performed, or which optimizer was used. $g_c$ takes the vectorized weights $\theta$ as input (see \secref{nws}). 

\vspace{5pt}
\noindent \textbf{Models:} We consider feature-based meta-classification, as well as deep models applied on the raw weight input. The features are specified from 8 different statistical measures of a weight vector: mean, variance, skewness (third standardized moment), and five-number summary (1, 25, 50, 75 and 99 percentiles). The measures are applied both directly on the weights $\theta$ and on weight gradients $\nabla \theta$, for a total of 16 features. The features are used for training support vector machines (SVMs), testing both linear and radial basis function (RBF) kernels. These models provide simple linear and non-linear baselines for comparison with a more advanced deep meta-classifier (DMC). We also tried performing logistic regression on the features, but performance was not better than random guessing.

A DMC is designed as a 1D CNN in order to handle the vectorized weights. Using convolutions on the vectorized weights can be motivated from three different perspectives: 1) there is spatial structure in the weight vector which we can explore, especially for the convolutional layers, 2) for local DMCs we are interested in spatial invariance, so that any subset of neighboring weights can be considered, and 3) for global DMCs we have a large input weight vector from which we need to extract a low-dimensional feature representation before using fully connected components.

\vspace{5pt}
\noindent \textbf{Data filtering:} 
Since the objective of a meta-classifier is to explore how hyper-parameters are encoded through the optimization process, we are only interested in models of the weight space that have learned something useful. Therefore, we discard trainings that have not converged, where convergence is defined as specified in \secref{analysis}.

\begin{figure}[t!]
	\centering
	\includegraphics[width=\linewidth, trim={2pt 3pt 2pt 1pt}, clip]{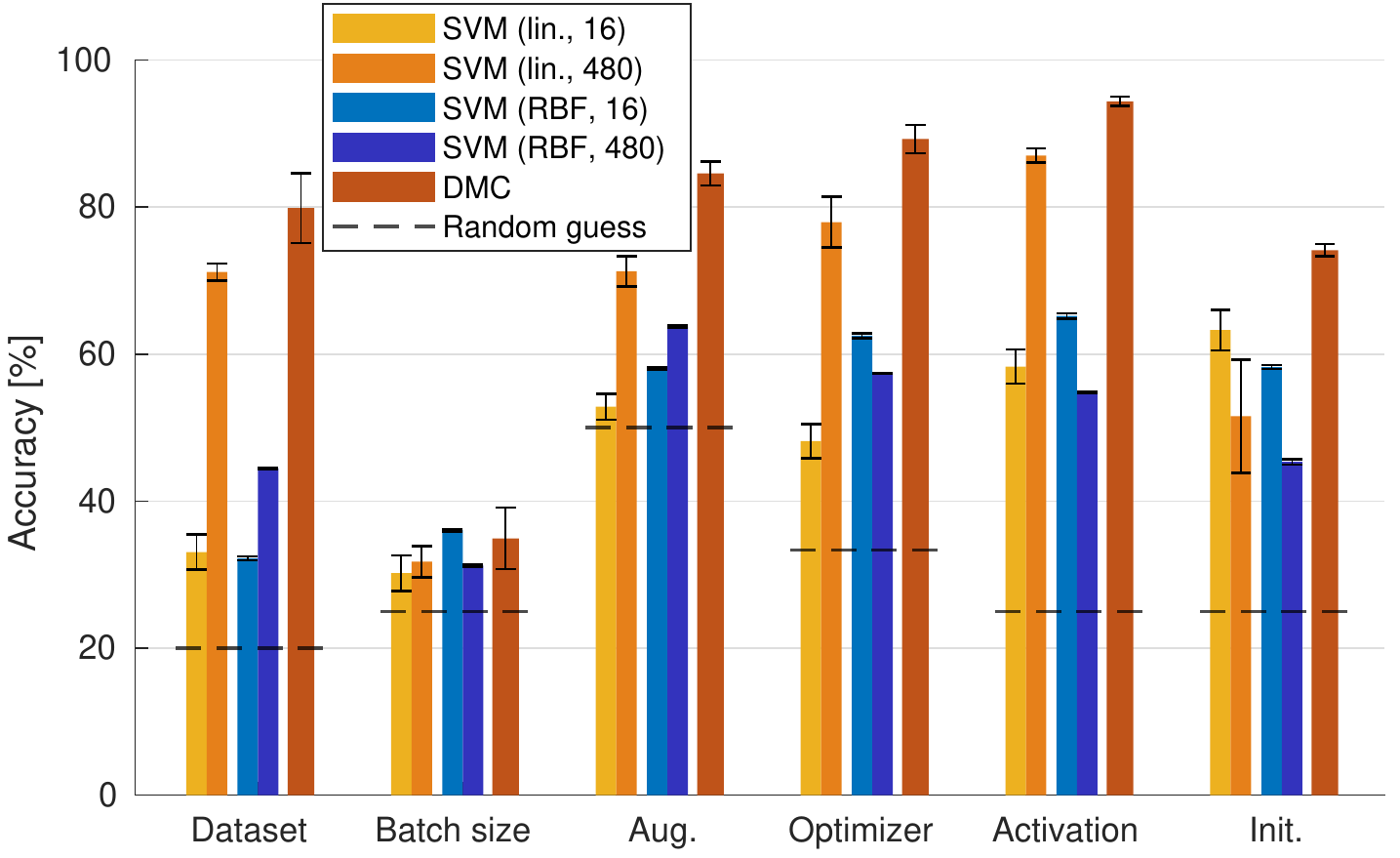}
	\caption{\label{fig:dmc_global} Global meta-classification performance for different hyper-parameters on the $C_{fixed}$ dataset (\tabref{hyperparams}). The numbers of the SVM classifiers specify the number of features used. Performance of random guessing is included for reference. The errorbars show standard deviations over 10 independent training rounds.}
\end{figure}

\begin{figure*}[t!]
	\centering
	\includegraphics[width=\textwidth, trim={0pt 0pt 0pt 0pt}, clip]{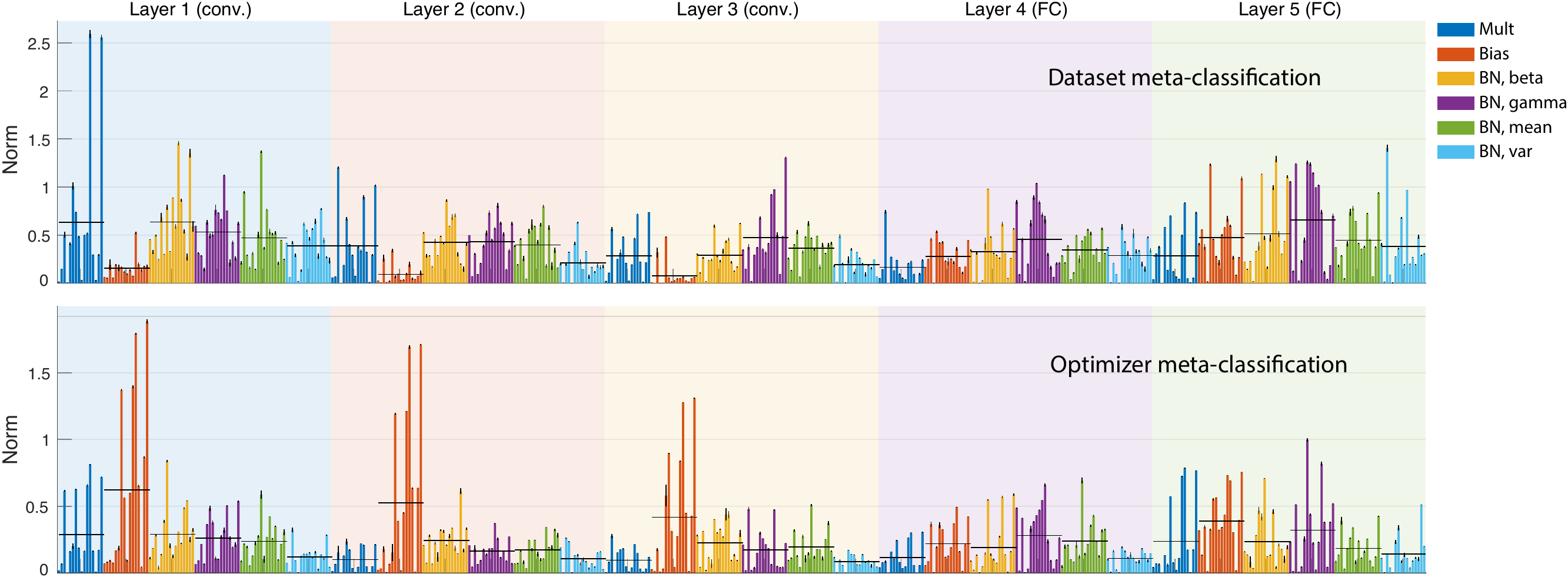}
	\caption{\label{fig:coeff_global_lwise} Estimated impact of individual features of linear SVMs for hyper-parameter classification on the set $C_{fixed}$ of the NWS dataset\add{, classifying dataset (top) and optimizer (bottom)}. Each of the different types of weights of a layer, on the x-axis, are described by 16 features. Horizontal bars denote mean over a type of weights. while error bars show standard deviations over 10 separate training runs. \emph{Mult} represent filters for convolutional layers and weight matrices for fully-connected (FC) layers. \emph{BN} denotes the parameters used for batch normalization.}
\end{figure*}

\subsection{Global meta-classification}\label{sec:dmc_global}
A global meta-classifier considers all trainable weights from each CNN. We train on the set $C_{fixed}$ in \tabref{datasets}, where each $\theta$ is composed of 92,868 weights. The DMC model consists of 15 1D convolutional layers followed by 6 FC layers. For the SVMs, we consider two methods for extracting the statistical measures mentioned in \secref{dmc_motivation} -- one is to evaluate statistics over the complete set of weights, and the other is to do this layer-by-layer. The layer-wise method extracts separate statistics for multiplicative weights, bias weights, and for each of the batch normalization weights in a layer, yielding a total of 480 training features. We refer to the supplementary material for details on the models and training.

\figref{dmc_global} shows the performance of global meta-classifiers trained on 6 different hyper-parameters. Considering the diversity of the training data, all of the hyper-parameters except for batch size can be predicted with fairly high accuracy using a DMC. This shows that there are many features in the weight space which are characteristic of different hyper-parameters. However, the most surprising results are achieved with a linear SVM and layer-wise weight statistics, with performance not very far from the deep classifiers, and especially for the dataset, optimizer and activation hyper-parameters. Apparently, using separate statistics for each layer and type of weight is enough to give a good estimation on which hyper-parameter was used in training.

By inspecting the decision boundaries of the linear SVMs, we can get a sense for which features that best explain a certain hyper-parameter, as illustrated in \figref{coeff_global_lwise}. 
Given the coefficients $\tau_c$ of a one-versus-rest SVM, it describes a vector that is orthogonal to the hyper-plane in feature space which separates the class $c$ from the rest of the classes. The vector is oriented along the feature axes which are most useful for separating the classes. Taking the norm of $\tau_c$ over the classes $c$, we can get an indication on which features were most important for separating the classes.
The most indicative features for determining the dataset are the 1 and 99 percentiles of the gradient of filter weights in the first convolutional layer. As these filters are responsible for extracting simple features, and the gradient of the filters are related to the strength of edge extraction, there is a close link to the statistics of the training images, which explains the good performance of the linear SVM. When evaluating the statistics over all weights, it is not possible to have this direct connection to training data, and performance suffers. For the optimizer meta-classifier, on the other hand, most information comes from the statistics of bias weights in the convolutional layers, and in particular from percentiles of $\theta$ and $\nabla \theta$. For activation function meta-classification, the running mean stored by batch normalization in the FC layers contribute most to the linear SVM decisions. For initialization it is more difficult to find isolated types of weights that contribute most, and looking at the SVMs trained on 16 features over the complete $\theta$ we can see how initialization is better described by statistics computed globally over the weights.

\begin{figure*}[t!]
	\centering
	\includegraphics[width=0.9\linewidth, trim={2pt 2pt 2pt 5pt}, clip]{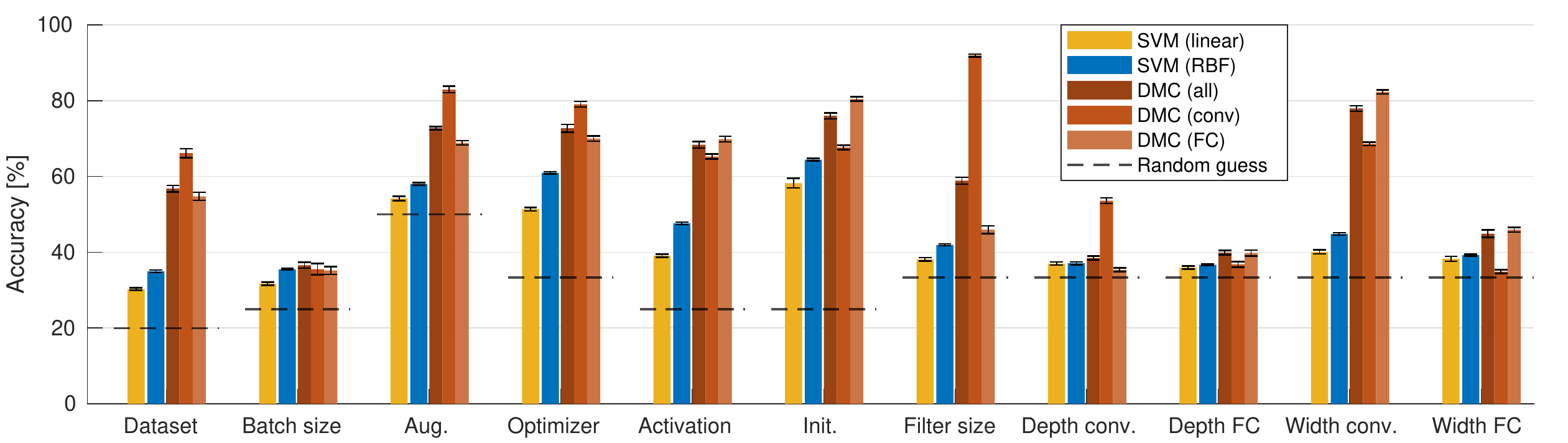}
	\caption{\label{fig:dmc_local} Local meta-classification performance for different hyper-parameters on the $C_{main}$ dataset (\tabref{datasets}), when using subsets of 5K weights. The SVM classifiers use 16 features from each weight subset. Local DMCs have been separately trained by considering subsets of all/convolutional/FC weights. Performance of random guessing is included for reference. The errorbars show standard deviations over 10 independent training rounds.}
\end{figure*}

\subsection{Local meta-classification}\label{sec:dmc_local}
A local meta-classifier is the model $g_c(\tau, \theta_{[a:b]})$, where $\theta_{[a:b]}$ is the subset of weights between indices $a$ and $b$. SVMs use features extracted from $\theta_{[a:b]}$, while a local DMC is trained directly on $\theta_{[a:b]}$. A local DMC consists of 12 1D convolutional layers followed by 6 FC layers. 
The training data is composed of the set $C_{main}$ in \tabref{datasets}. We use a subset size of $S=5000$ weights (on average around $5$\% of a weight vector), such that $b = a+S-1$. DMCs are trained by randomly picking $a$ for each mini-batch, while SVMs use a fixed number of 10 randomly selected subsets of each $\theta$.

\figref{dmc_local} shows the performance of local meta-classifiers $g_c$, trained on 11 different hyper-parameters $\phi_c$ from \tabref{hyperparams}. For each hyper-parameter, there are three individually trained DMCs based on: subsets of all weights in $\theta$, weights from only the convolutional layers, and only FC weights. 
In contrast to the global meta-classification it is not possible for an SVM to pinpoint statistics of one particular layer, which makes linear SVMs perform poorly. The RBF kernel improves performance, but is mostly far from the DMC accuracies. Still, the results are consistently better than random guessing for most hyper-parameters, so there is partial information contained in the statistical measures. The best SVM performance is achieved for the initialization hyper-parameter. This makes intuitive sense, as the differences are mainly described by simpler statistics of the weights. 

Considering that the local DMCs learn features that are invariant to the architecture, and use only a fraction of the weights, they perform very well compared to global DMCs. This is partly due to the larger training set, but also confirms how much information is stored locally in the vector $\theta$. By comparing DMCs trained on only convolutional or FC weights, we can analyze where most of the features of a certain hyper-parameter are stored, e.g. the dataset footprint is more pronounced in the convolutional layers. 
For the architectural hyper-parameters, the filter size can be predicted to some extent from only FC weights, which points towards how settings in the convolutional layers affect the FC weights. Compared to the global DMCs, initialization is a more profound local property as compared to e.g. dataset. 

Let $\theta_{a,j}$ denote the subset of $S$ weights starting at position $a$ from optimization step $j$. The trained model $g_c(\tau, \theta_{a,j})$ can then
probe for information across different depths of a model, and track how information evolve during training. Sampling at different $a$ and $j$, the result is a 2D performance map, see \figref{dmc}, where $(a,j) = (0,0)$ is in the upper left corner of the map. 
For the DMC trained to detect the optimizer used, the performance is approximately uniform across the weights except for a high peak close to the first layers. 
This roughly agrees with the feature importance of the linear SVM in \figref{coeff_global_lwise}, where information about optimizer can be encoded in the bias weights of the early convolutional layers. 
Inspecting how the DMC performance for initialization decreases faster in the first layers (\figref{dmc_initial}), we can see how learning faster diverges from the initialization point in the convolutional layers, which agrees with previous studies on how representations are learned \cite{Raghu2017}. However, we can also see the same tendency in the very last layers. That is, not only convolutional layers quickly diverge from the starting point to adapt to image content, but the last layers also do a similar thing when adapting to the output labels. However, looking at the minimum performance it is still easy to find patterns left from the initialization (see \figref{dmc_local}), and this hyper-parameter dominates the weight space locally. Connecting to the results in \secref{analysis}, initialization strategy was also the hyper-parameter of the NWS dataset that showed highest correlation with performance.


\section{Discussion}
\begin{figure*}[t!]
	\centering
	\begin{subfigure}{0.33\linewidth}
		\includegraphics[width=\linewidth, trim={2pt 2pt 8pt 2pt}, clip]{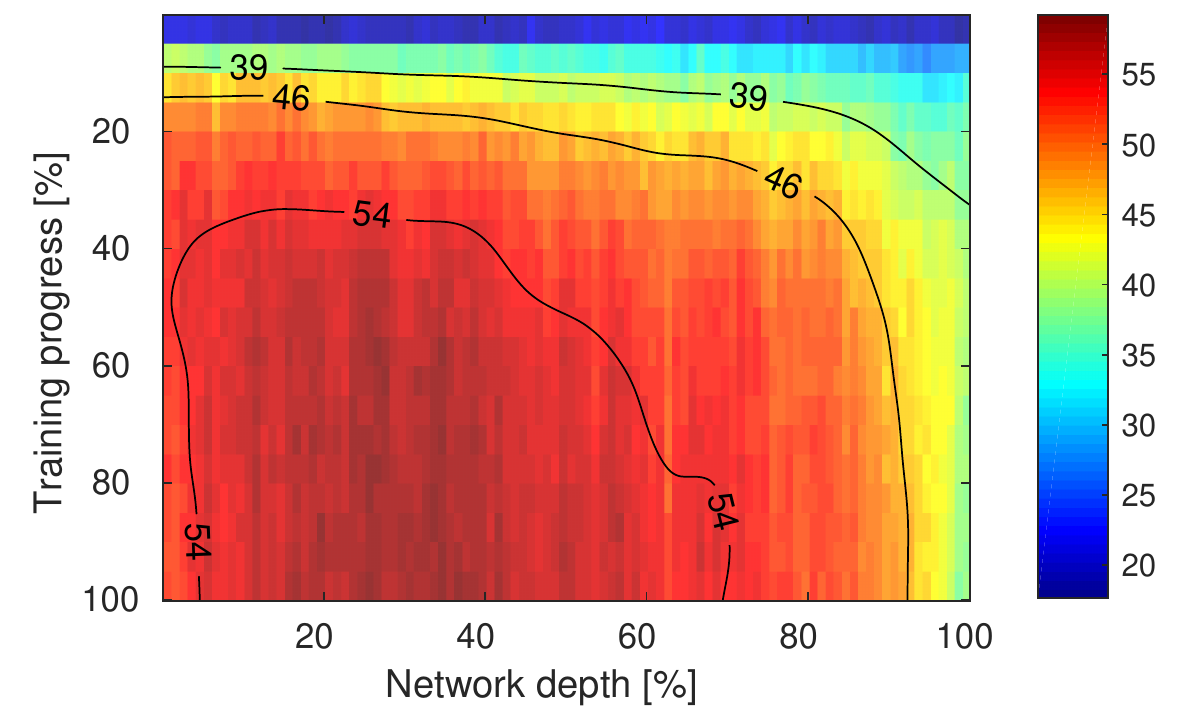}
		\caption{Dataset, DMC}
		\label{fig:dmc_ds}
	\end{subfigure}
	\begin{subfigure}{0.33\linewidth}
		\includegraphics[width=\linewidth, trim={2pt 2pt 8pt 2pt}, clip]{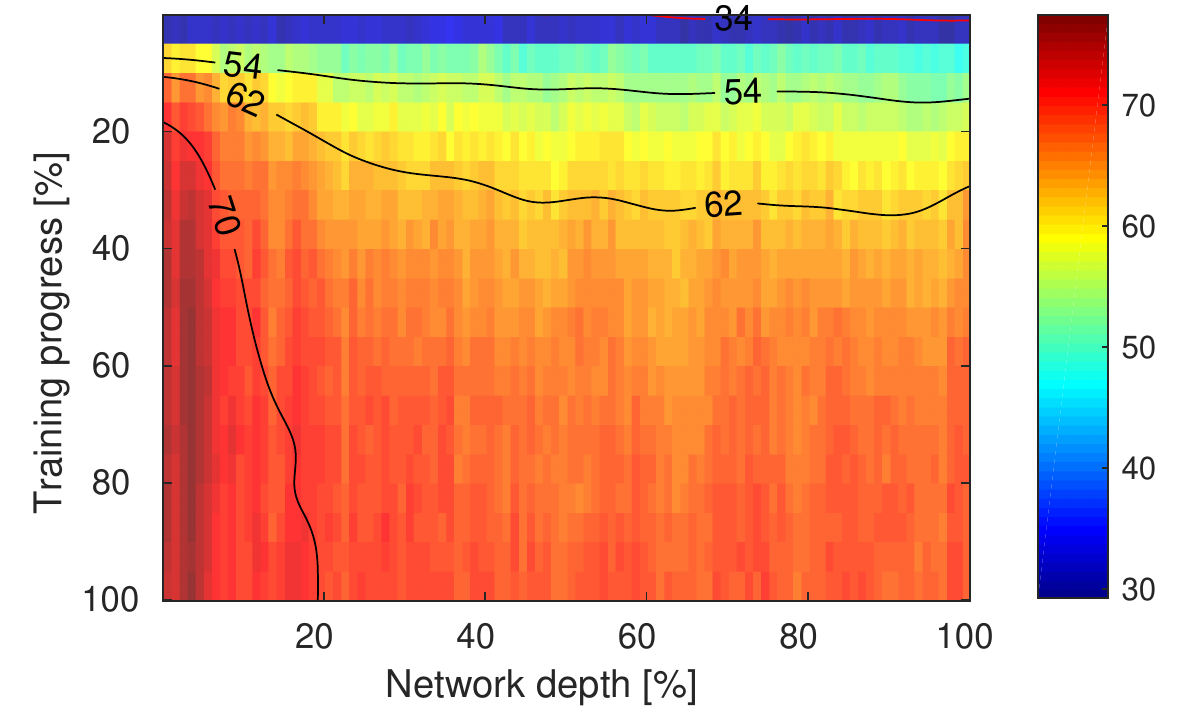}
		\caption{Optimizer, DMC}
		\label{fig:dmc_optim}
	\end{subfigure}
	\begin{subfigure}{0.33\linewidth}
		\includegraphics[width=\linewidth, trim={2pt 2pt 8pt 2pt}, clip]{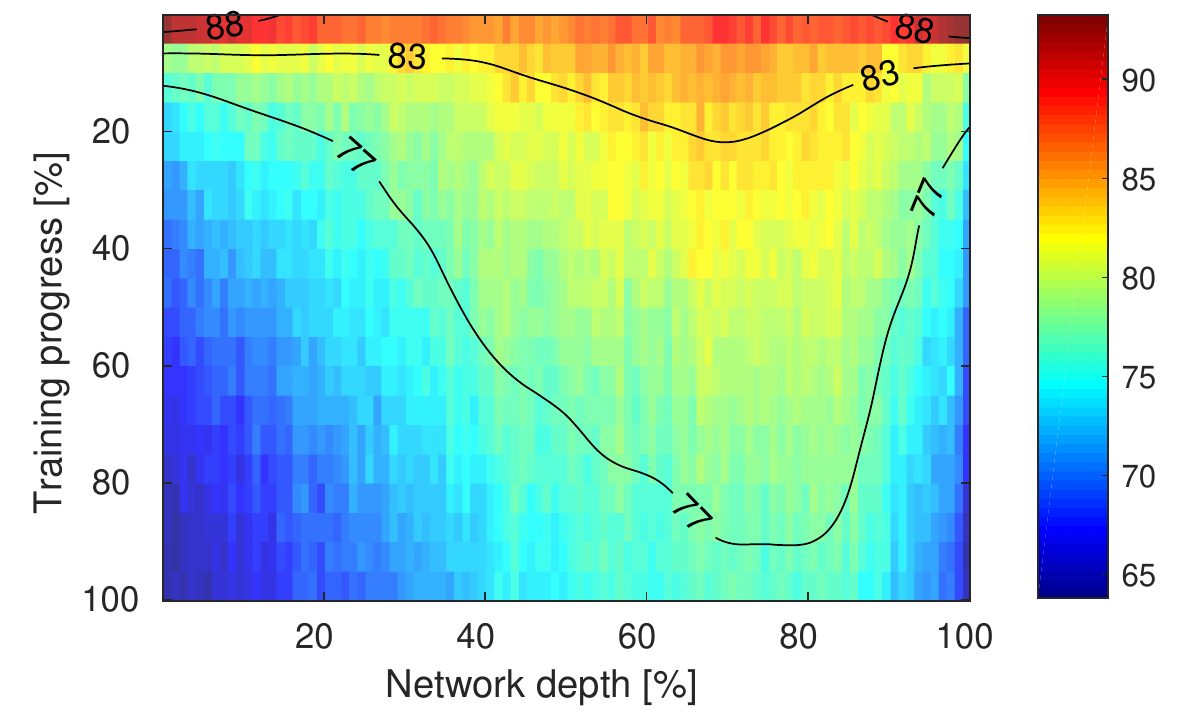}
		\caption{Initialization, DMC}
		\label{fig:dmc_initial}
	\end{subfigure}
	\vspace{0.1cm}\\
	\begin{subfigure}{0.33\linewidth}
		\includegraphics[width=\linewidth, trim={2pt 2pt 8pt 2pt}, clip]{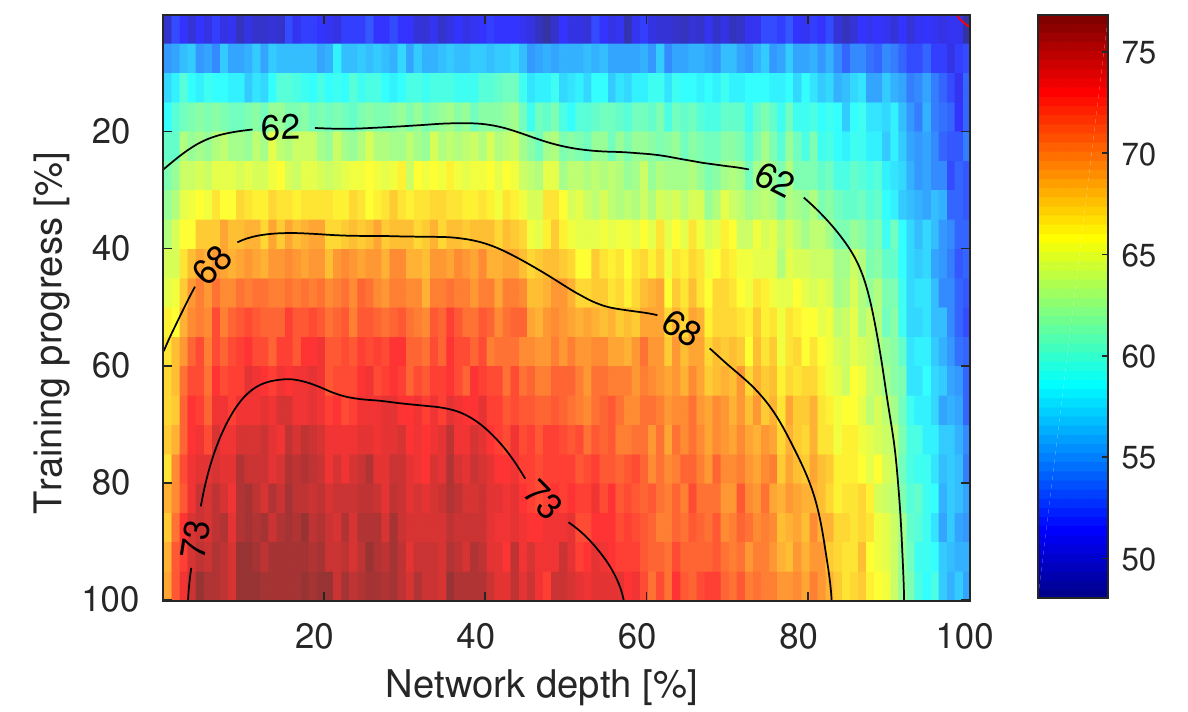}
		\caption{Augmentation, DMC}
		\label{fig:dmc_augment}
	\end{subfigure}
	\begin{subfigure}{0.33\linewidth}
		\includegraphics[width=\linewidth, trim={2pt 2pt 8pt 2pt}, clip]{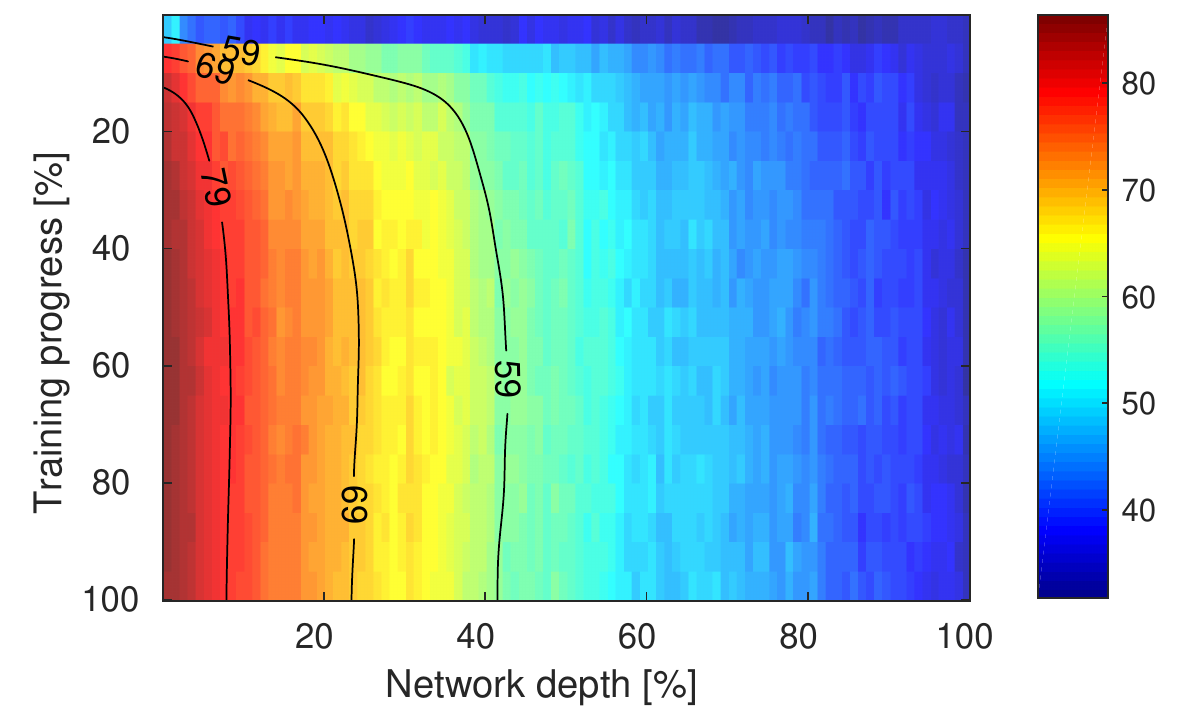}
		\caption{Filter size, DMC}
		\label{fig:dmc_filter}
	\end{subfigure}
	\begin{subfigure}{0.33\linewidth}
		\includegraphics[width=\linewidth, trim={2pt 2pt 8pt 2pt}, clip]{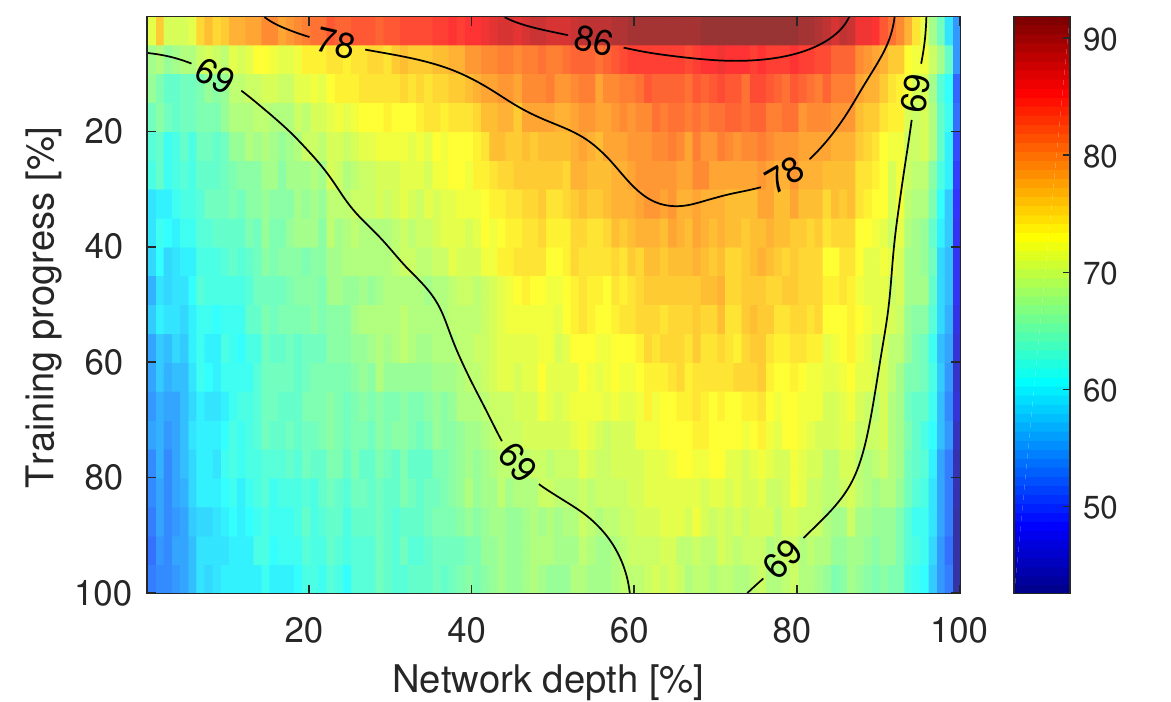}
		\caption{Initialization, SVM (RBF)}
		\label{fig:svm_init}
	\end{subfigure}
	\caption{\label{fig:dmc} Meta-classification performance maps, visualizing the average test accuracy at different model depths and training progress steps.
	\add{The y-axis illustrate training progress, from initialization (0\%) to converged model (100\%), while x-axis is the position of the weight vector where evaluation has been made, from the first weights of the convolutional layers (0\%) to the last weights of the fully connected layers (100\%).}
	Since the results are evaluated over models with different architectures, it is not possible to draw \add{separating} lines between individual layers. (c) and (f) show how similar trends are found by DMCs and SVMs. Note the different colormap ranges (e.g. higher lowest accuracy for initialization). For the results from all DMCs, \add{and individual evaluation on convolutional and fully connected layers}, we refer to the supplementary material.}
\end{figure*}

Looking at the results of the meta-classifications, SVMs can perform reasonably well when considering per-layer statistics on the whole set of weights. However, for extracting information from a random subset of weights, the DMCs are clearly superior, pointing to more complex patterns than the statistics used by the SVMs. It is interesting how much information a set of DMCs can extract from a very small subset of the weights, demonstrating how an abundance of information is locally encoded in the weight space. This is interesting from the viewpoint of privacy leakage, but the information can also be used for gaining valuable insight into how optimization shapes the weights of neural networks.
\add{This provides a new perspective for XAI, where we see our approach as a first step towards understanding neural networks from a direct inspection of the weight space.}
For example, in understanding and refining optimization algorithms, we may ask what are the differences in the learned weights caused by different optimizers? Or how does different activation functions affect the weights? Using a meta-classifier we can pinpoint how a majority of the differences caused by optimizer are due to different distributions in bias weights of the convolutional layers. The activation function used when training with batch normalization gives differences in the moving average used for batch normalization of the FC layers. Another interesting observation is the effect of the initialization in \figref{dmc_initial} and \ref{fig:svm_init}, which points to how convolutional and final layers diverge faster from the initialization point. A potential implication is to motivate studying what is referred to as \emph{differential learning rates} by the \emph{Fastai} library \cite{Howard2018}. This has been used for transfer learning, gradually increasing learning rate for deeper layers, but there could be reason to investigate the technique in a wider context, and to look at tuning learning rate of the last layers differently.

\subsection{Limitations and future work}
We have only scratched the surface of possible explorations in the neural weight space. There is a wealth of settings to be explored using meta-classification, where different combinations of hyper-parameters could reveal the structures of how DNNs learn. So far, we have only considered small vanilla CNNs. Larger and more diverse architectures and training regimes could be considered, e.g. ResNets~\cite{He2016}, GANs~\cite{Goodfellow2014}, RNNs, dilated and strided convolutions, as well as different input resolution, loss functions, and tasks.

The performance of DMCs can most likely be improved, e.g. by refining the representation of weights in \secref{nws}. And by aggregating information from the full weight vector using local DMCs, there is potential to learn many things about a black-box DNN with access only to the trained weights. 
Also, we consider only the weight space itself; a possible extension is to combine weights with layer activations for certain data samples.

\add{While XAI is one of the apparent applications of studying neural networks weights in closer detail, there are} many other important applications that would benefit from a large-scale analysis of the weight space, e.g. model compression, model privacy, model pruning, and distance metric learning. Another interesting direction would be to learn weight-generation, e.g. by means of GANs or VAEs, which could be used for initialization or ensemble learning.
The model in \secref{analysis} was used to show correlations between hyper-parameters and model performance. However, the topic of learning-based hyper-parameter optimization~\cite{Hutter2011,Mockus2012} could be explored in closer detail using the weight space sampling. Also, meta-learning could aim to include DMCs during training in order to steer optimization towards good regions in the NWS.


\section{Conclusions}\label{sec:conclusion}
This paper introduced the neural weight space (NWS) as a general setting for dissection of trained neural networks. We presented a dataset composed of 16K trained CNN classifiers, which we \add{make available for future research in explainable AI and meta-learning. The dataset was} studied both in terms of performance for different hyper-parameter selections, but most importantly we used meta-classifiers for reasoning about the weight space.
We showed how a significant amount of information on the training setup can be revealed by only considering a small fraction of random consecutive weights, pointing to the abundance of information locally encoded in DNN weights. We studied this information to learn properties of how optimization shapes the weights of neural networks. The results indicate how much, where, and when the optimization encodes information about the particular hyper-parameters in the weight space.

From the results, we pinpointed initialization as one of the most fundamental local features of the space, followed by activation function and optimizer. Although the actual dataset used for training a network also has a significant impact, the aforementioned properties are in general easier to distinguish, pointing to how optimization techniques can have a more profound effect on the weights as compared to the training data. 
We see many possible directions for future work, e.g. focusing on meta-learning for improving optimization in deep learning, for example using meta-classifiers during training in order to steer optimization towards good regions in the weight space.

\ack This project was supported by the Wallenberg Autonomous Systems and Software Program (WASP) and the
strategic research environment ELLIIT.

\bibliography{ecai}

\begin{thebibliography}{10}

\bibitem{Alain2016}
Guillaume Alain and Yoshua Bengio, `Understanding intermediate layers using
  linear classifier probes', {\em arXiv preprint arXiv:1610.01644}, (2016).

\bibitem{Antognini2018}
Joseph Antognini and Jascha Sohl-Dickstein, `{PCA} of high dimensional random
  walks with comparison to neural network training', in {\em Advances in Neural
  Information Processing Systems (NeurIPS 2018)}, (2018).

\bibitem{Ateniese2015}
Giuseppe Ateniese, Luigi~V Mancini, Angelo Spognardi, Antonio Villani, Domenico
  Vitali, and Giovanni Felici, `Hacking smart machines with smarter ones: How
  to extract meaningful data from machine learning classifiers', {\em
  International Journal of Security and Networks (IJSN)}, {\bf 10}(3), (2015).

\bibitem{Bau2017}
David Bau, Bolei Zhou, Aditya Khosla, Aude Oliva, and Antonio Torralba,
  `Network dissection: Quantifying interpretability of deep visual
  representations', in {\em IEEE Conference on Computer Vision and Pattern
  Recognition (CVPR 2017)}, (2017).

\bibitem{Bergstra2012}
James Bergstra and Yoshua Bengio, `Random search for hyper-parameter
  optimization', {\em Journal of Machine Learning Research (JMLR)}, {\bf 13},
  (2012).

\bibitem{Bilal2018}
Alsallakh Bilal, Amin Jourabloo, Mao Ye, Xiaoming Liu, and Liu Ren, `Do
  convolutional neural networks learn class hierarchy?', {\em IEEE transactions
  on visualization and computer graphics (TVCG)}, {\bf 24}(1), (2018).

\bibitem{Clevert2015}
Djork-Arn{\'e} Clevert, Thomas Unterthiner, and Sepp Hochreiter, `Fast and
  accurate deep network learning by exponential linear units (elus)', {\em
  arXiv preprint arXiv:1511.07289}, (2015).

\bibitem{Coates2011}
Adam Coates, Andrew Ng, and Honglak Lee, `An analysis of single-layer networks
  in unsupervised feature learning', in {\em International conference on
  artificial intelligence and statistics (AISTATS 2011)}, (2011).

\bibitem{Erhan2010}
Dumitru Erhan, Yoshua Bengio, Aaron Courville, Pierre-Antoine Manzagol, Pascal
  Vincent, and Samy Bengio, `Why does unsupervised pre-training help deep
  learning?', {\em Journal of Machine Learning Research (JMLR)}, {\bf 11},
  (2010).

\bibitem{Fredrikson2015}
Matt Fredrikson, Somesh Jha, and Thomas Ristenpart, `Model inversion attacks
  that exploit confidence information and basic countermeasures', in {\em ACM
  SIGSAC Conference on Computer and Communications Security (CCS 2015)},
  (2015).

\bibitem{Gabella2019}
Maxime Gabella, Nitya Afambo, Stefania Ebli, and Gard Spreemann, `Topology of
  learning in artificial neural networks', {\em arXiv preprint
  arXiv:1902.08160}, (2019).

\bibitem{Gallagher1997a}
Marcus Gallagher and Tom Downs, `Visualization of learning in neural networks
  using principal component analysis', in {\em International Conference on
  Computational Intelligence and Multimedia Applications (ICCIMA 1997)},
  (1997).

\bibitem{Gallagher1997b}
Marcus Gallagher and Tom Downs, `Weight space learning trajectory
  visualization', in {\em Australian Conference on Neural Networks (ACNN
  1997)}, (1997).

\bibitem{Glorot2010}
Xavier Glorot and Yoshua Bengio, `Understanding the difficulty of training deep
  feedforward neural networks', in {\em International conference on artificial
  intelligence and statistics (AISTATS 2010)}, (2010).

\bibitem{Goodfellow2014}
I.~Goodfellow, J.~Pouget-Abadie, M.~Mirza, B.~Xu, D.~Warde-Farley, S.~Ozair,
  A.~Courville, and Y.~Bengio, `Generative adversarial nets', in {\em
  International Conference on Neural Information Processing Systems (NIPS
  2014)}, (2014).

\bibitem{He2016}
Kaiming He, Xiangyu Zhang, Shaoqing Ren, and Jian Sun, `Deep residual learning
  for image recognition', in {\em IEEE conference on computer vision and
  pattern recognition (CVPR 2016)}, (2016).

\bibitem{Hinton2012}
Geoffrey Hinton, Nitish Srivastava, and Kevin Swersky, `Neural networks for
  machine learning lecture 6a overview of mini-batch gradient descent', (2012).

\bibitem{Hitaj2017}
Briland Hitaj, Giuseppe Ateniese, and Fernando P{\'e}rez-Cruz, `Deep models
  under the gan: information leakage from collaborative deep learning', in {\em
  ACM SIGSAC Conference on Computer and Communications Security (CCS 2017)},
  (2017).

\bibitem{Hohman2018}
Fred~Matthew Hohman, Minsuk Kahng, Robert Pienta, and Duen~Horng Chau, `Visual
  analytics in deep learning: An interrogative survey for the next frontiers',
  {\em IEEE transactions on visualization and computer graphics (TVCG)},
  (2018).

\bibitem{Howard2018}
Jeremy Howard et~al.
\newblock fastai.
\newblock \url{https://github.com/fastai/fastai}, 2018.

\bibitem{Hutter2011}
Frank Hutter, Holger~H Hoos, and Kevin Leyton-Brown, `Sequential model-based
  optimization for general algorithm configuration', in {\em International
  Conference on Learning and Intelligent Optimization (LION 2011)}, (2011).

\bibitem{Ioffe2015}
Sergey Ioffe and Christian Szegedy, `Batch normalization: Accelerating deep
  network training by reducing internal covariate shift', in {\em International
  Conference on Machine Learning (ICML 2015)}, (2015).

\bibitem{Kingma2014}
Diederik~P Kingma and Jimmy Ba, `{ADAM}: A method for stochastic optimization',
  {\em arXiv preprint arXiv:1412.6980}, (2014).

\bibitem{Krizhevsky2009}
Alex Krizhevsky and Geoffrey Hinton, `Learning multiple layers of features from
  tiny images', Technical report, Citeseer, (2009).

\bibitem{Krizhevsky2012}
Alex Krizhevsky, Ilya Sutskever, and Geoffrey~E Hinton, `Imagenet
  classification with deep convolutional neural networks', in {\em Advances in
  neural information processing systems (NIPS 2012)}, (2012).

\bibitem{Lecun1998}
Yann LeCun, L{\'e}on Bottou, Yoshua Bengio, Patrick Haffner, et~al.,
  `Gradient-based learning applied to document recognition', {\em Proceedings
  of the IEEE}, {\bf 86}(11), (1998).

\bibitem{Li2015}
Yixuan Li, Jason Yosinski, Jeff Clune, Hod Lipson, and John Hopcroft,
  `Convergent learning: Do different neural networks learn the same
  representations?', in {\em NIPS Workshop on Feature Extraction: Modern
  Questions and Challenges}, (2015).

\bibitem{Lipton2016}
Zachary~C Lipton, `Stuck in a what? adventures in weight space', {\em arXiv
  preprint arXiv:1602.07320}, (2016).

\bibitem{Liu2019}
Hanxiao Liu, Karen Simonyan, and Yiming Yang, `{DARTS}: Differentiable
  architecture search', in {\em International Conference on Learning
  Representations (ICLR 2019)}, (2019).

\bibitem{Liu2017}
Mengchen Liu, Jiaxin Shi, Zhen Li, Chongxuan Li, Jun Zhu, and Shixia Liu,
  `Towards better analysis of deep convolutional neural networks', {\em IEEE
  transactions on visualization and computer graphics (TVCG)}, {\bf 23}(1),
  (2017).

\bibitem{Lorch2016}
Eliana Lorch, `Visualizing deep network training trajectories with {PCA}', in
  {\em ICML Workshop on Visualization for Deep Learning}, (2016).

\bibitem{Mahendran2015}
Aravindh Mahendran and Andrea Vedaldi, `Understanding deep image
  representations by inverting them', in {\em IEEE conference on computer
  vision and pattern recognition (CVPR 2015)}, (2015).

\bibitem{Mcinnes2018}
Leland McInnes, John Healy, and James Melville, `Umap: Uniform manifold
  approximation and projection for dimension reduction', {\em arXiv preprint
  arXiv:1802.03426}, (2018).

\bibitem{Mockus2012}
Jonas Mockus, {\em Bayesian approach to global optimization: theory and
  applications}, volume~37, 2012.

\bibitem{Nair2010}
Vinod Nair and Geoffrey~E Hinton, `Rectified linear units improve restricted
  boltzmann machines', in {\em International conference on machine learning
  (ICML 2010)}, (2010).

\bibitem{Nasr2018}
Milad Nasr, Reza Shokri, and Amir Houmansadr, `Comprehensive privacy analysis
  of deep learning: Stand-alone and federated learning under passive and active
  white-box inference attacks', {\em arXiv preprint arXiv:1812.00910}, (2018).

\bibitem{Netzer2011}
Yuval Netzer, Tao Wang, Adam Coates, Alessandro Bissacco, Bo~Wu, and Andrew~Y
  Ng, `Reading digits in natural images with unsupervised feature learning', in
  {\em NIPS Workshop on Deep Learning and Unsupervised Feature Learning},
  (2011).

\bibitem{Novak2018}
Roman Novak, Yasaman Bahri, Daniel~A. Abolafia, Jeffrey Pennington, and Jascha
  Sohl-Dickstein, `Sensitivity and generalization in neural networks: an
  empirical study', in {\em International Conference on Learning
  Representations (ICLR 2018)}, (2018).

\bibitem{Raghu2017}
Maithra Raghu, Justin Gilmer, Jason Yosinski, and Jascha Sohl-Dickstein,
  `Svcca: Singular vector canonical correlation analysis for deep learning
  dynamics and interpretability', in {\em International Conference on Neural
  Information Processing Systems (NIPS 2017)}, (2017).

\bibitem{Rauber2017}
Paulo~E Rauber, Samuel~G Fadel, Alexandre~X Falcao, and Alexandru~C Telea,
  `Visualizing the hidden activity of artificial neural networks', {\em IEEE
  transactions on visualization and computer graphics (TVCG)}, {\bf 23}(1),
  (2017).

\bibitem{Real2017}
Esteban Real, Sherry Moore, Andrew Selle, Saurabh Saxena, Yutaka~Leon Suematsu,
  Jie Tan, Quoc~V Le, and Alexey Kurakin, `Large-scale evolution of image
  classifiers', in {\em International Conference on Machine Learning (ICML
  2017)}, (2017).

\bibitem{Selvaraju2017}
Ramprasaath~R Selvaraju, Michael Cogswell, Abhishek Das, Ramakrishna Vedantam,
  Devi Parikh, and Dhruv Batra, `Grad-cam: Visual explanations from deep
  networks via gradient-based localization', in {\em IEEE International
  Conference on Computer Vision (CVPR 2017)}, (2017).

\bibitem{Shokri2017}
Reza Shokri, Marco Stronati, Congzheng Song, and Vitaly Shmatikov, `Membership
  inference attacks against machine learning models', in {\em IEEE Symposium on
  Security and Privacy (SP)}. IEEE, (2017).

\bibitem{Simonyan2013}
Karen Simonyan, Andrea Vedaldi, and Andrew Zisserman, `Deep inside
  convolutional networks: Visualising image classification models and saliency
  maps', {\em arXiv preprint arXiv:1312.6034}, (2013).

\bibitem{Simonyan2014}
Karen Simonyan and Andrew Zisserman, `Very deep convolutional networks for
  large-scale image recognition', {\em arXiv preprint arXiv:1409.1556}, (2014).

\bibitem{Srivastava2014}
Nitish Srivastava, Geoffrey Hinton, Alex Krizhevsky, Ilya Sutskever, and Ruslan
  Salakhutdinov, `Dropout: a simple way to prevent neural networks from
  overfitting', {\em The Journal of Machine Learning Research (JMLR)}, {\bf
  15}(1), (2014).

\bibitem{Xiao2017}
Han Xiao, Kashif Rasul, and Roland Vollgraf, `Fashion-{MNIST}: a novel image
  dataset for benchmarking machine learning algorithms', {\em arXiv preprint
  arXiv:1708.07747}, (2017).

\bibitem{Yosinski2014}
Jason Yosinski, Jeff Clune, Yoshua Bengio, and Hod Lipson, `How transferable
  are features in deep neural networks?', in {\em International Conference on
  Neural Information Processing Systems (NIPS 2014)}, (2014).

\bibitem{Yosinski2015}
Jason Yosinski, Jeff Clune, Anh Nguyen, Thomas Fuchs, and Hod Lipson,
  `Understanding neural networks through deep visualization', in {\em ICML
  Workshop on Deep Learning}, (2015).

\bibitem{Zamir2018}
Amir~R Zamir, Alexander Sax, William Shen, Leonidas~J Guibas, Jitendra Malik,
  and Silvio Savarese, `Taskonomy: Disentangling task transfer learning', in
  {\em IEEE Conference on Computer Vision and Pattern Recognition (CVPR 2018)},
  (2018).

\bibitem{Zeiler2014}
Matthew~D Zeiler and Rob Fergus, `Visualizing and understanding convolutional
  networks', in {\em European conference on computer vision (ECCV 2014)}.
  Springer, (2014).

\bibitem{Zhou2018}
Bolei Zhou, David Bau, Aude Oliva, and Antonio Torralba, `Interpreting deep
  visual representations via network dissection', {\em IEEE transactions on
  pattern analysis and machine intelligence (TPAMI)}, (2018).

\bibitem{Zhou2015}
Bolei Zhou, Aditya Khosla, Agata Lapedriza, Aude Oliva, and Antonio Torralba,
  `Object detectors emerge in deep scene {CNNs}', in {\em International
  Conference on Learning Representations (ICLR 2015)}, (2015).

\bibitem{Zoph2017}
Barret Zoph and Quoc~V. Le, `Neural architecture search with reinforcement
  learning', in {\em International Conference on Learning Representations (ICLR
  2017)}, (2017).

\bibitem{Zoph2018}
Barret Zoph, Vijay Vasudevan, Jonathon Shlens, and Quoc~V Le, `Learning
  transferable architectures for scalable image recognition', in {\em IEEE
  conference on computer vision and pattern recognition (CVPR 2018)}, (2018).

\end{thebibliography}

\newpage
\newpage

\renewcommand\thesection{\Alph{section}}
\setcounter{section}{0}

\section*{\huge Supplementary material}
\vspace{0.2cm}

\section{NWS sampling}

\subsection{Hyper-parameters}
The different hyper-parameter choices are listed in Table 1 in the main paper. Here, we provide details on how the hyperparameters are specified. Note that, as in the paper, we use a broad definition of hyper-parameters which includes dataset and architectural design.

\paragraph{Dataset:} MNIST \cite{Lecun1998} uses 55K/10K train/test images at 28$\times$28 pixels resolution. These are up-sampled and replicated to be 32$\times$32$\times$3 pixels. CIFAR-10 \cite{Krizhevsky2009} uses 45K/10K train/test images at 32$\times$32$\times$3 pixels. SVHN \cite{Netzer2011} uses 73,257/26,032 train/test images at 32$\times$32$\times$3 pixels. STL-10 \cite{Coates2011} uses 5K/8K train/test images at 96$\times$96$\times$3 pixels resolution. These are down-sampled to 32$\times$32$\times$3 pixels. STL-10 also provides unlabeled images, but these are not utilized in our trainings. Fashion-MNIST \cite{Xiao2017} uses 55K/10K train/test images at 28$\times$28 pixels resolution. These are up-sampled and replicated to be 32$\times$32$\times$3 pixels.

All datasets have 10 classes each. All datasets except for SVHN have an equal number of images for each class. The class imbalance of SVHN is handled as described in \secref{opt}.

\paragraph{Architecture:} There are 5 hyper-parameters ($s, d_c, d_f, w_c, w_f$) for specifying the CNN architecture, which are used by Algorithm \ref{algo} to build the network. With the random selection of these hyper-parameters, the sampled weight space contains models with between 6 and 10 layers, and between $\sim$20K and $\sim$390K weights each. The distribution of model sizes is shown in \figref{dist_weights}.

\paragraph{Learning rate:} Randomly selected in the range $0.0002-0.005$, and then decayed by a factor $0.96$ in each epoch of training.

\paragraph{Augmentation:} Augmentation is performed online, i.e. by randomly transforming each of the training images for every mini-batch during training. The random transformations include: horizontal and vertical translations in the range $[-3,3]$ pixels, rotations in the range $[-15,15]$ degrees, zooming by a factor in the range $[-3.3,3.3]$ \%, shearing by a factor in the range $[-3,3]$ degrees, brightness adjustment by adding a constant in the range $[-0.2,0.2]$, contrast adjustment by the operation $k(x - \mu) + \mu$ where $\mu$ is the mean of image $x$ and k is picked in the range $[0.5,1.5]$, hue adjustment by an offset $[-0.08,0.08]$ to the H channel in the HSV color space, color saturation adjustment by multiplying the S channel in the HSV color space by a factor in the range $[0.3,1.5]$, and finally corruption by normally distributed noise with standard deviation in the range $[0,0.03]$.

\begin{algorithm}[t!]
	\small
	\SetAlgoLined
	\KwOut{CNN architecture}
	\KwIn{Filter size ($s$), depth Conv/FC ($d_c$/$d_f$), width Conv/FC ($w_c$/$w_f$)}
	CNN $=$ []\;
	CNN $+$$=$ conv2D($filter size = s\times s$, $channels = w_c$)\;
	CNN $+$$=$ max\_pool\_2x2()\;
	CNN $+$$=$ conv2D($filter size = s\times s$, $channels = 2\times w_c$)\;
	\If{$d_c > 4$}{
		CNN $+$$=$ conv2D($filter size = s\times s$, $channels = 2\times w_c$)\;
	}
	CNN $+$$=$ max\_pool\_2x2()\;
	CNN $+$$=$ conv2D($filter size = s\times s$, $channels = 4\times w_c$)\;
	\If{$d_c > 3$}{
		CNN $+$$=$ conv2D($filter size = s\times s$, $channels = 4\times w_c$)\;
	}
	CNN $+$$=$ max\_pool\_2x2()\;
	
	CNN $+$$=$ reshape()\;	
	\For{$c = 1,...,d_f$}{
		CNN $+$$=$ fc($width = 2^{-c+1} w_f$)\;
		CNN $+$$=$ dropout($0.5$)\;
	}
	CNN $+$$=$ fc($width = 20$)\;
	\caption{\label{algo}CNN construction from input filter size, depth, and width. Each conv. and FC layer is followed by batch normalization and activation function.}
\end{algorithm}

\paragraph{Optimizer:} The optimizers are used with default parameters in Tensorflow. For ADAM \cite{Kingma2014} this is $\beta_1=0.9$, $\beta_2=0.999$, and $\epsilon=10^{-8}$. For RMSprop \cite{Hinton2012} the decay is $0.9$, momentum is not used, and $\epsilon=10^{-10}$. For momentum SGD the momentum term is set to $0.95$.

\paragraph{Initialization:} The different initialization schemes are only applied to convolutional filters and FC weight matrices. The bias terms are always initialized as $0.0$. Constant initialization always uses the constant $0.1$, random normal initialization uses mean $0.0$ and standard deviation $1.0$, and the Glorot uniform and normal initialization schemes are parameter free \cite{Glorot2010}.

\subsection{Training procedure}\label{sec:opt}
For optimization, we use 10\% randomly selected training images as validation set. Since we train with a very diverse set of hyper-parameters it is difficult to specify for how many steps optimization should be performed. Thus, we use the validation set to provide early stopping criteria. The procedure is as follows: We evaluate the validation accuracy, $v_t$, after each epoch $t$ of training. We compute a filtered validation accuracy $\hat{v}_t = \alpha v_t + (1-\alpha) \hat{v}_{t-1}$ (we set $\alpha = 0.5$). Early stopping is performed if $t\ge20$ and one or more of the following criteria are full-filled:

\begin{enumerate}
	\item There are 5 consecutive decreasing validation accuracy evaluations $v_t < v_{t-1}$. This criteria is for detecting over-fitting.
	\item There are in total 30 decreasing validation accuracy evaluations, $v_t < \hat{v}_{t-1}$, when comparing to the smoothed validation accuracy. This criteria is used when there is noise in the validation accuracy evaluations across epochs, so that criteria 1 does not kick in, or if pronounced over-fitting does not occur.
	\item There are 30 consecutive stationary iterations $| v_t - v_{t-1} | < 10^{-8}$. This situation can occur if optimization gets stuck.
\end{enumerate}

\begin{figure}[t!]
	\centering
	\begin{subfigure}{0.4\textwidth}
		\includegraphics[width=\textwidth, trim={0pt 2pt 0pt 2pt}, clip]{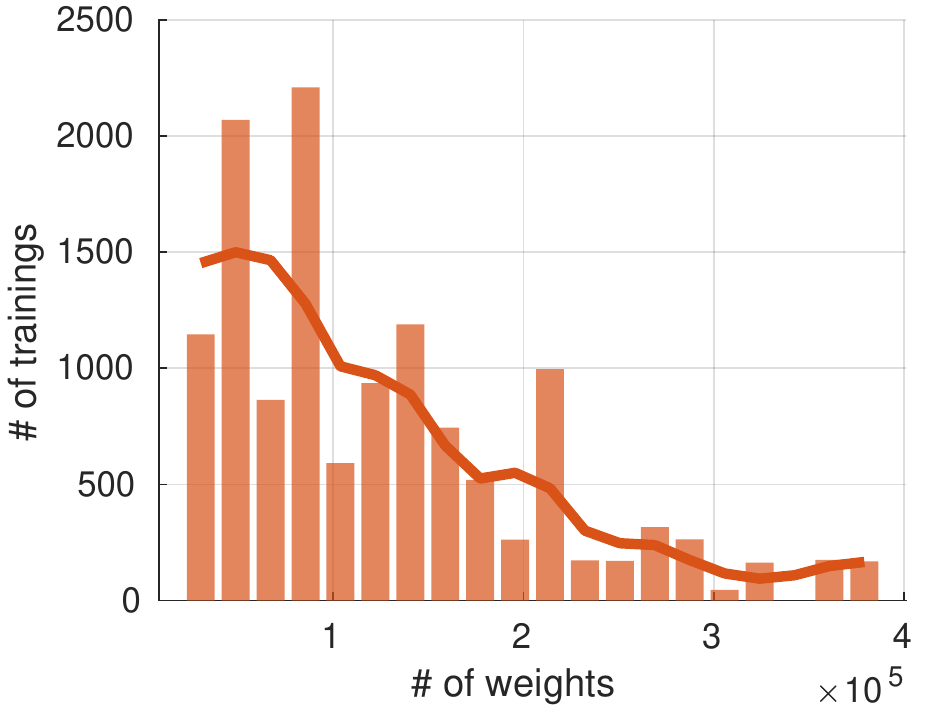}
		\caption{Model size (number of weights).}
		\label{fig:dist_weights}
	\end{subfigure}\\
	\vspace{0.2cm}
	\begin{subfigure}{0.4\textwidth}
		\includegraphics[width=\textwidth, trim={0pt 2pt 0pt 2pt}, clip]{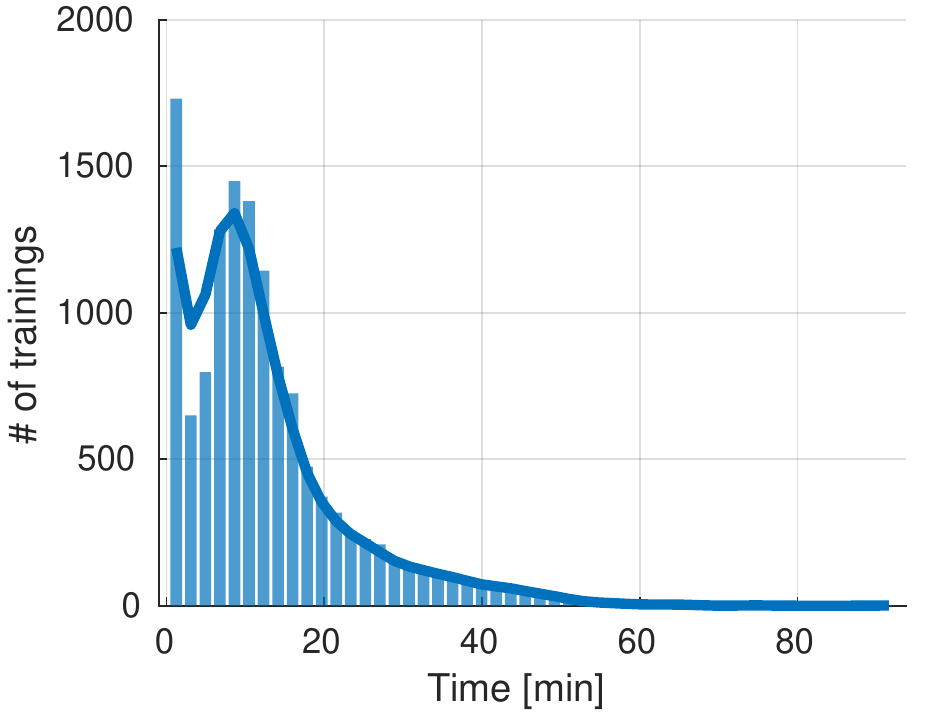}
		\caption{Time.}
		\label{fig:dist_time}
	\end{subfigure}
	\caption{\label{fig:dist} Distribution of model size and training time of the set $C_{main}$ of the sampled neural weight space (NWS) dataset.}
\end{figure}

Although these criteria capture many of the variations that can occur, there is some room for improvement. For example, training on STL-10, which is a smaller dataset than the others, there is more noise in the validation loss between epochs. This means that sometimes, mainly with difficult hyper-parameter setup, early stopping can kick earlier than optimal.

The wide diversity of hyper-parameters means that early stopping results in very different training lengths. The distribution of training times is shown in \figref{dist_time}.

Once per epoch, the current weights are exported. Since early stopping never applies before 20 epochs, we always have exported weights at 20 points along the optimization trajectory. However, in most cases there are a lot more. To be consistent and to be able to manage the amount of data, we always only keep 20 weight snapshots. This means that if training is longer than 20 epochs, we keep only the exported weights at 20 uniformly sampled epochs from initialization to convergence.

The SVHN dataset has an unbalanced number of images. We address this problem by randomly selecting $K$ images from each class, where $K$ is the minimum number of images of a class. In order to utilize all images, this random selection is repeated each epoch.

\subsection{Training statistics}
\begin{figure*}[t!]
	\centering
	\begin{subfigure}{0.325\textwidth}
		\includegraphics[width=\textwidth, trim={0pt 2pt 0pt 2pt}, clip]{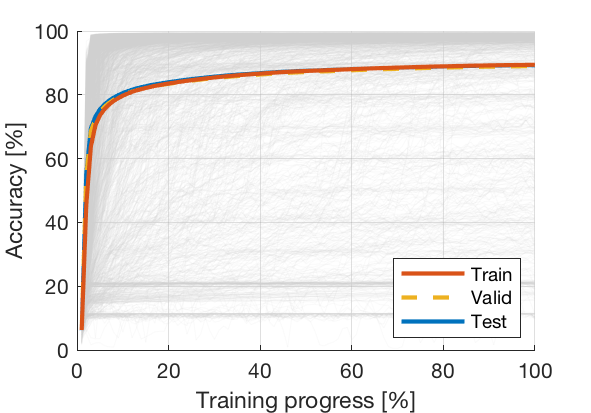}
		\caption{MNIST (2529 trainings)}
		\label{fig:progress_mnist}
	\end{subfigure}
	\begin{subfigure}{0.325\textwidth}
		\includegraphics[width=\textwidth, trim={0pt 2pt 0pt 2pt}, clip]{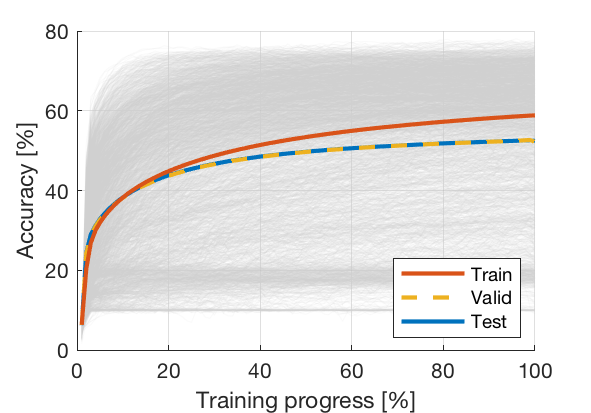}
		\caption{CIFAR-10 (2650 trainings)}
		\label{fig:progress_cifar}
	\end{subfigure}
	\begin{subfigure}{0.325\textwidth}
		\includegraphics[width=\textwidth, trim={0pt 2pt 0pt 2pt}, clip]{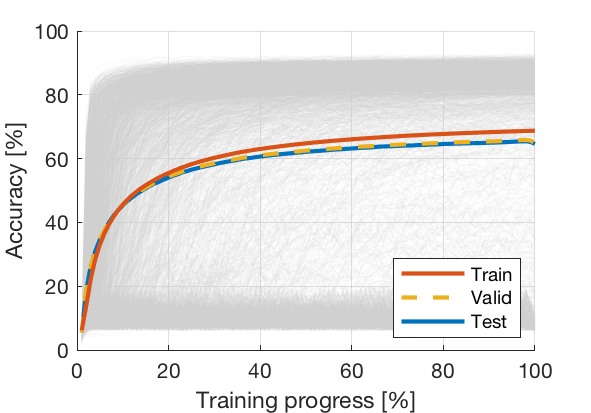}
		\caption{SVHN (2561 trainings)}
		\label{fig:progress_svhn}
	\end{subfigure}\\
	\vspace{0.2cm}
	\begin{subfigure}{0.325\textwidth}
		\includegraphics[width=\textwidth, trim={0pt 2pt 0pt 2pt}, clip]{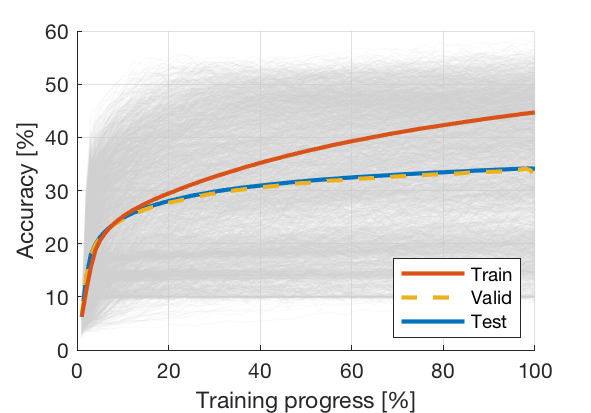}
		\caption{STL-10 (2588 trainings)}
		\label{fig:progress_stl}
	\end{subfigure}
	\begin{subfigure}{0.325\textwidth}
		\includegraphics[width=\textwidth, trim={0pt 2pt 0pt 2pt}, clip]{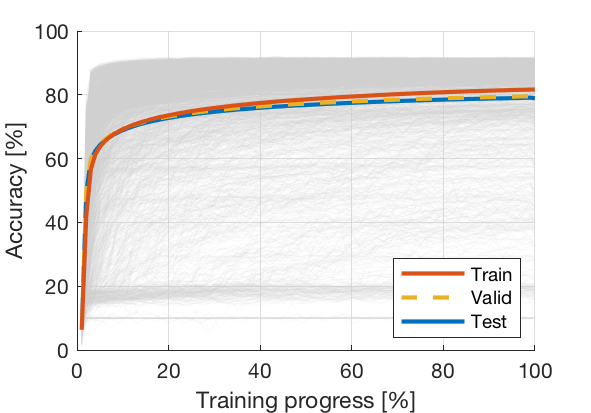}
		\caption{Fashion-MNIST (2672 trainings)}
		\label{fig:progress_fashion}
	\end{subfigure}
	\hspace{0.325\textwidth}
	\caption{\label{fig:progress} Training progress for each dataset. The gray curves are the individual test accuracies during training, while the red/yellow/blue curves show the mean train/valid/test accuracies. Note that the x-axes show the progress in percentage, which means that there is no differentiation of the time it takes to train.}
	\vspace{0.2cm}
\end{figure*}

\figref{progress} shows the training, validation and test accuracy during the progress of optimization, for all 13K trainings in the NWS set $C_{main}$. The figure is separated between the 5 datasets, and also provides mean accuracies over all the trainings. For all datasets there is a wide spread in end accuracy and convergence behavior, which is tightly linked to the hyper-parameter setup. Also, there is a distinct difference between datasets when it comes to over-fitting. For example, inspecting the difference between training and validation accuracy, over-fitting hardly occurs when training on MNIST, while CIFAR-10 and STL-10 are more susceptible to this phenomena.

\begin{figure*}[t!]
	\centering
	\begin{subfigure}{0.325\textwidth}
		\includegraphics[width=\textwidth, trim={0pt 2pt 0pt 2pt}, clip]{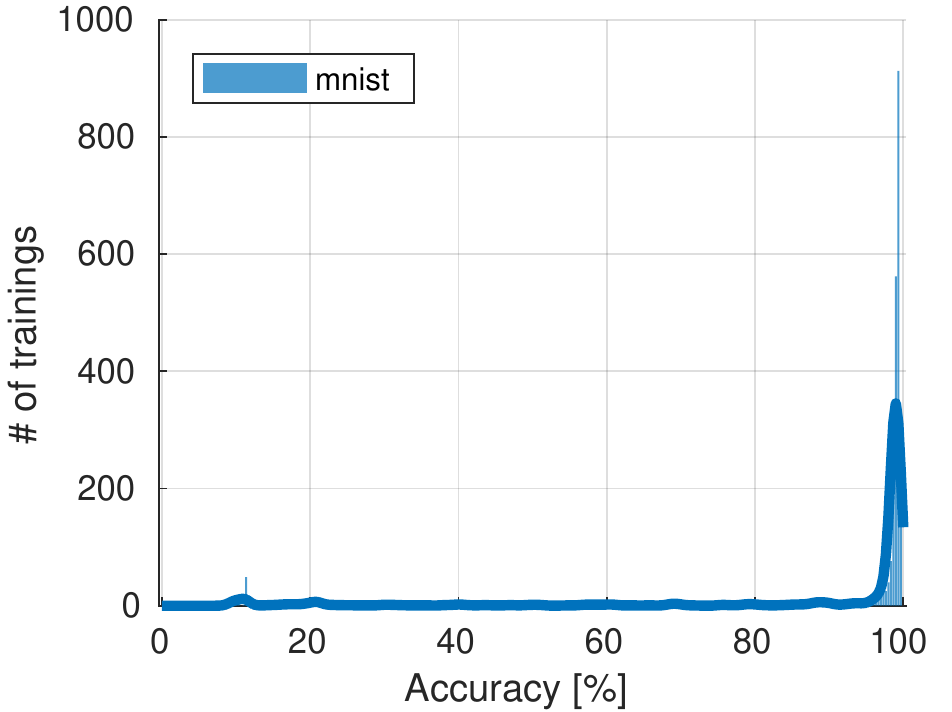}
		\caption{MNIST (2529 trainings)}
		\label{fig:accuracy_mnist}
	\end{subfigure}
	\begin{subfigure}{0.325\textwidth}
		\includegraphics[width=\textwidth, trim={0pt 2pt 0pt 2pt}, clip]{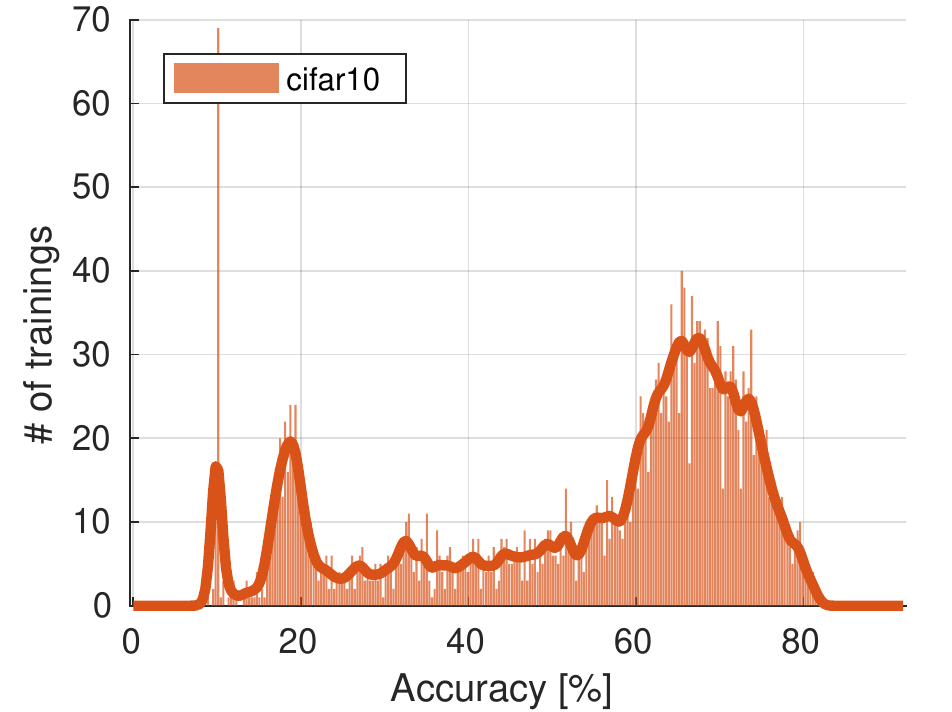}
		\caption{CIFAR-10 (2650 trainings)}
		\label{fig:accuracy_cifar}
	\end{subfigure}
	\begin{subfigure}{0.325\textwidth}
		\includegraphics[width=\textwidth, trim={0pt 2pt 0pt 2pt}, clip]{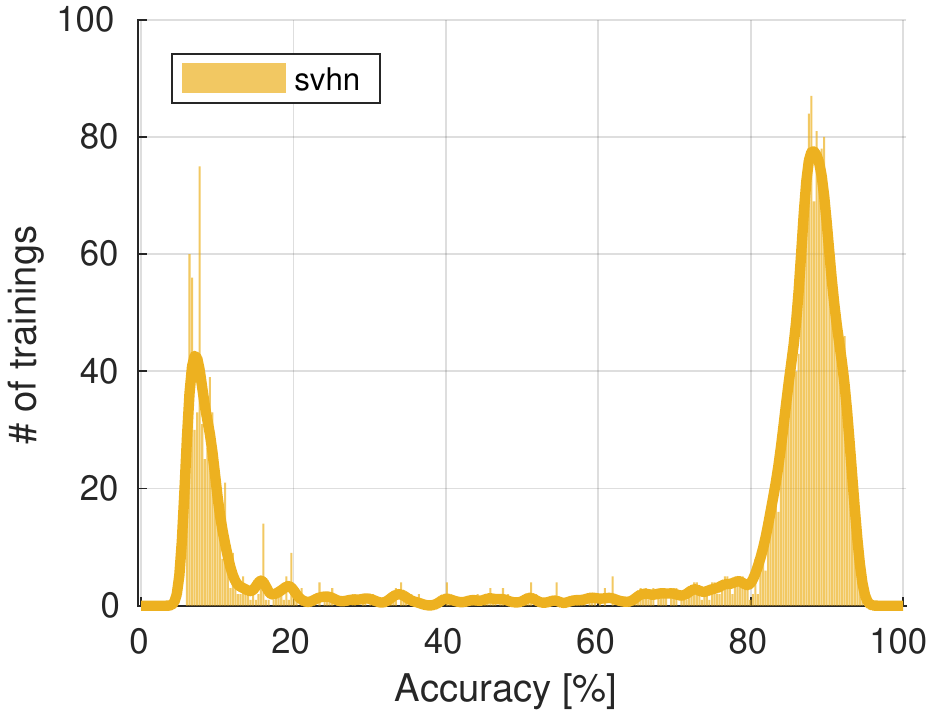}
		\caption{SVHN (2561 trainings)}
		\label{fig:accuracy_svhn}
	\end{subfigure}\\
	\vspace{0.2cm}
	\begin{subfigure}{0.325\textwidth}
		\includegraphics[width=\textwidth, trim={0pt 2pt 0pt 2pt}, clip]{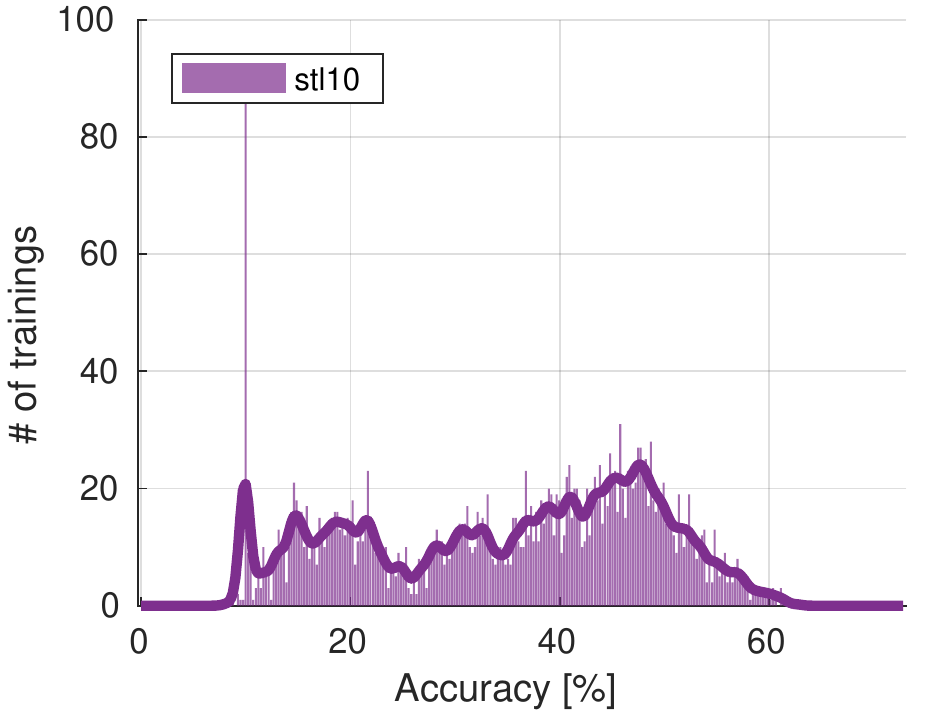}
		\caption{STL-10 (2588 trainings)}
		\label{fig:accuracy_stl}
	\end{subfigure}
	\begin{subfigure}{0.325\textwidth}
		\includegraphics[width=\textwidth, trim={0pt 2pt 0pt 2pt}, clip]{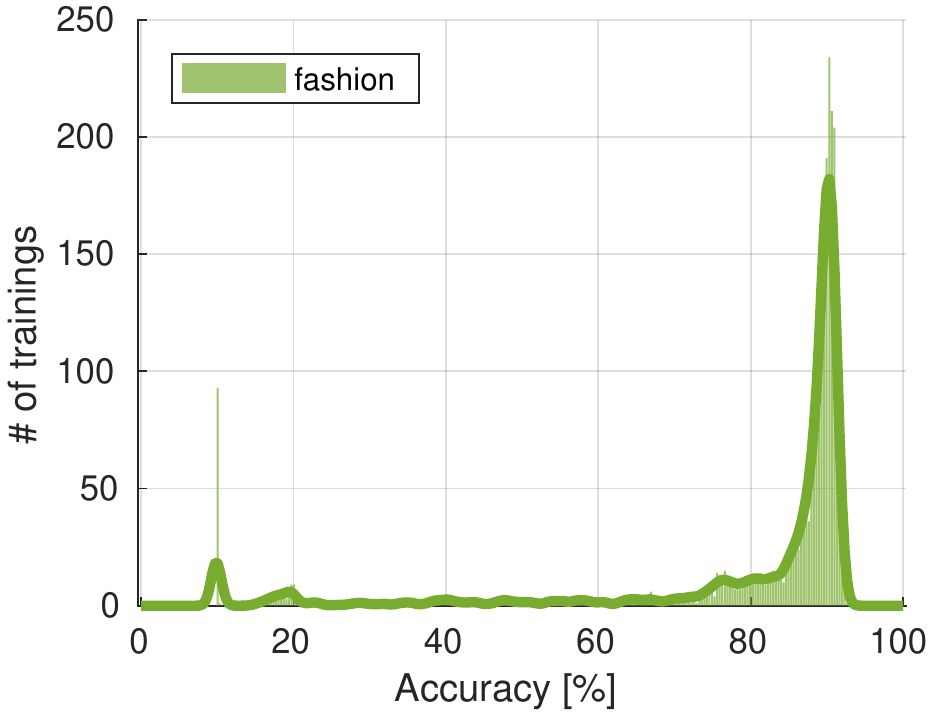}
		\caption{Fashion-MNIST (2672 trainings)}
		\label{fig:accuracy_fashion}
	\end{subfigure}
	\hspace{0.325\textwidth}
	\caption{\label{fig:accuracy} Distribution of training accuracy for each dataset. The curves show smoothed histograms to convey the general shape of the distributions.}
	\vspace{0.2cm}
\end{figure*}

To get a better picture of the spread in performance of trained models, \figref{accuracy} shows the distribution of test accuracy for each dataset. While each dataset generates performances over a wide range of values, there are always two distinct modes; one with the failed trainings and one with more successful trainings. Clearly, some datasets are more prone to fail (CIFAR-10, STL-10) than others (MNIST, Fashion-MNIST). Also, there is always an approximately normally distributed set of successful trainings, and this is also the case when only selecting the best hyper-parameters. That is, random initialization and/or SGD will result in that test performance is normally distributed over repeated training runs. 

For the failure modes, it is also possible to discern a tendency to have two peaks. One is at 10\%, i.e. random guessing, meaning that nothing is learned. However, for all datasets there is also a more or less pronounced peak around 20\%.

\begin{figure*}[t!]
	\centering
	\begin{subfigure}{0.325\textwidth}
		\includegraphics[width=\textwidth, trim={0pt 2pt 0pt 2pt}, clip]{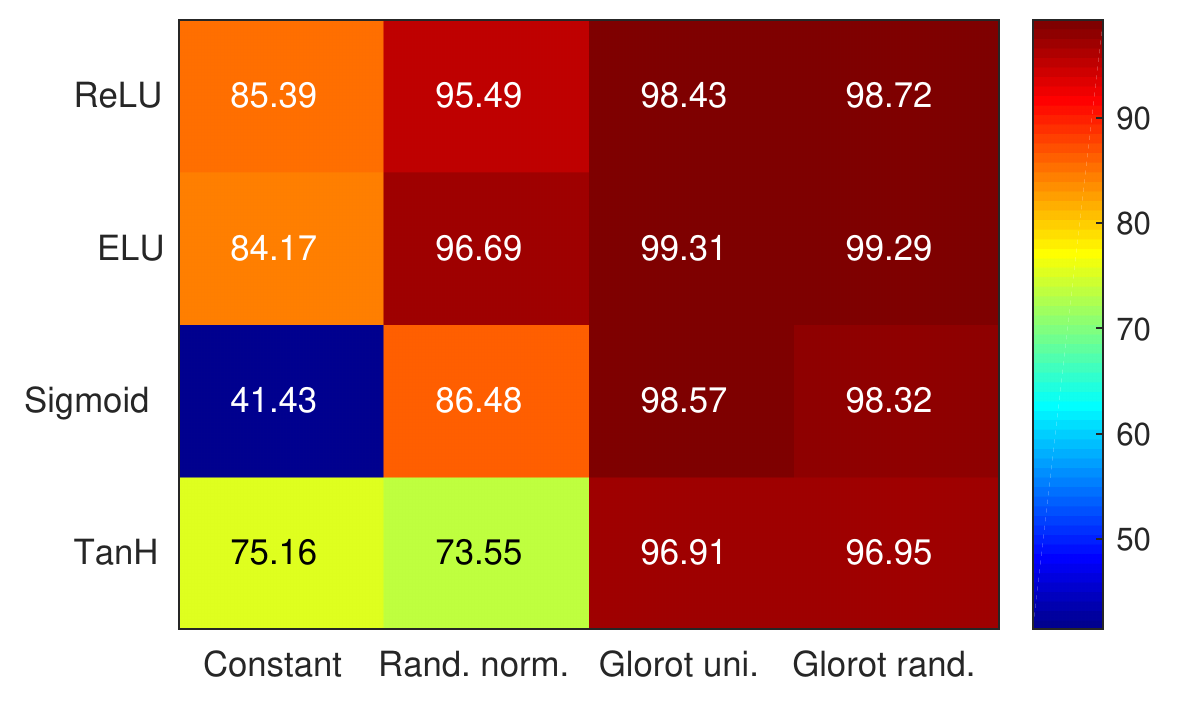}
		\caption{MNIST}
		\label{fig:init_act_mnist}
	\end{subfigure}
	\begin{subfigure}{0.325\textwidth}
		\includegraphics[width=\textwidth, trim={0pt 2pt 0pt 2pt}, clip]{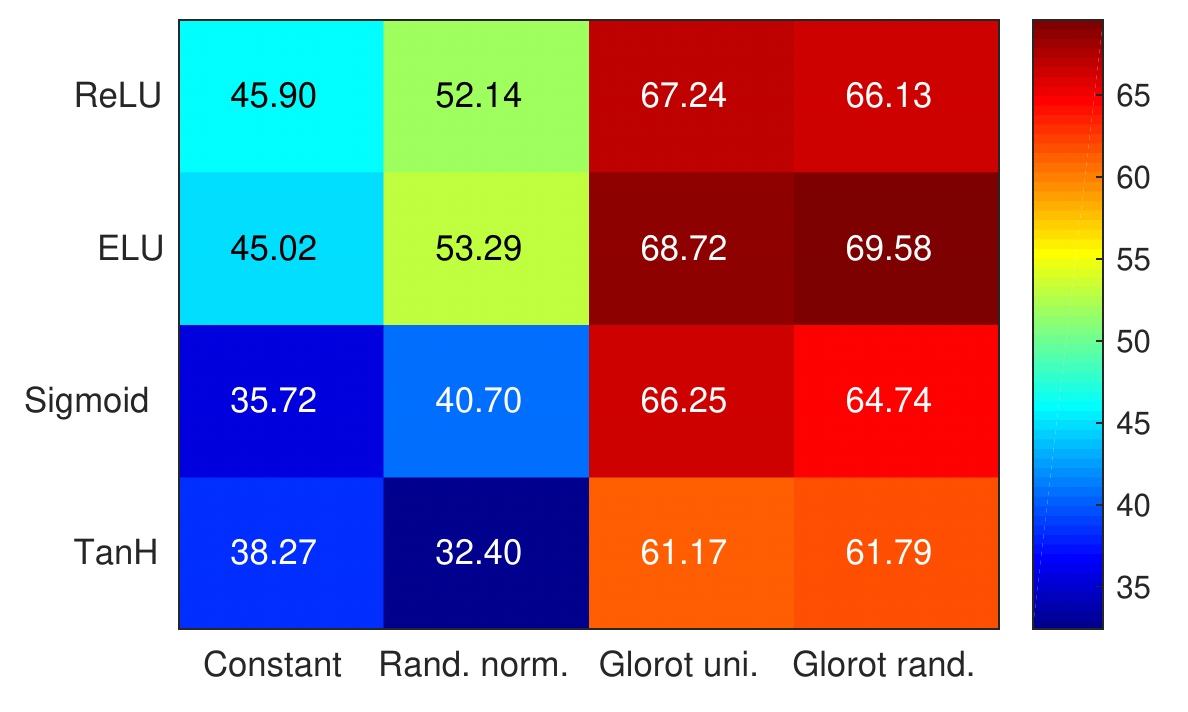}
		\caption{CIFAR-10}
		\label{fig:init_act_cifar}
	\end{subfigure}
	\begin{subfigure}{0.325\textwidth}
		\includegraphics[width=\textwidth, trim={0pt 2pt 0pt 2pt}, clip]{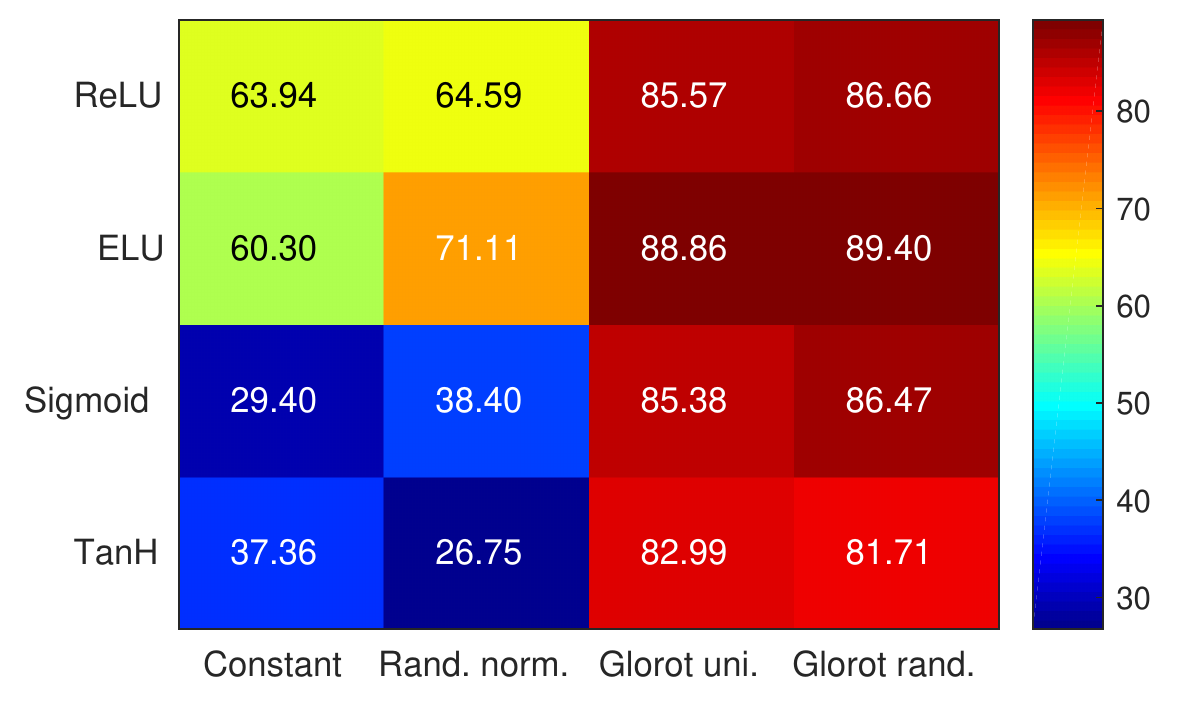}
		\caption{SVHN}
		\label{fig:init_act_svhn}
	\end{subfigure}\\
	\vspace{0.2cm}
	\begin{subfigure}{0.325\textwidth}
		\includegraphics[width=\textwidth, trim={0pt 2pt 0pt 2pt}, clip]{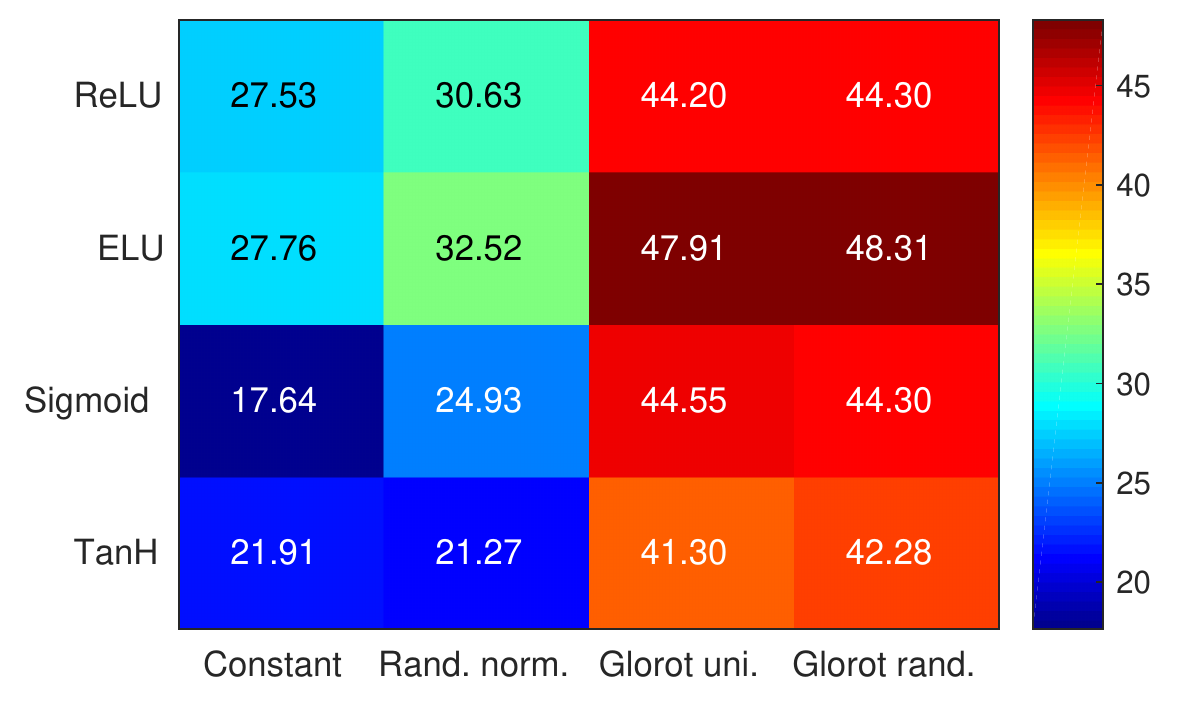}
		\caption{STL-10}
		\label{fig:init_act_stl}
	\end{subfigure}
	\begin{subfigure}{0.325\textwidth}
		\includegraphics[width=\textwidth, trim={0pt 2pt 0pt 2pt}, clip]{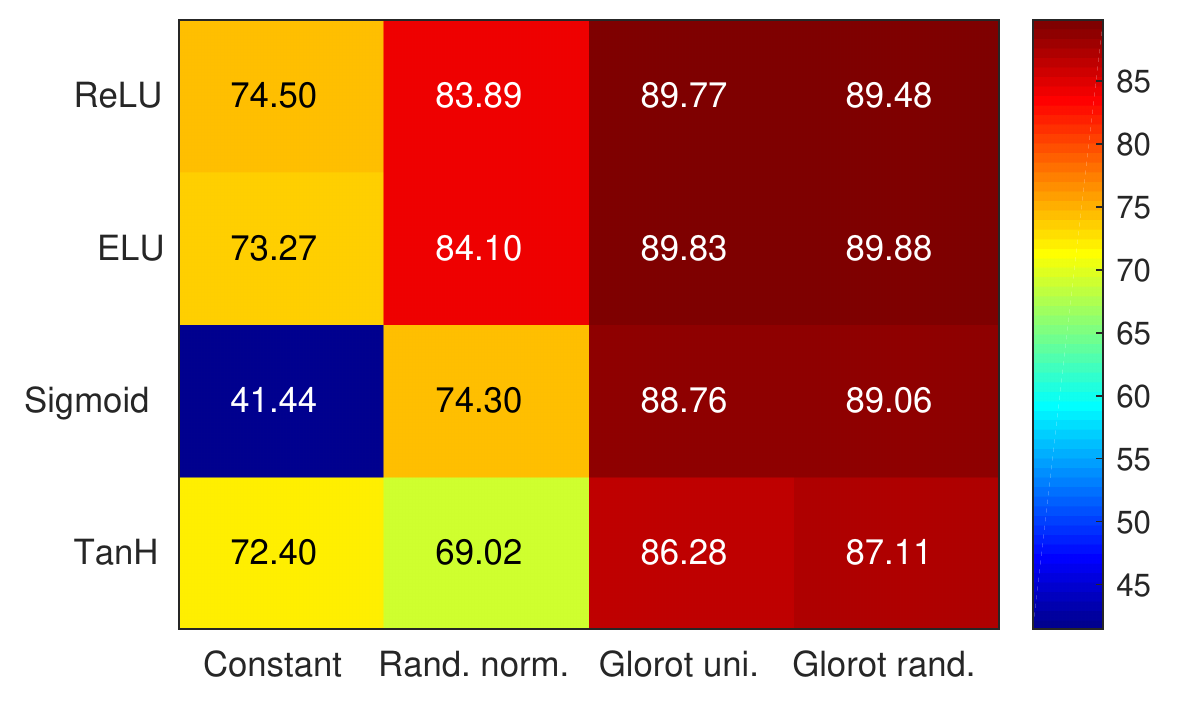}
		\caption{Fashion-MNIST}
		\label{fig:init_act_fashion}
	\end{subfigure}
	\hspace{0.325\textwidth}
	\caption{\label{fig:init_act} Average test accuracy for different combinations of initializations and activation functions.}
\end{figure*}

\begin{figure*}[t!]
	\centering
	\begin{subfigure}{0.325\textwidth}
		\includegraphics[width=\textwidth, trim={0pt 2pt 0pt 2pt}, clip]{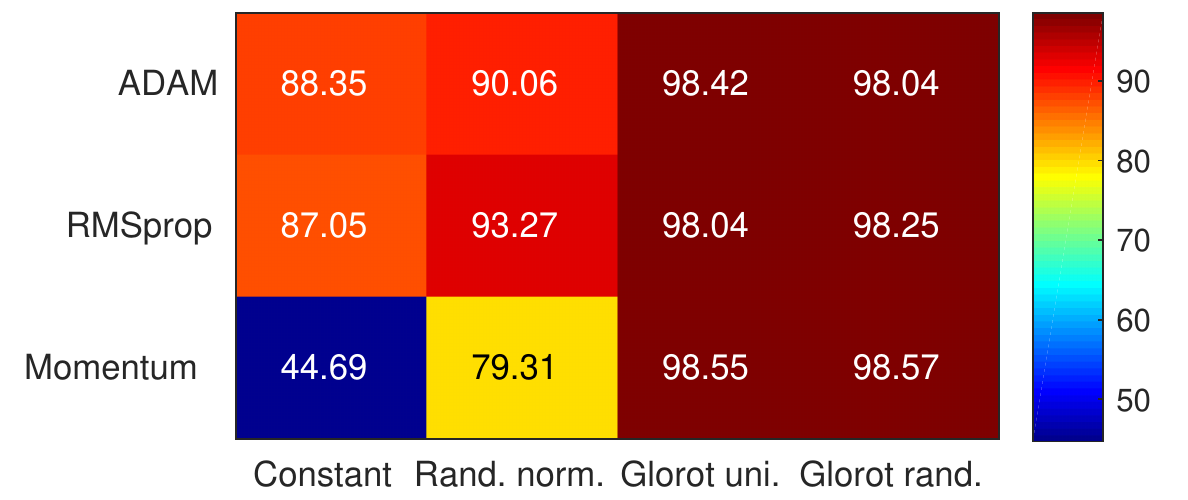}
		\caption{MNIST}
		\label{fig:init_opt_mnist}
	\end{subfigure}
	\begin{subfigure}{0.325\textwidth}
		\includegraphics[width=\textwidth, trim={0pt 2pt 0pt 2pt}, clip]{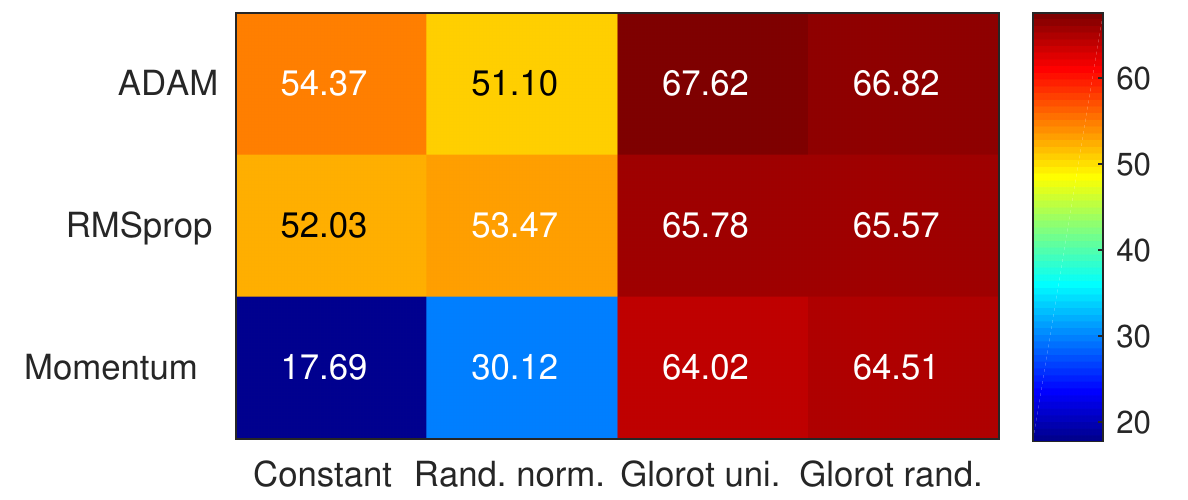}
		\caption{CIFAR-10}
		\label{fig:init_opt_cifar}
	\end{subfigure}
	\begin{subfigure}{0.325\textwidth}
		\includegraphics[width=\textwidth, trim={0pt 2pt 0pt 2pt}, clip]{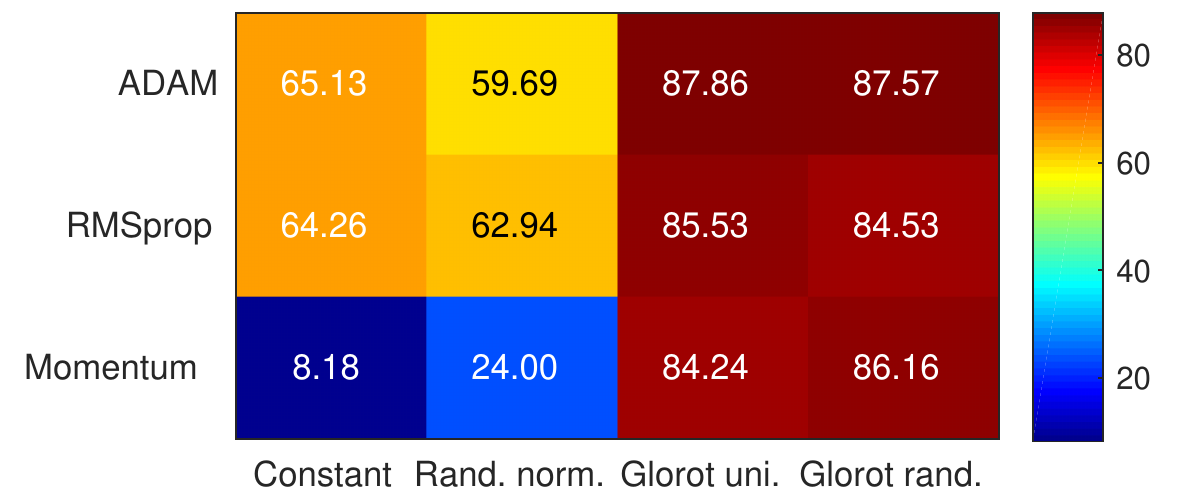}
		\caption{SVHN}
		\label{fig:init_opt_svhn}
	\end{subfigure}\\
	\vspace{0.2cm}
	\begin{subfigure}{0.325\textwidth}
		\includegraphics[width=\textwidth, trim={0pt 2pt 0pt 2pt}, clip]{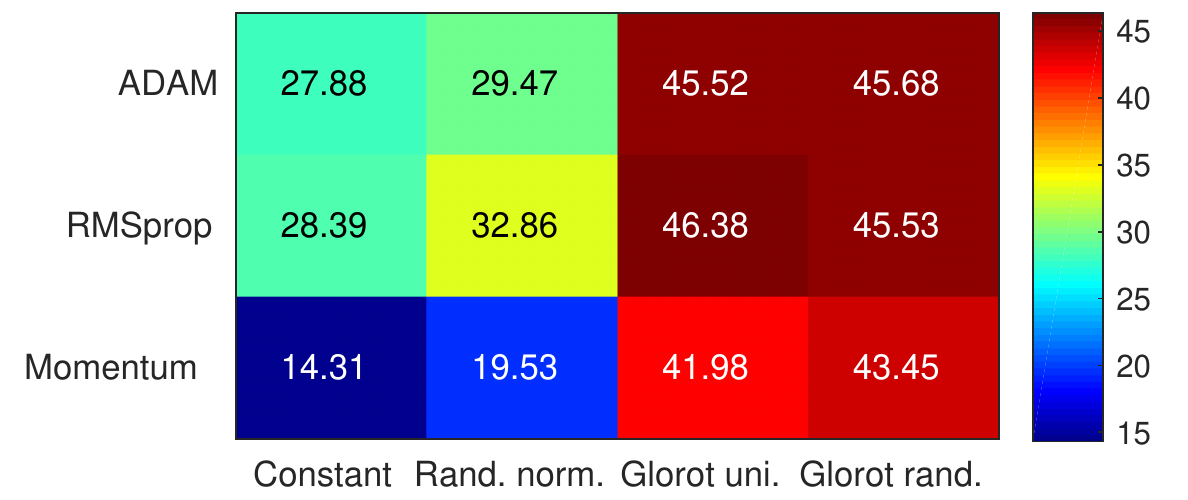}
		\caption{STL-10}
		\label{fig:init_opt_stl}
	\end{subfigure}
	\begin{subfigure}{0.325\textwidth}
		\includegraphics[width=\textwidth, trim={0pt 2pt 0pt 2pt}, clip]{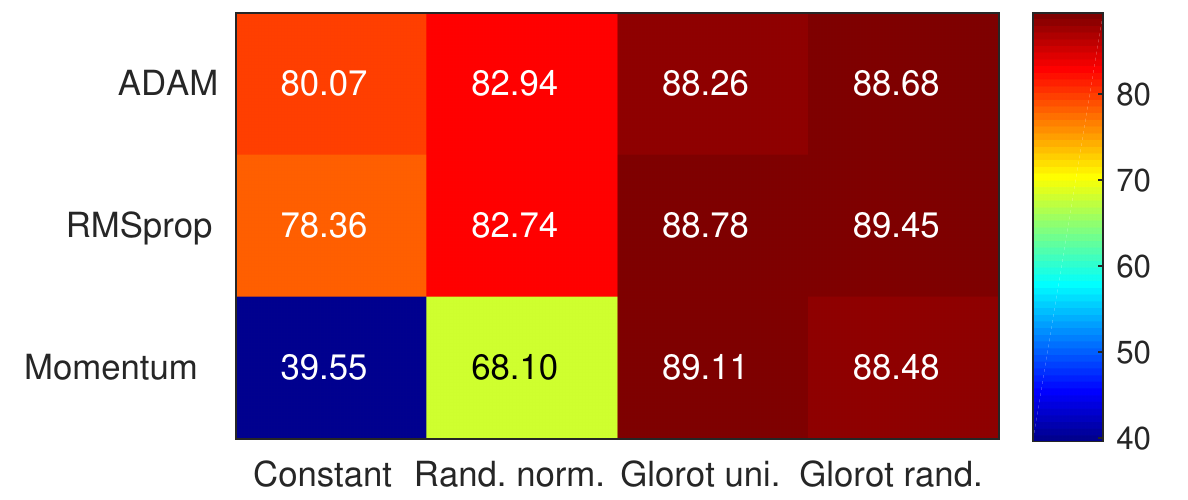}
		\caption{Fashion-MNIST}
		\label{fig:init_opt_fashion}
	\end{subfigure}
	\hspace{0.325\textwidth}
	\caption{\label{fig:init_opt} Average test accuracy for different combinations of initializations and optimizer.}
	\vspace{0.2cm}
\end{figure*}

\begin{figure*}[t!]
	\centering
	\begin{subfigure}{0.325\textwidth}
		\includegraphics[width=\textwidth, trim={0pt 2pt 0pt 2pt}, clip]{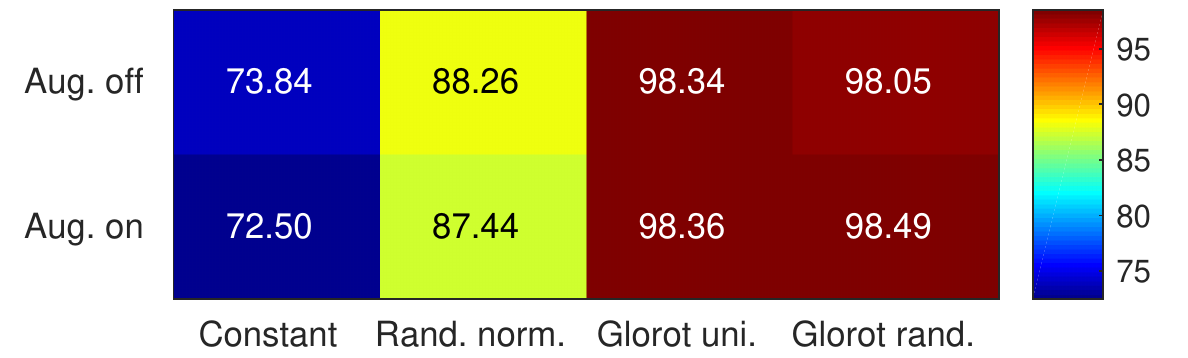}
		\caption{MNIST}
		\label{fig:init_aug_mnist}
	\end{subfigure}
	\begin{subfigure}{0.325\textwidth}
		\includegraphics[width=\textwidth, trim={0pt 2pt 0pt 2pt}, clip]{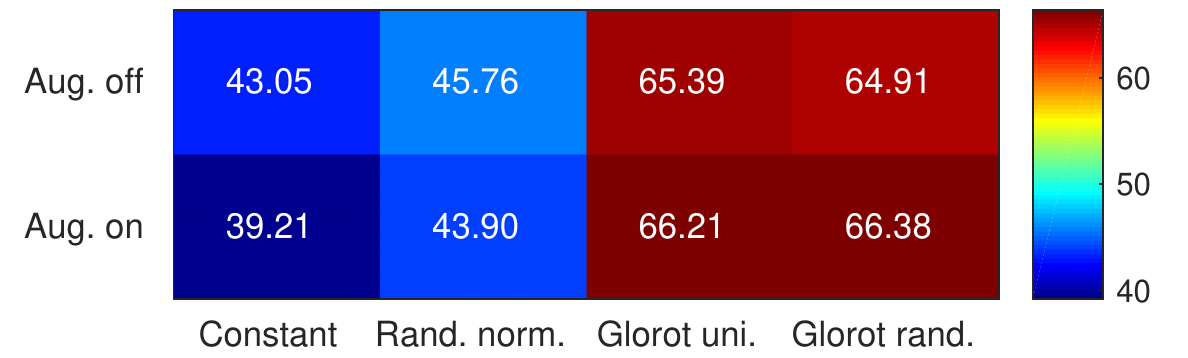}
		\caption{CIFAR-10}
		\label{fig:init_aug_cifar}
	\end{subfigure}
	\begin{subfigure}{0.325\textwidth}
		\includegraphics[width=\textwidth, trim={0pt 2pt 0pt 2pt}, clip]{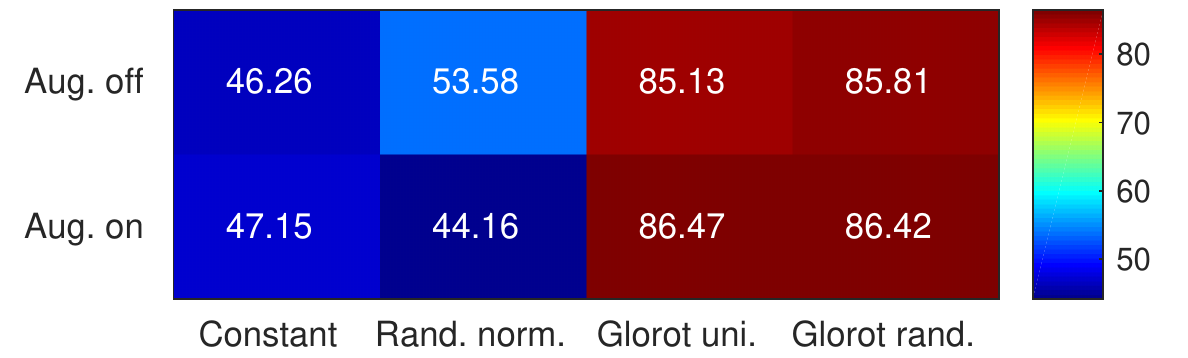}
		\caption{SVHN}
		\label{fig:init_aug_svhn}
	\end{subfigure}\\
	\vspace{0.2cm}
	\begin{subfigure}{0.325\textwidth}
		\includegraphics[width=\textwidth, trim={0pt 2pt 0pt 2pt}, clip]{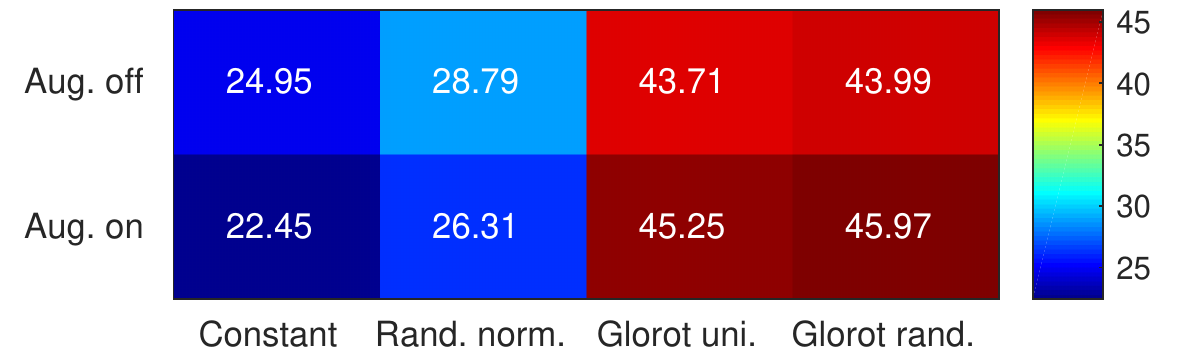}
		\caption{STL-10}
		\label{fig:init_aug_stl}
	\end{subfigure}
	\begin{subfigure}{0.325\textwidth}
		\includegraphics[width=\textwidth, trim={0pt 2pt 0pt 2pt}, clip]{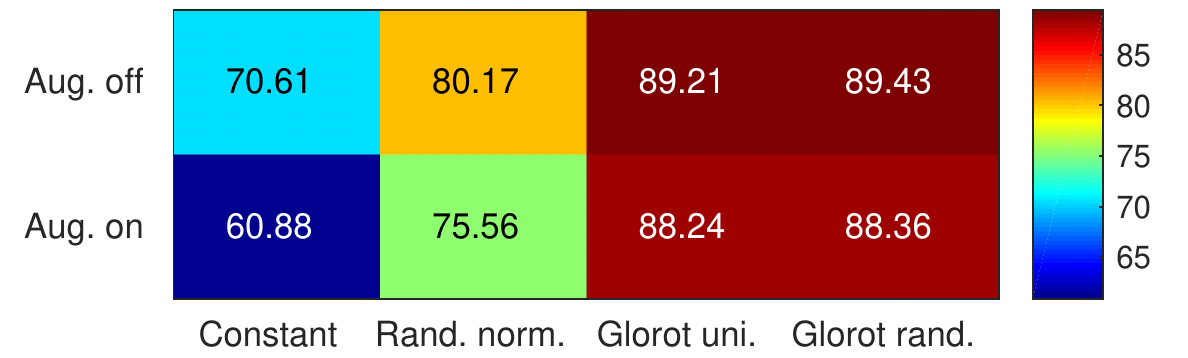}
		\caption{Fashion-MNIST}
		\label{fig:init_aug_fashion}
	\end{subfigure}
	\hspace{0.325\textwidth}
	\caption{\label{fig:init_aug} Average test accuracy for different combinations of initializations and augmentation.}
	\vspace{0.2cm}
\end{figure*}

\begin{figure*}[t!]
	\centering
	\begin{subfigure}{0.325\textwidth}
		\includegraphics[width=\textwidth, trim={0pt 2pt 0pt 2pt}, clip]{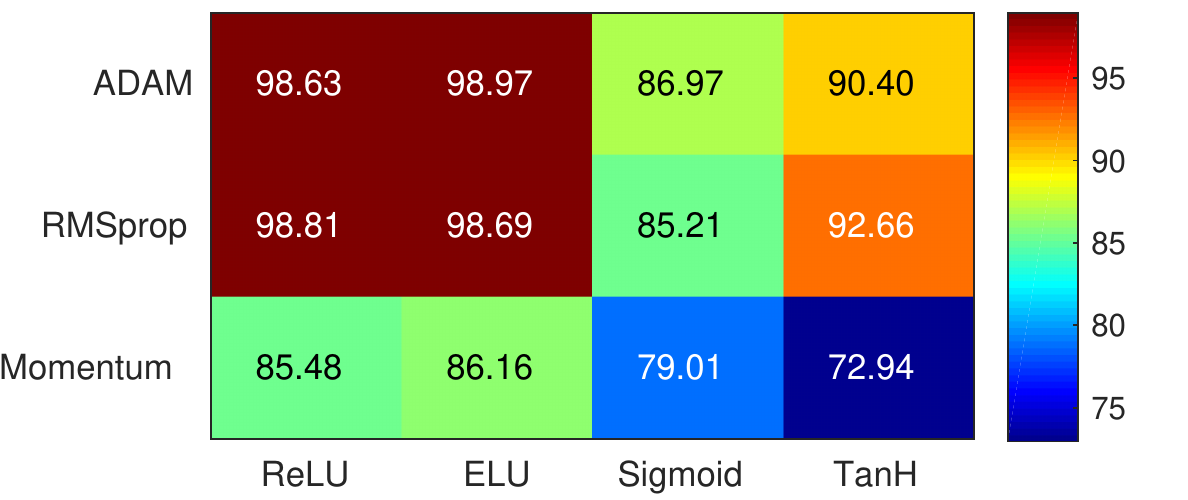}
		\caption{MNIST}
		\label{fig:act_opt_mnist}
	\end{subfigure}
	\begin{subfigure}{0.325\textwidth}
		\includegraphics[width=\textwidth, trim={0pt 2pt 0pt 2pt}, clip]{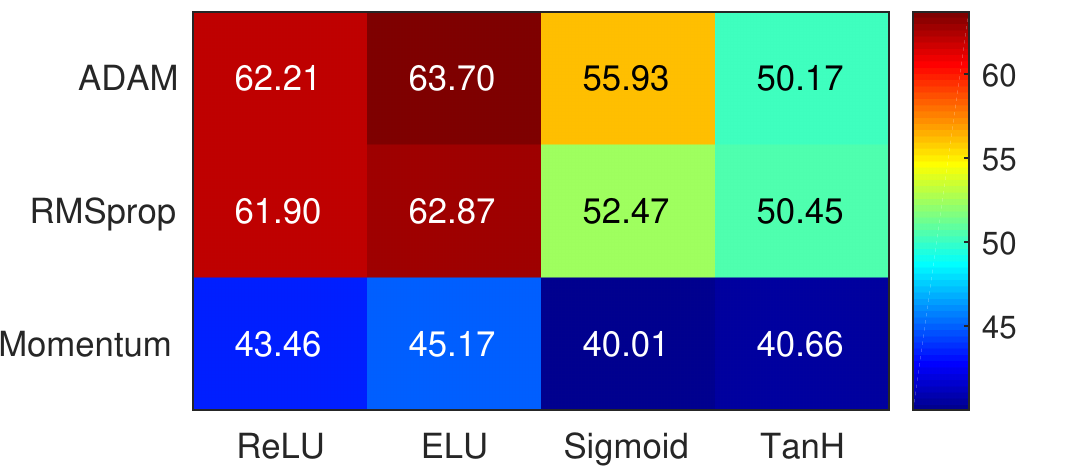}
		\caption{CIFAR-10}
		\label{fig:act_opt_cifar}
	\end{subfigure}
	\begin{subfigure}{0.325\textwidth}
		\includegraphics[width=\textwidth, trim={0pt 2pt 0pt 2pt}, clip]{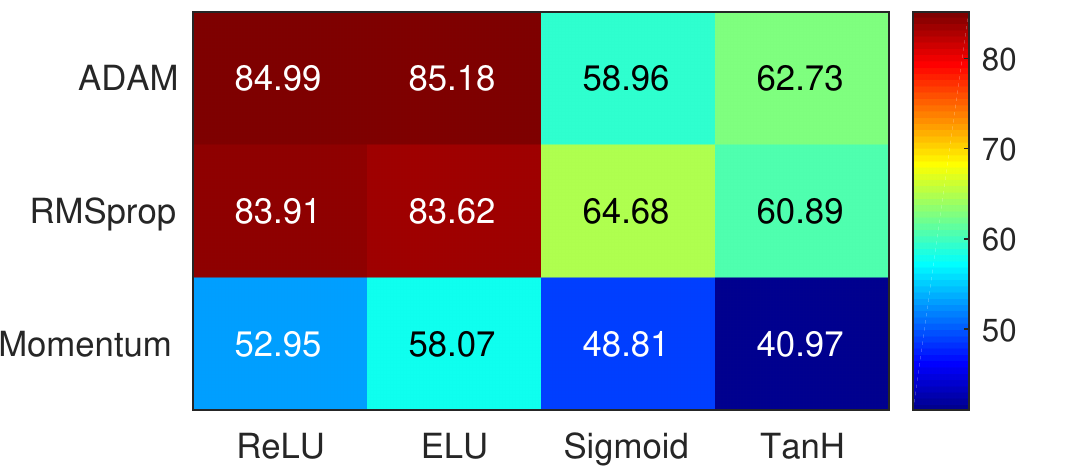}
		\caption{SVHN}
		\label{fig:act_opt_svhn}
	\end{subfigure}\\
	\vspace{0.2cm}
	\begin{subfigure}{0.325\textwidth}
		\includegraphics[width=\textwidth, trim={0pt 2pt 0pt 2pt}, clip]{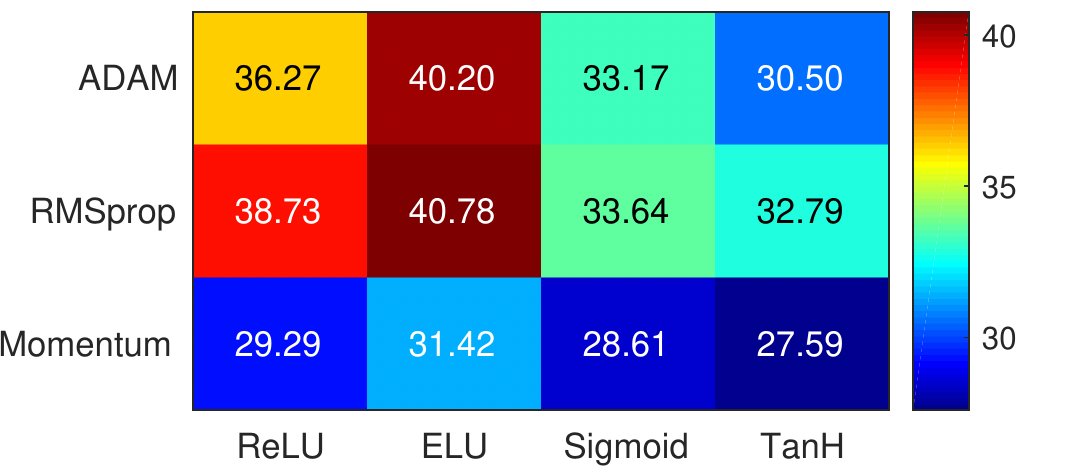}
		\caption{STL-10}
		\label{fig:act_opt_stl}
	\end{subfigure}
	\begin{subfigure}{0.325\textwidth}
		\includegraphics[width=\textwidth, trim={0pt 2pt 0pt 2pt}, clip]{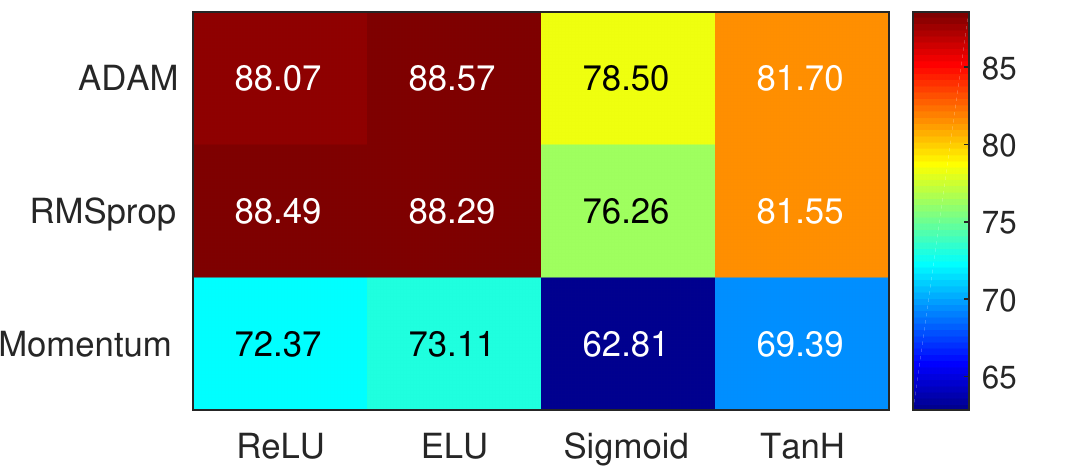}
		\caption{Fashion-MNIST}
		\label{fig:act_opt_fashion}
	\end{subfigure}
	\hspace{0.325\textwidth}
	\caption{\label{fig:act_opt} Average test accuracy for different combinations of activation function and optimizer.}
\end{figure*}

Using the test accuracies, we can also analyze the performance of different combinations of hyper-parameters. \figref{init_act} shows all combinations of activation function and initialization strategy, split between different datasets. We can make a number of observations from the result. For example, Glorot initialization is clearly the superior choice of the initialization schemes included, both using uniformly and normally distributed weights. ELU consistently provides a slight advantage over ReLU, and especially for the more difficult datasets (CIFAR-10 and STL-10). There is also an interesting pattern of sigmoid and TanH being clearly inferior to ReLU and ELU when combined with constant or random normal initialization. Constant initialization combined with sigmoid activation is the far worst combination, but for other initialization strategies sigmoid performs better than TanH. When using Glorot initialization, the difference between ReLU/ELU and sigmoid/Tanh is much smaller. This means that on average ReLU and ELU are much better at handling difficult initialization.

\figref{init_opt} shows all combinations of optimizer and initialization strategy for the different datasets. When using Glorot initialization, the momentum optimizer performance is close to that of ADAM and RMSprop. However, ADAM and RMSprop are much better at dealing with the more difficult initialization schemes, although the performance is quite far from the results when using Glorot initialization. It is possible that momentum optimization could behave better in the beginning of optimization with a lower momentum setting \cite{Hinton2012}, when gradients are large, whereas a higher momentum is preferable later on. That is, a different momentum could give the opposite pattern of what we see here -- better handling of bad initialization, but worse at converging to the more optimal end points.

\figref{init_aug} shows performances of different initialization strategies when augmentations are turned off and on. When studying correlation between hyper-parameters and test performance in \figref{linear_reg} it seems like augmentation has little effect. However, this does not reveal the complete picture. Looking at performance for different initializations in \figref{init_aug}, for Glorot initialization it improves performance on most datasets, while decreasing performance for less effective initialization schemes.

\figref{act_opt} shows all combinations of optimizer and activation function. The trends are similar as in \figref{init_act} and \ref{fig:init_opt}, with ADAM or RMSprop together with ReLU or ELU being the optimal hyper-parameter selection.

\section{Dimensionality reduction}
\begin{figure*}[t!]
	\centering
	\begin{subfigure}{0.49\textwidth}
		\includegraphics[width=\textwidth, trim={40pt 28pt 32pt 20pt}, clip]{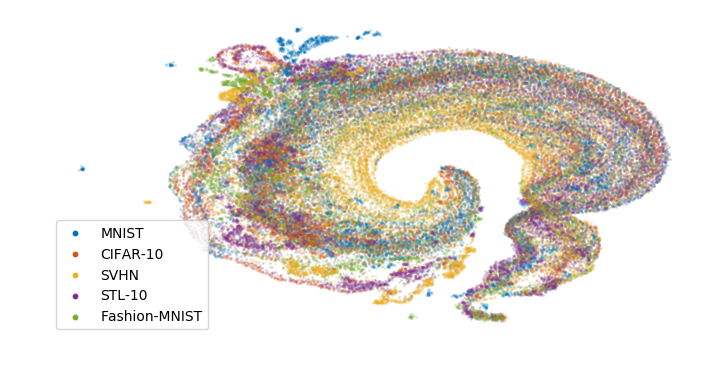}
		\caption{Dataset}
		\label{fig:umap_00}
	\end{subfigure}
	\begin{subfigure}{0.49\textwidth}
		\includegraphics[width=\textwidth, trim={40pt 28pt 43pt 15pt}, clip]{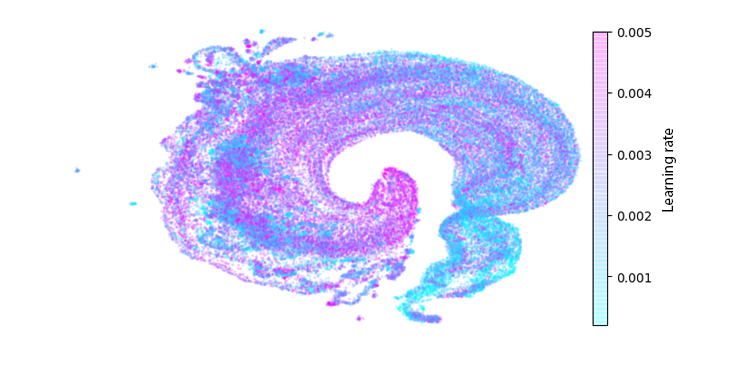}
		\caption{Learning rate}
		\label{fig:umap_01}
	\end{subfigure}\\
	\vspace{0.4cm}
	\begin{subfigure}{0.49\textwidth}
		\includegraphics[width=\textwidth, trim={40pt 28pt 32pt 20pt}, clip]{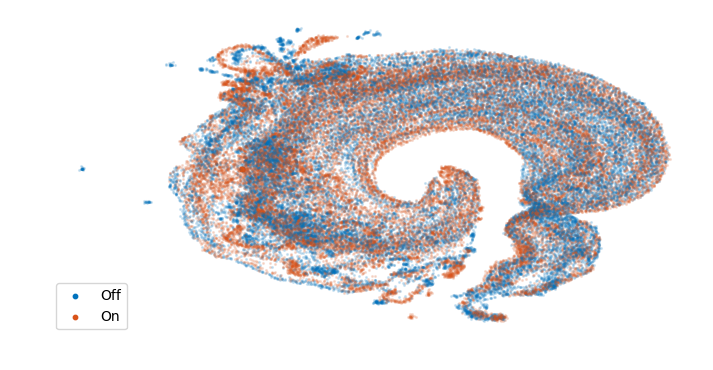}
		\caption{Augmentation}
		\label{fig:umap_04}
	\end{subfigure}
	\begin{subfigure}{0.49\textwidth}
		\includegraphics[width=\textwidth, trim={40pt 28pt 32pt 20pt}, clip]{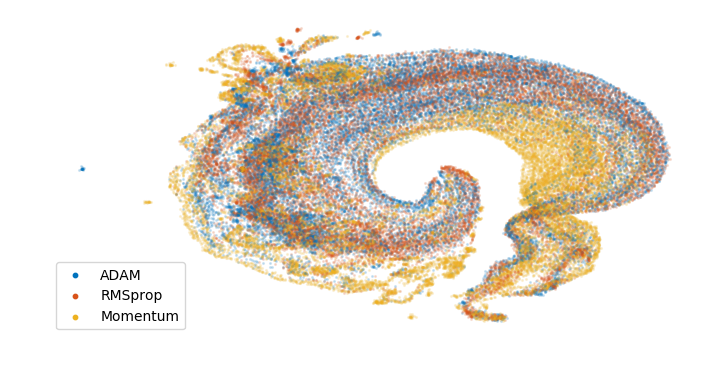}
		\caption{Optimizer}
		\label{fig:umap_05}
	\end{subfigure}\\
	\vspace{0.4cm}
	\begin{subfigure}{0.49\textwidth}
		\includegraphics[width=\textwidth, trim={40pt 28pt 32pt 20pt}, clip]{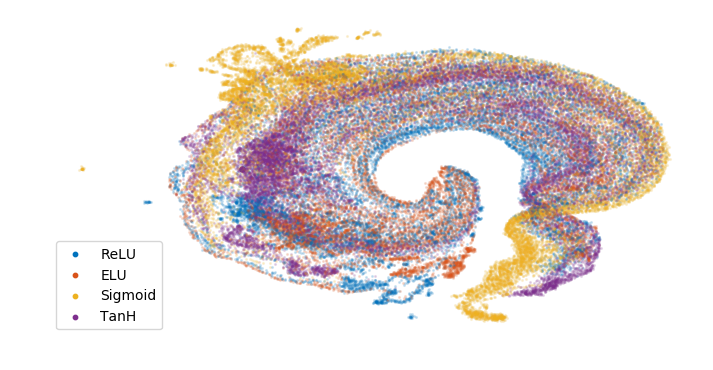}
		\caption{Activation function}
		\label{fig:umap_06}
	\end{subfigure}
	\begin{subfigure}{0.49\textwidth}
		\includegraphics[width=\textwidth, trim={40pt 28pt 32pt 20pt}, clip]{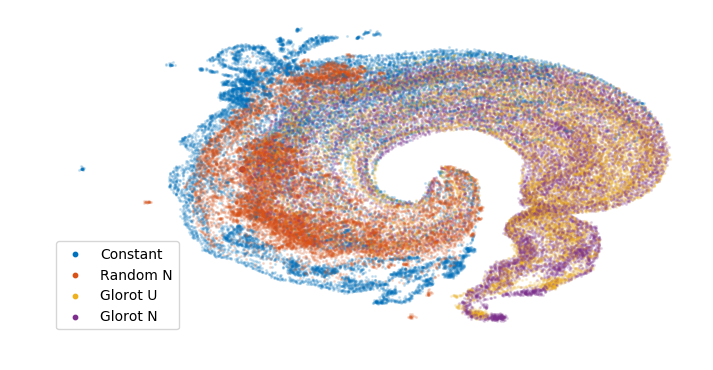}
		\caption{Initialization}
		\label{fig:umap_07}
	\end{subfigure}\\
	\vspace{0.4cm}
	\begin{subfigure}{0.49\textwidth}
		\includegraphics[width=\textwidth, trim={40pt 28pt 45pt 20pt}, clip]{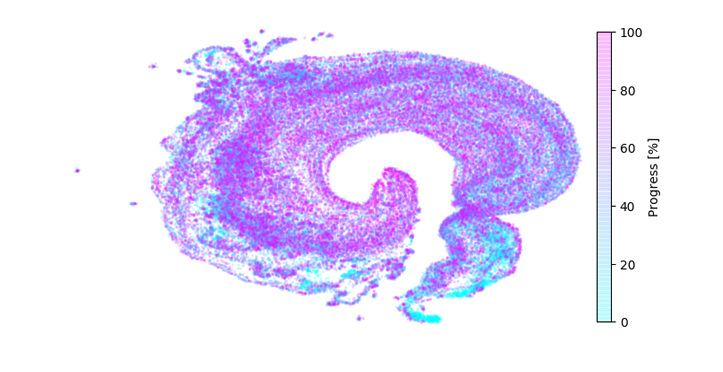}
		\caption{Training progress}
		\label{fig:umap_progress}
	\end{subfigure}
	\begin{subfigure}{0.49\textwidth}
		\includegraphics[width=\textwidth, trim={40pt 28pt 45pt 20pt}, clip]{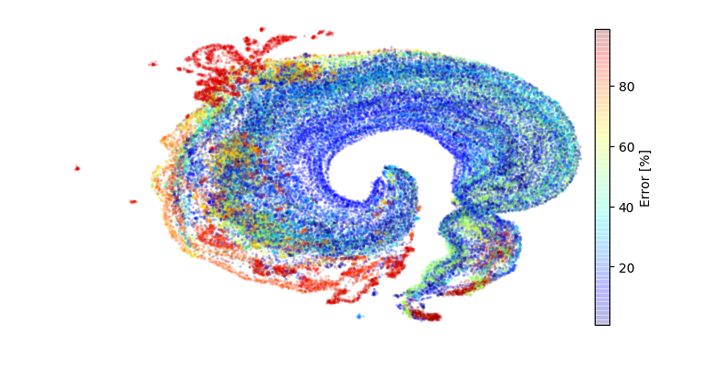}
		\caption{Test error}
		\label{fig:umap_error}
	\end{subfigure}
	\hspace{0.49\textwidth}
	\caption{\label{fig:umap} UMAP embeddings of the 10 first PCA components of 60K weight samples from 3K individual trainings of $C_{fixed}$ from the NWS dataset, color coded according to the different hyper-parameters used in training (a-f). The bottom row (g-h) also illustrates training progress and test error.}
\end{figure*}

Previous work have demonstrated the weights of a single or few trainings from the perspective of PCA components~\cite{Gallagher1997a,Lorch2016,Antognini2018}. However, we can make use of thousands of separate trainings to perform PCA, and a rather small set of components holds the majority of variance of the data. \figref{umap} shows a UMAP embedding~\cite{Mcinnes2018} of the 10 first principal components of 3K separate trainings ($C_{fixed}$ in Table 1 in the paper). 
These have in total 92,868 model weights each, of which 16,880 are from the convolutional layers and 75,988 from the FC layers. While it is possible to run PCA on the full weight representations, we choose to focus only on the convolutional weights to be able to use more models for the embedding. 
For each training, 20 points in the weight space have been included along the training trajectory from initialization to final converged model, for a total of $60K$ weight vectors. For embedding in 2D, we perform UMAP \cite{Mcinnes2018} on the 10 first principal components. The result is 60K 2D points, which can be scattered with different colors to represent hyper-parameters. \figref{umap} shows the same embedding, but color coded according to 6 different hyper-parameter settings used in the training of the different weights. The figure also illustrates the training progress and test error of each point. Some of the hyper-parameter options show clear patterns, such as optimizer, activation function and initialization. 

The majority of initialization points cluster in the bottom right (\figref{umap_progress}). From the different types of initialization (\figref{umap_07}) and the error (\figref{umap_error}), it is clear how many trainings with constant initialization fail, especially when using Sigmoid activation (\figref{umap_06}). Constant initialization and Sigmoid activation was also shown to be the worst combination of hyper-parameters in \figref{init_act}.
There are also small clusters of points outside the main manifold of points; these are examples of trainings that get stuck and fail to learn useful information.


\section{Sub-sampling of the NWS}\label{sec:subsampling}
For the meta-learning experiments (regression of test accuracy from hyper-parameters and meta-classification), we use a sub-sampled set of the NWS. As shown in \figref{accuracy}, the test accuracy distributions show different modes of ``failed'' and ``successful'' trainings. For our meta-learning experiments we are interested in differentiating between the successful trainings. Therefore, we only include trainings of this mode, by using a threshold on test accuracy for rejecting the failed models.

The threshold accuracies for separating the modes are 80\%, 25\%, 50\%, 25\%, and 50\% for MNIST, CIFAR-10, SVHN, STL-10, and Fashion-MNIST, respectively. The exact choice of threshold is not critical since there are relatively few trainings around these values. For the 10K/3K training/test set of $C_{main}$, this selection means that we have 8,035/2,448 training/test samples. For the 2K/1K training/test set of $C_{fixed}$, it means 1,758/880 training/test samples.

\section{Regressing the test accuracy}
\begin{figure*}[t!]
	\centering
	\begin{subfigure}{0.48\textwidth}
		\includegraphics[width=\textwidth, trim={0pt 2pt 0pt 2pt}, clip]{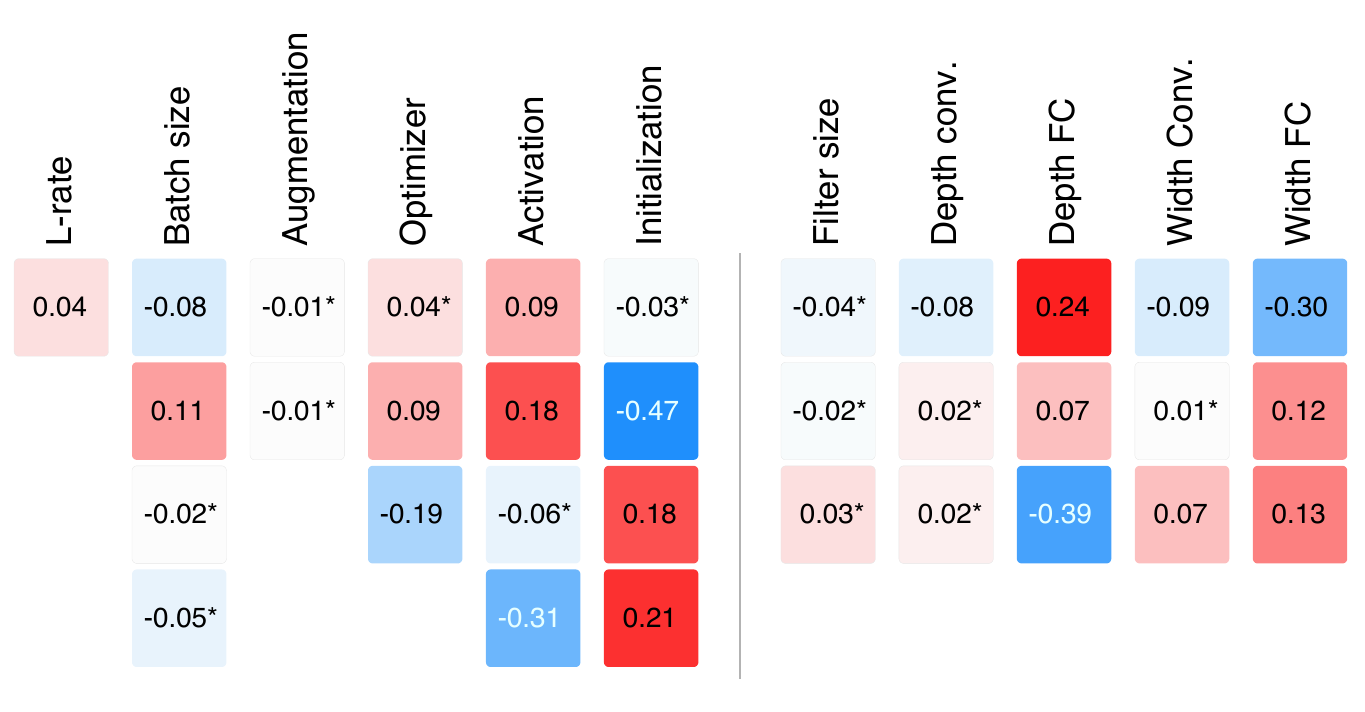}
		\caption{MNIST}
		\label{fig:linear_reg_mnist}
	\end{subfigure}
	\hspace{10pt}
	\begin{subfigure}{0.48\textwidth}
		\includegraphics[width=\textwidth, trim={0pt 2pt 0pt 2pt}, clip]{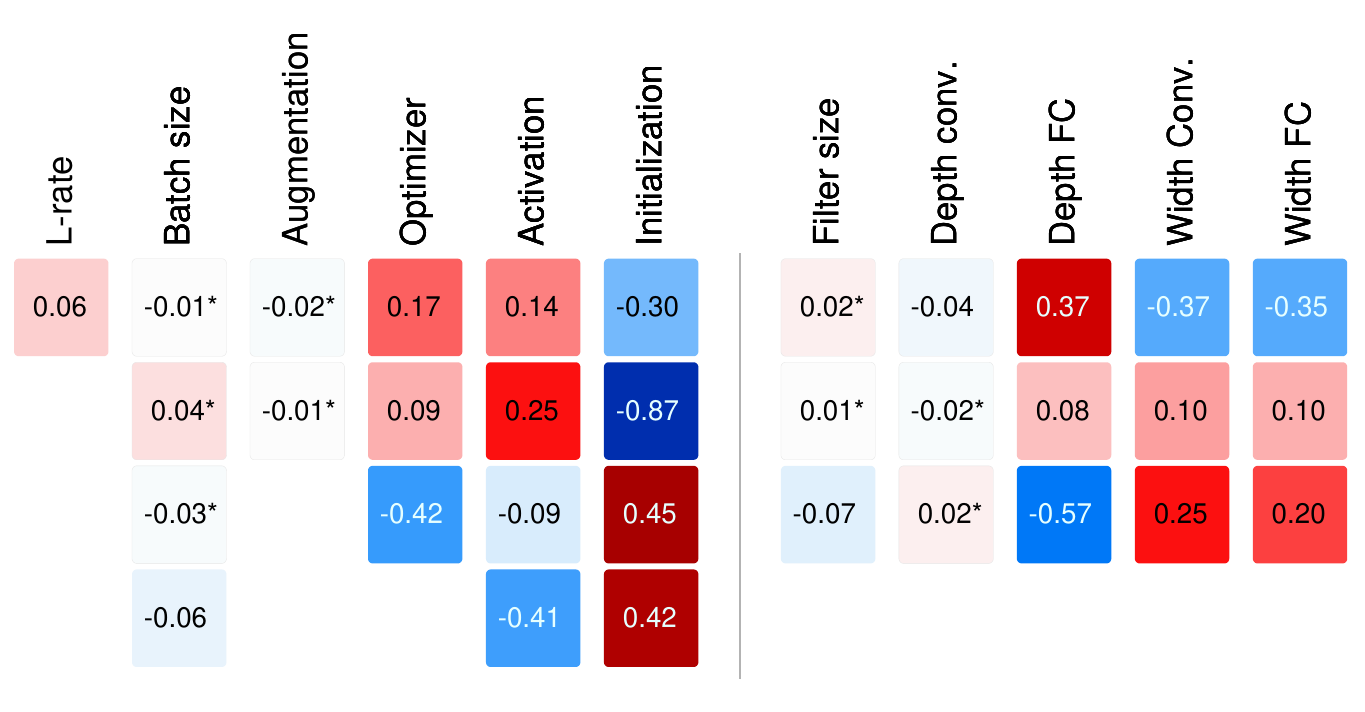}
		\caption{CIFAR-10}
		\label{fig:linear_reg_cifar}
	\end{subfigure}
	\begin{subfigure}{0.48\textwidth}
		\includegraphics[width=\textwidth, trim={0pt 2pt 0pt 2pt}, clip]{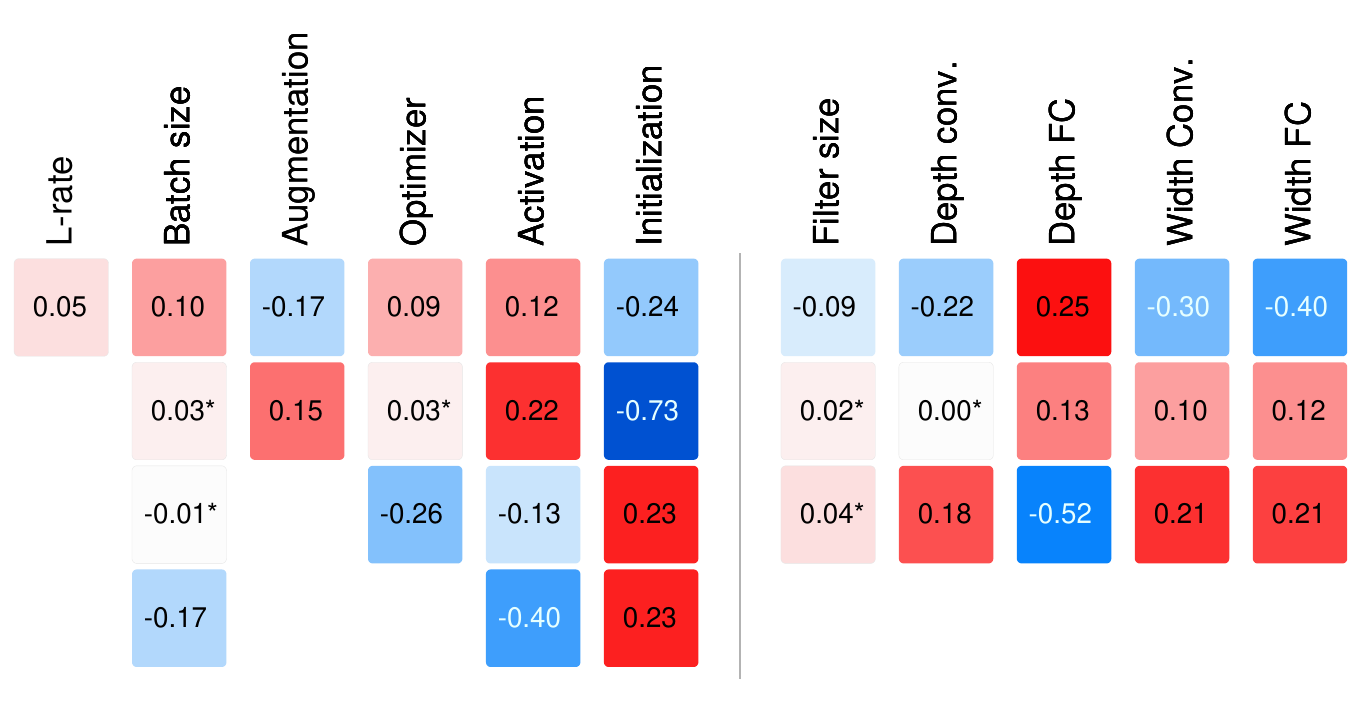}
		\caption{SVHN}
		\label{fig:linear_reg_svhn}
	\end{subfigure}
	\hspace{10pt}
	\begin{subfigure}{0.48\textwidth}
		\includegraphics[width=\textwidth, trim={0pt 2pt 0pt 2pt}, clip]{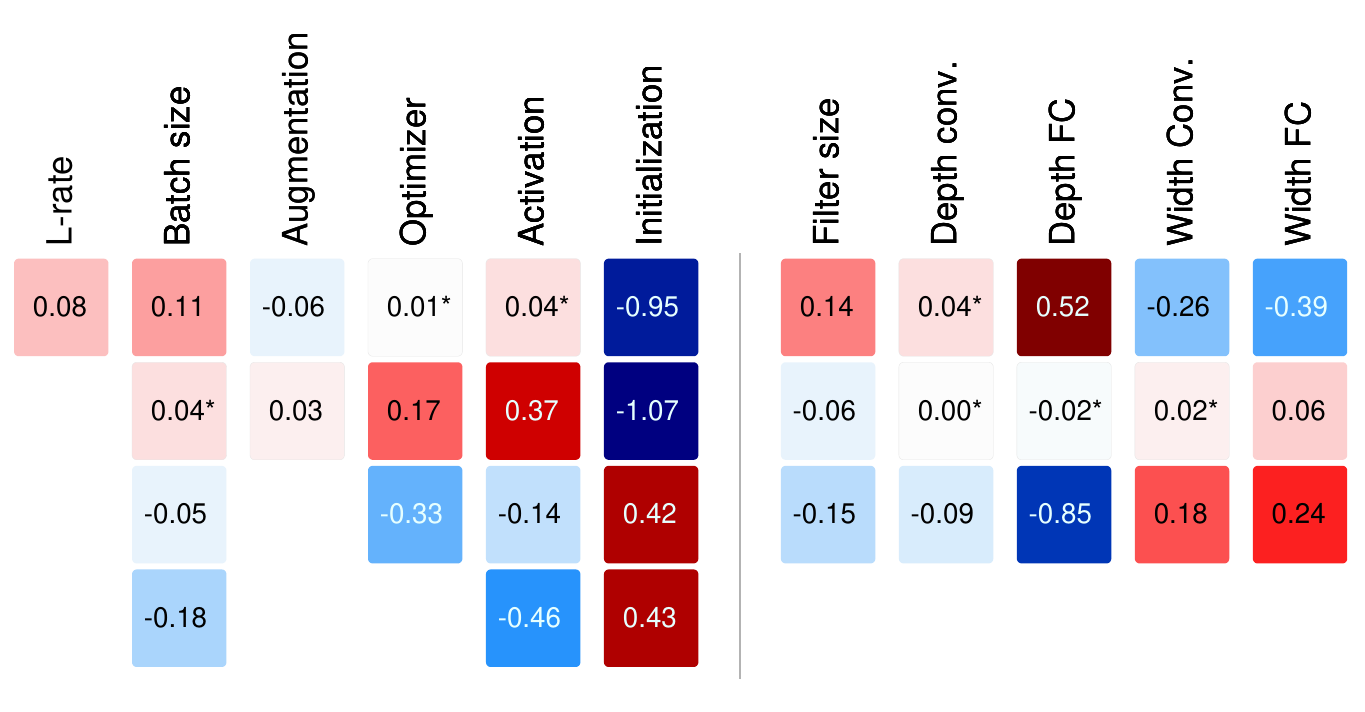}
		\caption{STL-10}
		\label{fig:linear_reg_stl}
	\end{subfigure}
	\begin{subfigure}{0.48\textwidth}
		\includegraphics[width=\textwidth, trim={0pt 2pt 0pt 2pt}, clip]{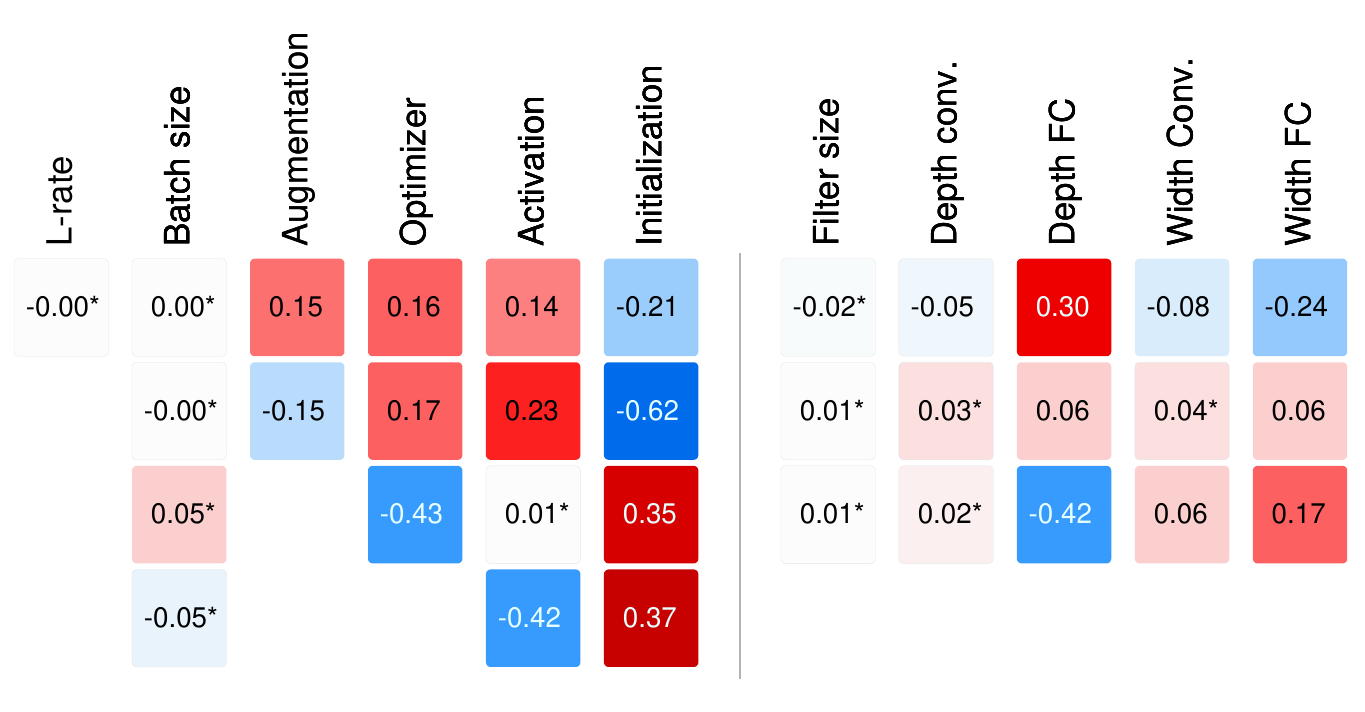}
		\caption{Fashion-MNIST}
		\label{fig:linear_reg_fashion}
	\end{subfigure}
	\hspace{0.48\textwidth}
	\caption{\label{fig:linear_reg} Regression of test error from hyper-parameters. Asterisk denotes a p-value $>$ 5\%.}
\end{figure*}

\figref{linear_reg} shows the regression coefficients for linear models fitted to the test accuracies of models from each dataset. The sub-sampled NWS dataset $C_{main}$ was used for these models. The different options for each hyper-parameter are listed in Table 1 in the main paper.

The results clearly demonstrate a vast improvement using Glorot initialization~\cite{Glorot2010} as compared to constant or random normal initialization. There is also a clear indication on the advantages of using more well-thought-out optimization strategies (ADAM, RMSprop) as compared to conventional momentum SGD. However, if we compare different combinations of optimizer and initialization (\figref{init_opt}), it is evident how ADAM and RMSprop are better at handling less optimal initializations. For Glorot initialization, momentum SGD can perform on par with these. 
Clearly, optimization is very much dependent on the initialization and optimizer. 
Another important factor is the activation function, where ReLU and ELU are clearly superior compared to Sigmoid and TanH. Morover, there is an overall tendency to favor ELU over ReLU. For architecture specific hyper-parameters, there is a general trend to promote wider models. However, the number of FC layers has a consistently negative correlation. Although a large number of FCs do not necessarily increase performance (for example, AlexNet \cite{Krizhevsky2012} and VGG \cite{Simonyan2014} have only 3), it is interesting how more than 3 layers overall results in decreased performance. One possible explanation could be that many FCs are hard to optimize without using skip-connection/resnet designs~\cite{He2016}, especially when using less optimal hyperparameters.

\begin{figure*}[t]
	\centering
	\begin{subfigure}{0.4\textwidth}
		\includegraphics[width=\textwidth, trim={0pt 2pt 0pt 2pt}, clip]{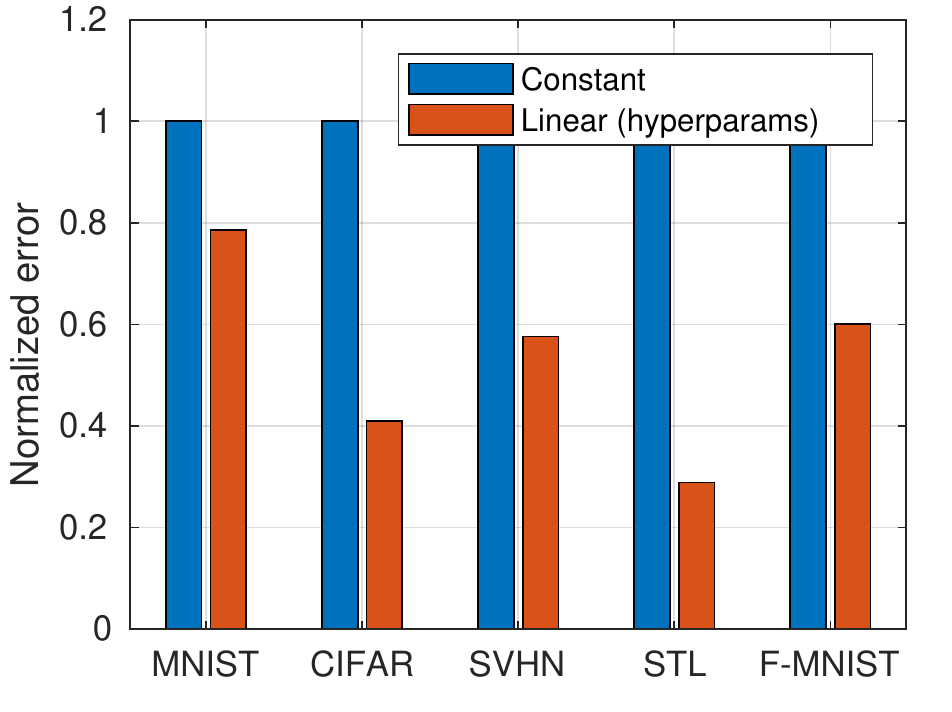}
		\caption{Linear regression errors.}
		\label{fig:linear_reg_error}
		\vspace{5pt}
	\end{subfigure}
	\hspace{20pt}
	\begin{subfigure}{0.4\textwidth}
		\includegraphics[width=\textwidth, trim={0pt 2pt 0pt 2pt}, clip]{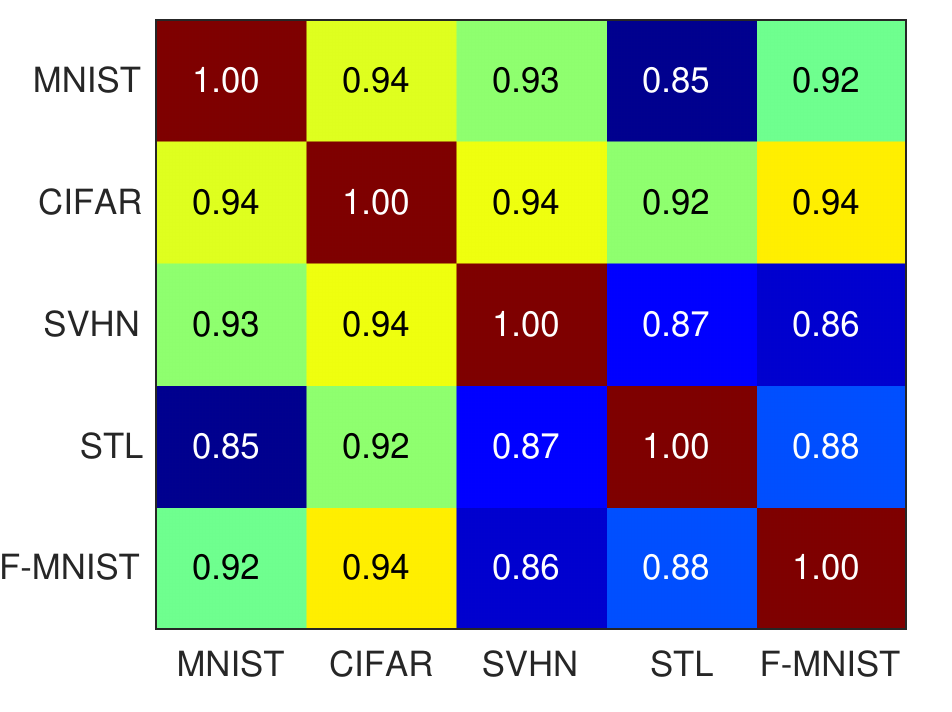}
		\caption{Linear model coefficient correlations.}
		\label{fig:linear_reg_corr}
	\end{subfigure}
	\caption{\label{fig:linear_reg_stat} The error of the linear models fitted to the test accuracy of individual datasets (left) and the Pearson's correlation between regression coefficients of all combinations of datasets (right).}
\end{figure*}

To get an idea of how well the models capture the test accuracies of the datasets, \figref{linear_reg_error} shows the average error of the models. The errors of the linear models are compared to constant models, i.e. which simply uses the mean test accuracy. Since the test accuracies have been normalized before fitting the models, the mean test accuracy is 1. For the easier datasets (MNIST, F-MNIST), the error is higher than for the more difficult ones (CIFAR-10, STL-10).

From \figref{linear_reg} we see similar patterns of optimal hyper-parameters for all datasets, and it is difficult to say which datasets require the most similar hyper-parameter tuning. In order to measure how well hyper-parameter tuning correlates between different datasets, we measure the Pearson's correlation between the model coefficients of all combinations of datasets. The results are displayed in \figref{linear_reg_corr}. From the correlations, we can for example see that MNIST requires a hyper-parameter tuning that is more similar to CIFAR-10 than STL-10, and that STL-10 on average has the least similar coefficients. However, all datasets are strongly correlated in terms of optimal hyper-parameters.

\section{SVM meta-classifier trainings}

\subsection{Experimental setup}
SVMs are trained by considering both a weight vector $\theta$ and its gradients $\nabla\theta$. From these, standard statistical measures are calculated, and include: mean, variance, skewness, 1\% percentile, 25\% percentile, 50\% percentile, 75\% percentile, and 99\% percentile. The skewness is described by the third order moment,
\begin{equation}
s = E\left[ \left( \frac{\theta-\mu}{\sigma} \right)^3 \right],
\end{equation}
where $\mu$ and $\sigma$ is the mean and standard deviation of $\theta$, respectively. This yields a total of 16 features. For the layer-wise statistics, each type of weight in each layer is considered separately. The type of weights include multiplicative weights (filters in convolutional layers, and weight matrices of fully connected layers), bias weights, and the $\beta$, $\gamma$, running mean and variance used for the batch normalization of a layer.

SVMs are trained in a one-versus-rest strategy, i.e. where one SVM is trained for each class of a dataset, drawing a decision boundary in feature space between that class and the rest of the classes.

\subsection{Feature importance}
Using a linear kernel of an SVM, we can get an indication on which features are most important for a decision. Given the coefficients $\tau_c$ of a one-versus-rest SVM, it describes a vector that is orthogonal to the hyper-plane in feature space which separates the class $c$ from the rest of the classes. The vector is oriented along the feature axes which are most useful for separating the classes. Taking the norm of $\tau_c$ over the classes $c$, we can get an indication on which features were most important for separating the classes.

\begin{figure*}[t!]
	\centering
	\begin{subfigure}{0.9\textwidth}
		\includegraphics[width=\textwidth, trim={0pt 2pt 0pt 2pt}, clip]{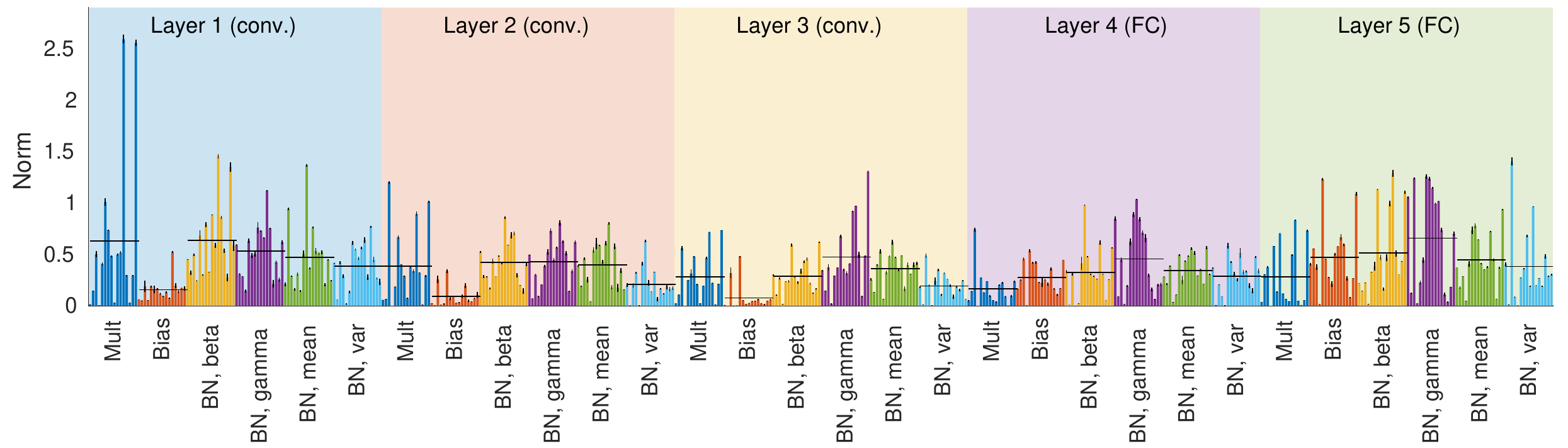}
		\caption{Dataset}
		\label{fig:coeff_global_lw_dataset}
	\end{subfigure}\\
	\vspace{0.2cm}
	\begin{subfigure}{0.9\textwidth}
		\includegraphics[width=\textwidth, trim={0pt 2pt 0pt 2pt}, clip]{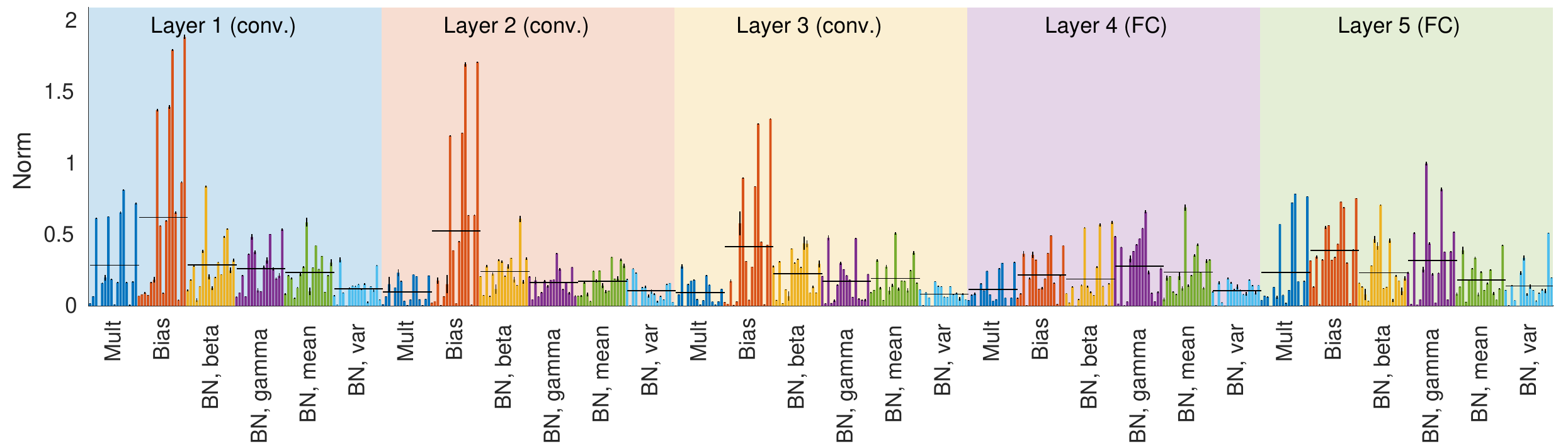}
		\caption{Optimizer}
		\label{fig:coeff_global_lw_opt}
	\end{subfigure}\\
	\vspace{0.2cm}
	\begin{subfigure}{0.9\textwidth}
		\includegraphics[width=\textwidth, trim={0pt 2pt 0pt 2pt}, clip]{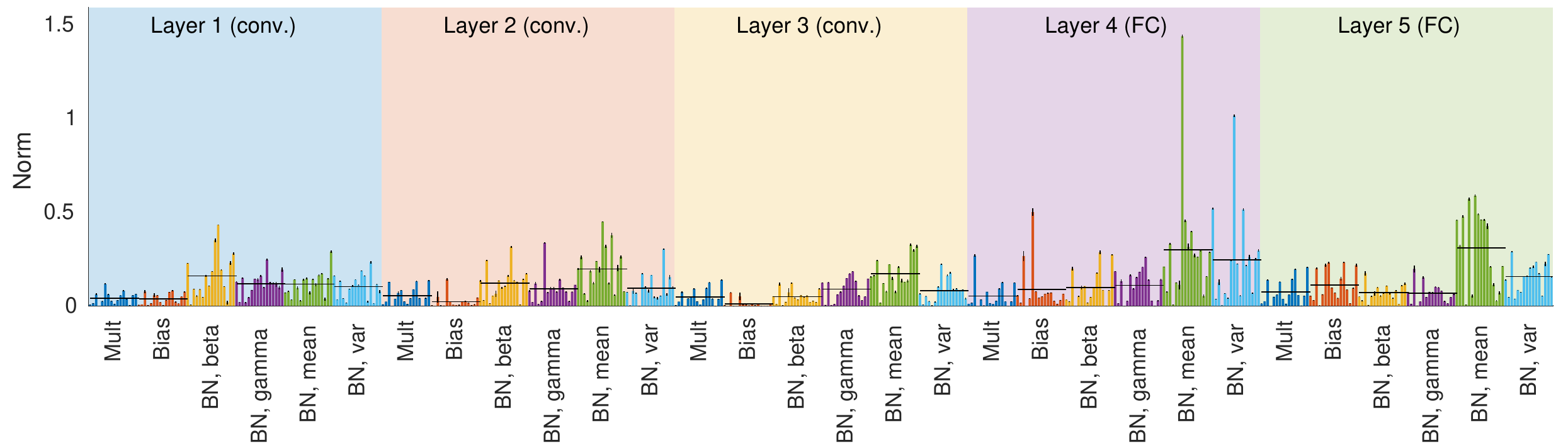}
		\caption{Activation}
		\label{fig:coeff_global_lw_act}
	\end{subfigure}\\
	\vspace{0.2cm}
	\begin{subfigure}{0.9\textwidth}
		\includegraphics[width=\textwidth, trim={0pt 2pt 0pt 2pt}, clip]{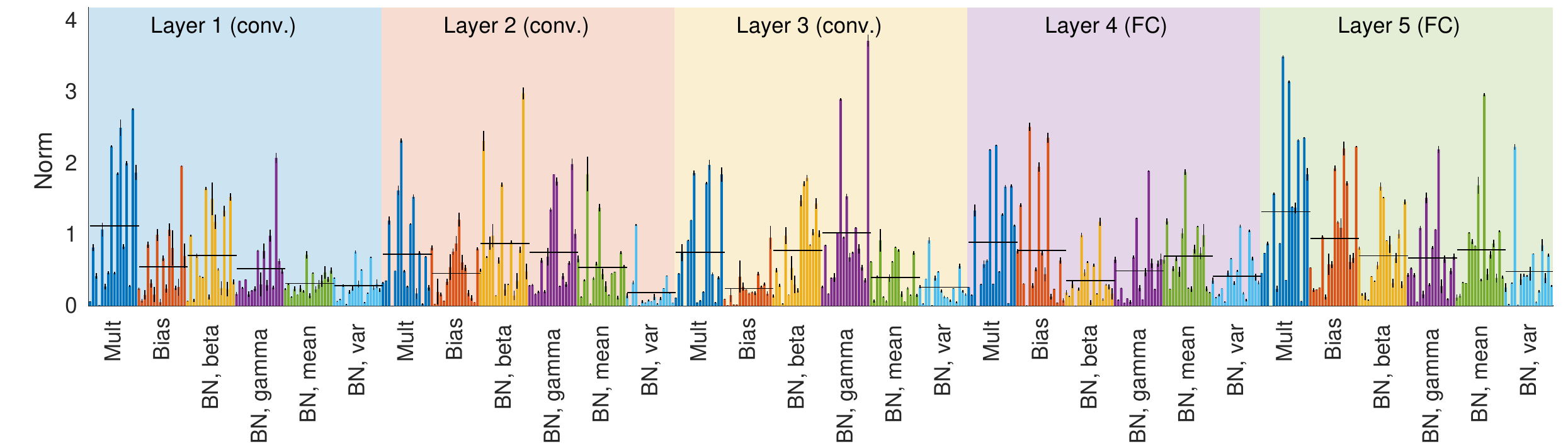}
		\caption{Initialization}
		\label{fig:coeff_global_lw_init}
	\end{subfigure}
	\caption{\label{fig:coeff_global_lw} Estimated impact of individual features of linear SVMs for hyper-parameter classification on the set $C_{fixed}$ of the NWS dataset. Each of the different types of weights of a layer, on the x-axis, are described by 16 features. Horizontal bars denote mean over a type of weights. Error bars show standard deviations over 10 separate training runs.}
\end{figure*}

In \figref{coeff_global_lw}, the coefficient norm of each of the in total 480 features of the layer-wise linear SVMs are illustrated, together with information on which layer and type of weight a feature is computed from. For dataset, optimizer and activation, there are distinct features which are more pronounced. For example, filter gradients of the first layer are most descriptive for dataset, biases from all convolutional layers are used for classifying optimizer, and running mean and variance are used for classifying activation function. For initialization classification, however, there are no clear individual features or layers which are most descriptive.

\begin{figure*}[t!]
	\centering
	\begin{subfigure}{0.325\textwidth}
		\includegraphics[width=\textwidth, trim={0pt 2pt 0pt 2pt}, clip]{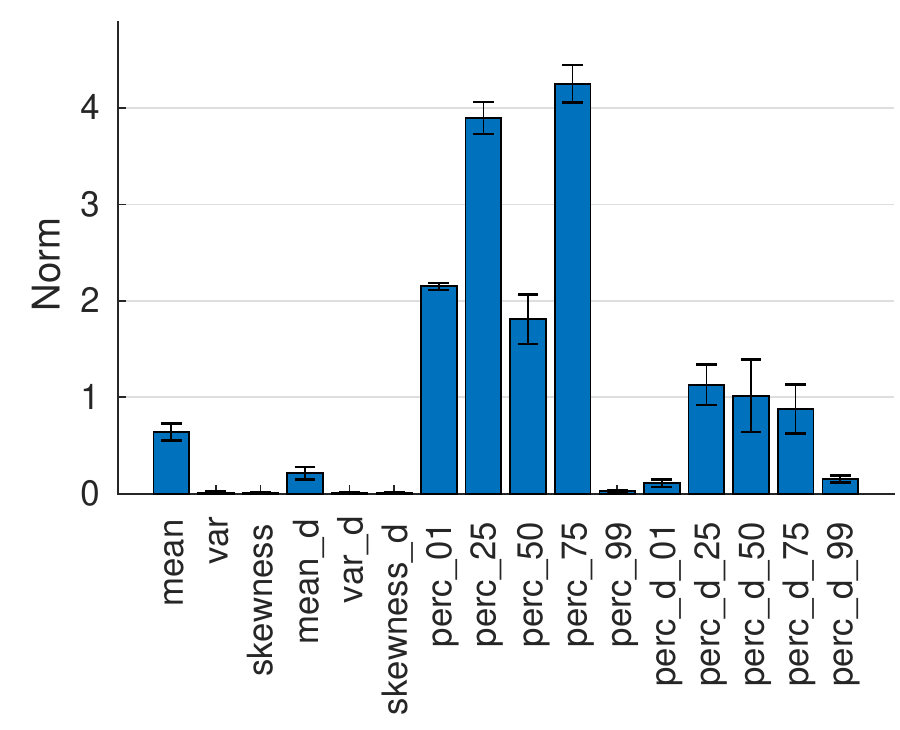}
		\caption{Dataset}
		\label{fig:coeff_local_dataset}
	\end{subfigure}
	\begin{subfigure}{0.325\textwidth}
		\includegraphics[width=\textwidth, trim={0pt 2pt 0pt 2pt}, clip]{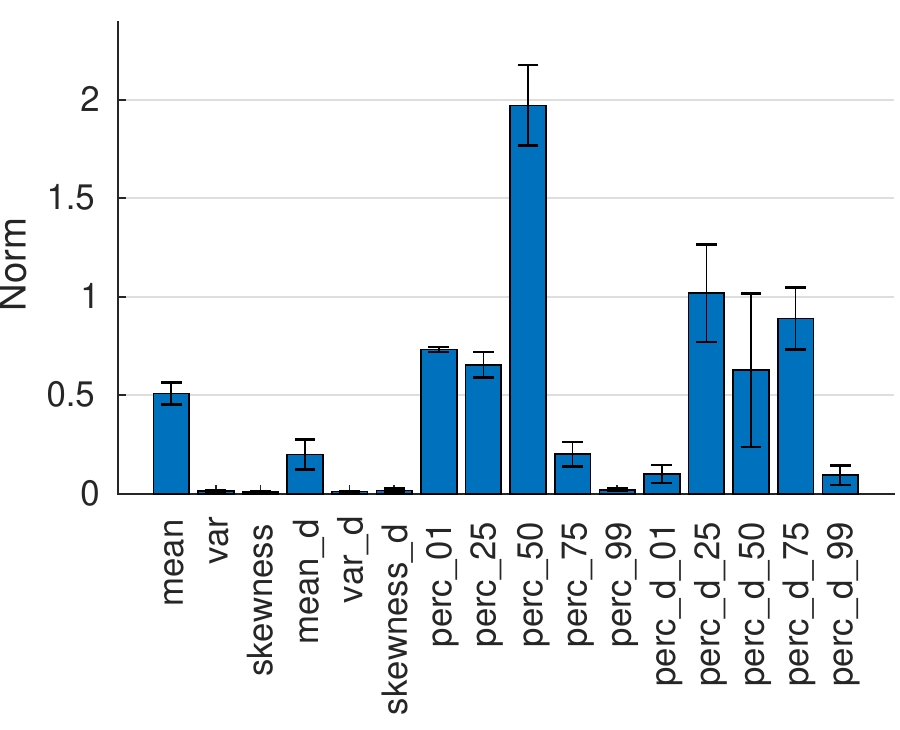}
		\caption{Batch size}
		\label{fig:coeff_local_batch_size}
	\end{subfigure}
	\begin{subfigure}{0.325\textwidth}
		\includegraphics[width=\textwidth, trim={0pt 2pt 0pt 2pt}, clip]{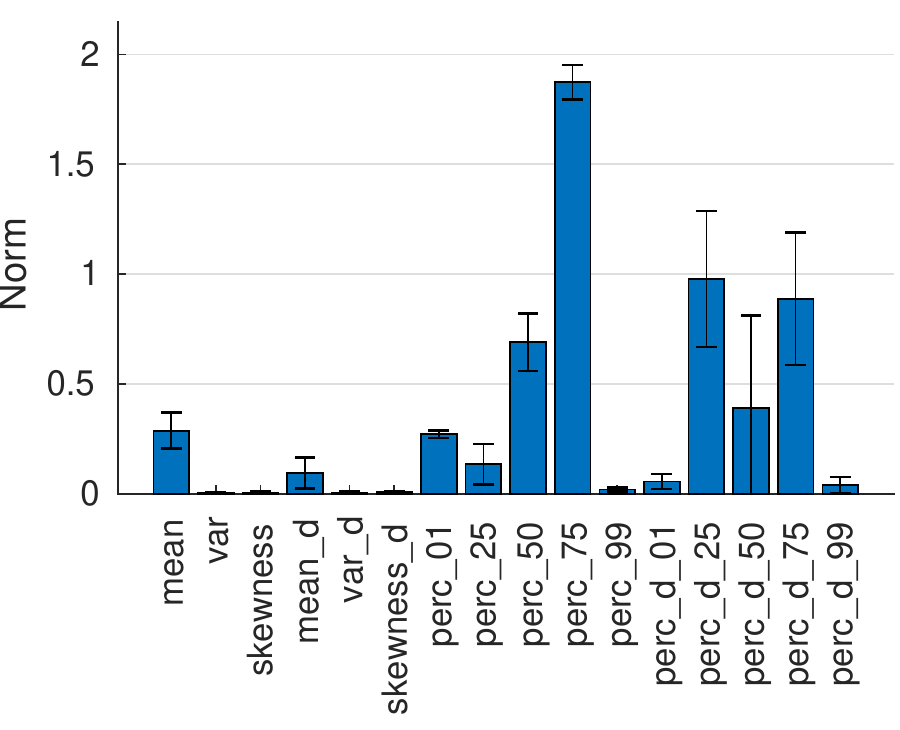}
		\caption{Augmentation}
		\label{fig:coeff_local_augmentation}
	\end{subfigure}\\
	\vspace{0.2cm}
	\begin{subfigure}{0.325\textwidth}
		\includegraphics[width=\textwidth, trim={0pt 2pt 0pt 2pt}, clip]{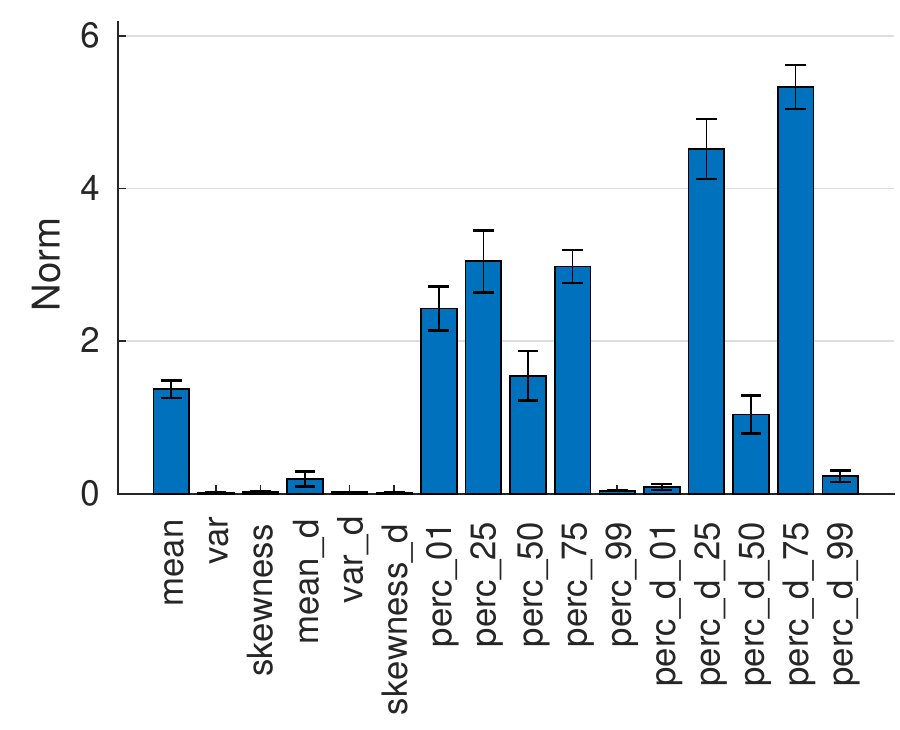}
		\caption{Optimizer}
		\label{fig:coeff_local_opt}
	\end{subfigure}
	\begin{subfigure}{0.325\textwidth}
		\includegraphics[width=\textwidth, trim={0pt 2pt 0pt 2pt}, clip]{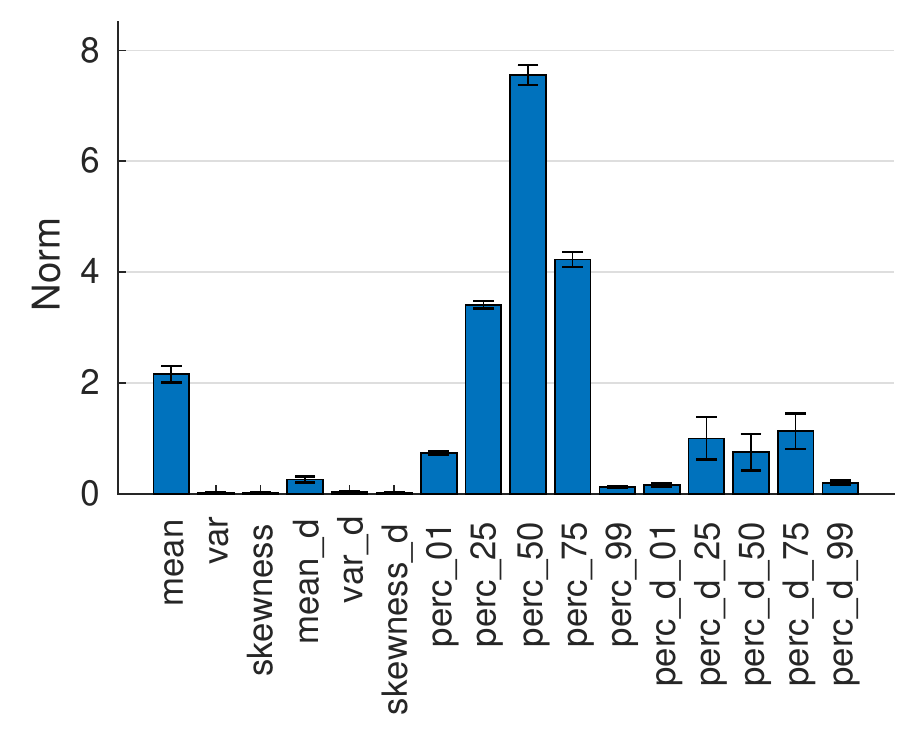}
		\caption{Activation}
		\label{fig:coeff_local_activation}
	\end{subfigure}
	\begin{subfigure}{0.325\textwidth}
		\includegraphics[width=\textwidth, trim={0pt 2pt 0pt 2pt}, clip]{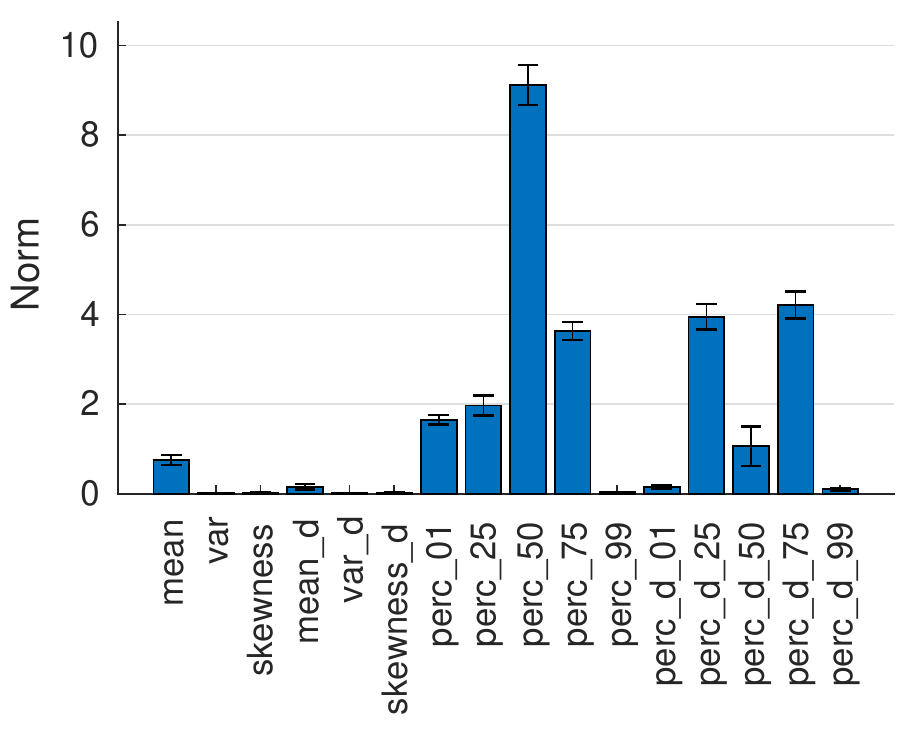}
		\caption{Initialization}
		\label{fig:coeff_local_init}
	\end{subfigure}
	\caption{\label{fig:coeff_local} Estimated impact of individual features of linear SVMs for hyper-parameter classification on the set $C_{main}$ of the NWS dataset. The features described using ``\_d'' are extracted from weight gradients $\nabla\theta$. Error bars show standard deviations over 10 separate training runs.}
\end{figure*}

\figref{coeff_local} shows coefficient norms of the 16 features of SVMs trained on local subsets of weights. The information used for classification is predominantly located in the features containing percentiles of weights and weight gradients.

\section{DMC trainings}	
\subsection{Experimental setup}
\begin{table*}
	\centering
	\def\arraystretch{1.05}
	\small
	\caption{DMC 1D CNN architectures. The arrows denote the number of channels from one layer to the next. Before the first FC layer, the output of the conv. layer is reshaped to a 1D vector. All layers except for the last are followed by batch normalization and ReLU activation.}
	\vspace{0.2cm}
	\begin{tabular}{r | l | l }
		\textbf{Layer} &\textbf{Global DMC} (10,948,997 weights in total) & \textbf{Local DMC} (10,407,685 weights in total)\\
		\hline
		\textbf{Input} & 92,868$\times$1 & 5,000$\times$1\\
		\textbf{1} & Conv. 5$\times$1: (1 $\rightarrow$ 8) $+$ Max-pooling 2$\times$1 & Conv. 5$\times$1: (1 $\rightarrow$ 8) $+$ Max-pooling 2$\times$1\\
		\textbf{2} & Conv. 5$\times$1 (8 $\rightarrow$ 16)  $+$ Max-pooling 2$\times$1 & Conv. 5$\times$1 (8 $\rightarrow$ 16)  $+$ Max-pooling 2$\times$1\\
		\textbf{3} & Conv. 5$\times$1 (16 $\rightarrow$ 32)  $+$ Max-pooling 2$\times$1 & Conv. 5$\times$1 (16 $\rightarrow$ 32)\\
		\textbf{4} & Conv. 5$\times$1 (32 $\rightarrow$ 64)  $+$ Max-pooling 2$\times$1 & Conv. 5$\times$1 (32 $\rightarrow$ 64)  $+$ Max-pooling 2$\times$1\\
		\textbf{5} & Conv. 5$\times$1 (64 $\rightarrow$ 128)  $+$ Max-pooling 2$\times$1 & Conv. 5$\times$1 (64 $\rightarrow$ 128)\\
		\textbf{6} & Conv. 5$\times$1 (128 $\rightarrow$ 128)  $+$ Max-pooling 2$\times$1 & Conv. 5$\times$1 (128 $\rightarrow$ 256)  $+$ Max-pooling 2$\times$1\\
		\textbf{7} & Conv. 5$\times$1 (128 $\rightarrow$ 128)  $+$ Max-pooling 2$\times$1 & Conv. 5$\times$1 (256 $\rightarrow$ 256)  $+$ Max-pooling 2$\times$1\\
		\textbf{8} & Conv. 5$\times$1 (128 $\rightarrow$ 128)  $+$ Max-pooling 2$\times$1 & Conv. 5$\times$1 (256 $\rightarrow$ 256)  $+$ Max-pooling 2$\times$1\\
		\textbf{9} & Conv. 5$\times$1 (128 $\rightarrow$ 128)  $+$ Max-pooling 2$\times$1 & Conv. 5$\times$1 (256 $\rightarrow$ 256)  $+$ Max-pooling 2$\times$1\\
		\textbf{10} & Conv. 5$\times$1 (128 $\rightarrow$ 128) & Conv. 5$\times$1 (256 $\rightarrow$ 256)\\
		\textbf{11} & Conv. 5$\times$1 (128 $\rightarrow$ 256)  $+$ Max-pooling 2$\times$1 & Conv. 5$\times$1 (256 $\rightarrow$ 256)\\
		\textbf{12} & Conv. 5$\times$1 (256 $\rightarrow$ 256) & Conv. 5$\times$1 (256 $\rightarrow$ 256) $+$ Max-pooling 2$\times$1\\
		\textbf{13} & Conv. 5$\times$1 (256 $\rightarrow$ 256)  $+$ Max-pooling 2$\times$1 & FC (4864 $\rightarrow$ 1024) $+$ Dropout 0.5\\
		\textbf{14} & Conv. 5$\times$1 (256 $\rightarrow$ 256) & FC (1024 $\rightarrow$ 1024) $+$ Dropout 0.5\\
		\textbf{15} & Conv. 5$\times$1 (256 $\rightarrow$ 256)  $+$ Max-pooling 2$\times$1 & FC (1024 $\rightarrow$ 1024) $+$ Dropout 0.5\\
		\textbf{16} & FC (5632 $\rightarrow$ 1024) $+$ Dropout 0.5 & FC (1024 $\rightarrow$ 1024) $+$ Dropout 0.5\\
		\textbf{17} & FC (1024 $\rightarrow$ 1024) $+$ Dropout 0.5 & FC (1024 $\rightarrow$ 64) $+$ Dropout 0.5\\
		\textbf{18} & FC (1024 $\rightarrow$ 1024) $+$ Dropout 0.5 & FC (64 $\rightarrow$ C)\\
		\textbf{19} & FC (1024 $\rightarrow$ 1024) $+$ Dropout 0.5 & \\
		\textbf{20} & FC (1024 $\rightarrow$ 64) $+$ Dropout 0.5 & \\
		\textbf{21} & FC (64 $\rightarrow$ C)\\
		\label{tab:dmc}
	\end{tabular}
\end{table*}

The deep meta-classifiers (DMCs) are specified as 1D CNNs. A global DMC takes as input a significantly larger number of weights compared to a local DMC, and this is accounted for by having an appropriate number of pooling operations throughout the convolutional part of a network. The exact layer specifications of the global and local DMCs are listed in \tabref{dmc}. The CNNs use batch normalization and ReLU activation on each layer. Batch normalization turns out to be a crucial component to make the DMCs converge. Initialization is performed by the Glorot normal scheme \cite{Glorot2010}.

The choice of the number of layers and their width is made to get a reasonably large CNN that performs well. However, as we have not made an extensive effort to search for the optimal design, this could most probably be improved, and we leave this for future work.

For training of DMCs, we use the sub-sampled NWS (\secref{subsampling}). Since each model of the dataset has weight snapshots exported at 20 different occasions during training, we can also use more than one weight sample of the models. Although consecutive weight snapshots from the same training are expected to be similar, there are also differences that can improve DMC trainings, similar to how data augmentation is commonly used for improving generalization. For the global DMCs, we use the 4 last weight snapshots, for a total of 7,032 weight vectors used in training. For the local DMCs, we use the 2 last weight snapshots, for at total of 16,070 weight vectors.

When training local DMCs, for each new mini-batch we pick at random the location of the subset $\theta_{[a:b]}$ within the weight vector. This means that the effective size of the training set is much larger than the number of weight samples, and we can train for many epochs without over-fitting. Thus, we train local DMCs for 500 epochs, while global DMCs are trained for 100 epochs. 

Since the sub-sampling in \secref{subsampling} removes failed trainings, the distribution of hyper-parameters in the training set may be unbalanced (e.g. less trainings that use constant initialization, or sigmoid activation). We enforce class balance by simply only using $K$ samples of each class used for the meta-classification, where $K$ is the number of samples of the class with least samples.

Optimization of DMCs is performed using ADAM with default settings in Tensorflow ($\beta_1=0.9$, $\beta_2=0.999$, and $\epsilon=10^{-8}$), a batch size of 64, and a learning rate of $10^{-3}$. The learning rate is decayed by a factor $0.95$ 50 times during training, so that the end learning rate is $7.7 \times 10^{-5}$. A local DMC trained with a subset size of 5,000 takes approximately 2 hours to optimize on an Nvida GeForce GTX 1080 Ti GPU, while a global DMC takes $\sim$70 minutes.

\subsection{Subset size}
The choice of using a subset size of 5,000 weight elements as input to a local DMC is arbitrary, and means that on average 5\% of the weights of a model are used. In order to get a sense for the impact of the choice of subset size, we train DMCs using a range of different values, between 100 and 20,000. This means that the size of the DMC CNN in \tabref{dmc} will vary greatly. To account for this, we adjust the number of max-poolings performed in layers 6-11. That is, if the input is small, these layers are not followed by max-pooling, and if it is large all use max-pooling. Using this strategy, the CNNs will approximately use equally many trainable weights for different input sizes.

\begin{figure}[t!]
	\centering
	\includegraphics[width=0.9\linewidth, trim={0pt 2pt 0pt 2pt}, clip]{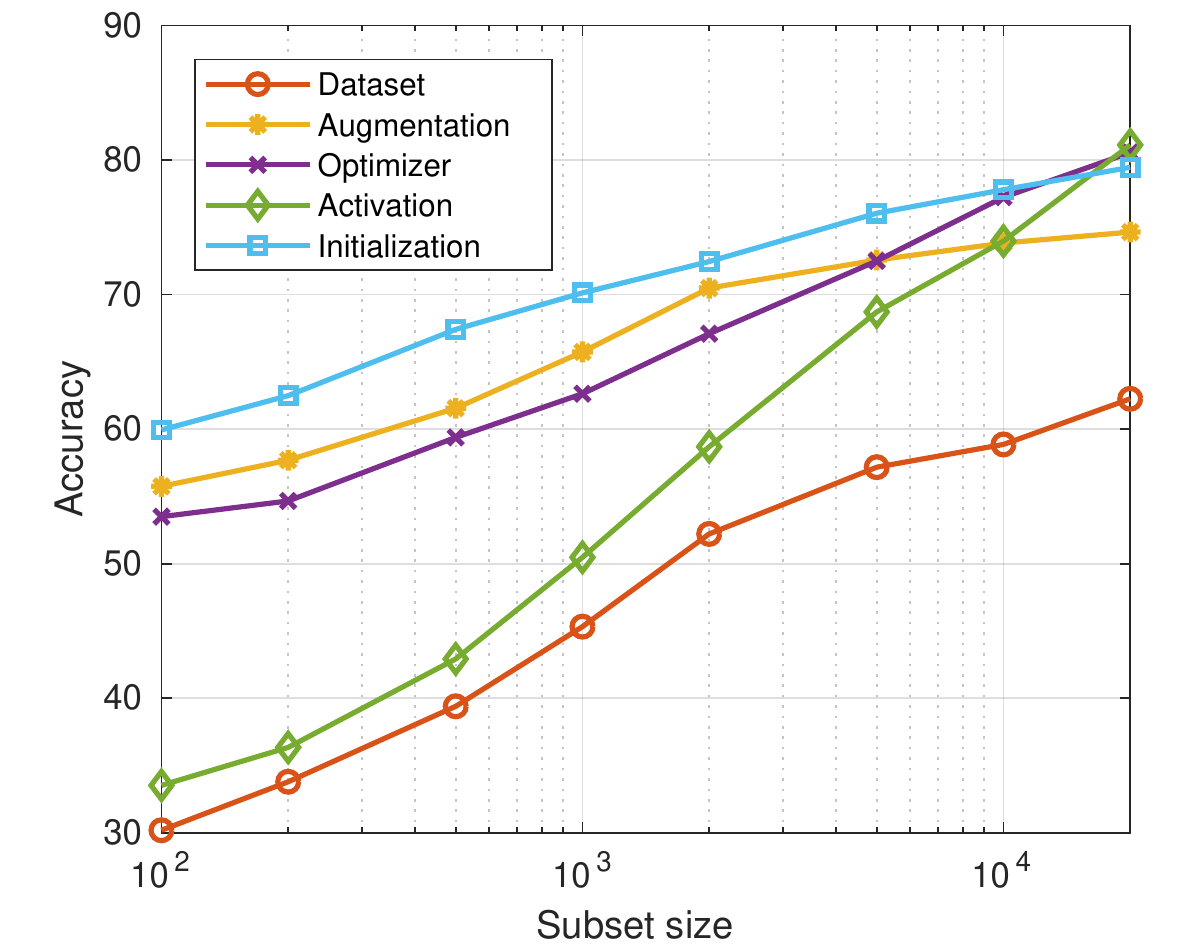}
	\caption{\label{fig:subset} Impact of subset size on DMC performance. Note that the x-axis is logarithmic.}
\end{figure}

\figref{subset} shows the relationship between the input size and DMC performance for 5 different hyper-parameter classifications. Generally, there is a logarithmic relation between subset size and DMC performance. Looking at the DMC trained to detect activation function, this shows a steeper slope, so that compared to e.g. the dataset DMC, activation function benefits more from having a larger number of weights from a model. This gives an indication on how local the information is stored, i.e. that activation function is a more global property of the weights than dataset.

\subsection{Batch normalization}
\begin{figure}[t!]
	\centering
	\includegraphics[width=\linewidth, trim={50pt 2pt 50pt 2pt}, clip]{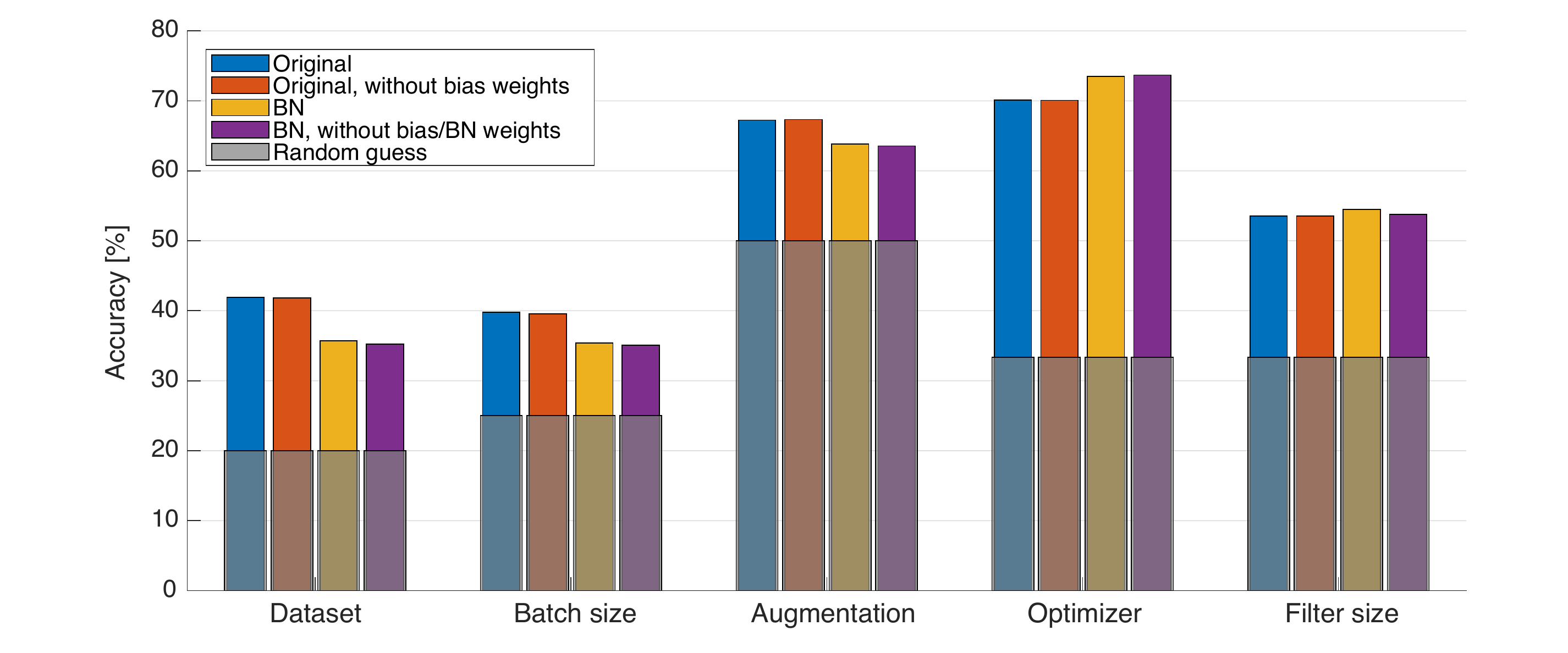}
	\caption{\label{fig:bn} The difference of DMC performance on models training with and without batch normalization (BN). The DMCs have been trained on a preliminary set of NWS samples. The performances have also been evaluated on weight vectors where BN and/or bias weights have been removed, in order to show that these do not contain the important part of the features learned by the DMCs.}
\end{figure}

As discussed in the main paper, and specified in Algorithm \ref{algo}, we use batch normalization (BN) for all CNN trainings, since otherwise the majority of trainings fail if difficult hyper-parameters are used (constant initialization, sigmoid activation, etc.). As discussed by Bau et al. \cite{Bau2017}, BN seems to have a whitening effect on the activations at each layer, which makes for decreased interpretability. The question is if the same thing is true also for the weights, and if this makes it more difficult to use a DMC to classify the weights? Potentially, the reduction in internal covariate shift provided by BN could make it more difficult to find descriptive features that describe the training properties.

In a preliminary experiment, shown in \figref{bn}, we compared DMCs trained on models with and without BN. Sampling of the NWS for this experiment was made using a less diverse sampling of hyper-parameters, with approximately 2K models trained using BN and 2K trained without BN. Performance for the BN-trained models is slightly lower for some of the hyper-parameters (dataset, batch size, augmentation), but also slightly higher for other hyper-parameters (optimizer, filter size). We conclude that there is a smaller difference using BN in the context of DMC training. However, it is still possible to see information about hyper-parameters, which is also confirmed by the DMC results we provide in the main paper.

Another aspect of BN is that it uses learnable parameters (running mean/variance and $\gamma$/$\beta$). These are part of the vectorized model weights $\theta$. Since the BN parameters carries statistics of the training, this raises the question of how much of the information found by a DMC is actually contained in the BN specific weights? And on the same line of questions; how much of the information is found in the bias weights as compared to other weights? \figref{coeff_global_lw} partially answers these questions for SVMs on the full set of weights. However, to also answer the questions for local DMCs, the results in \figref{bn} have also been computed on stripped weight vectors, where both BN specific and bias weights have been removed. That is, we use the same DMCs, trained on weights with BN/bias weights, but remove these weights during inference. As seen in the results, there are no significant differences in performance, which means that most information is contained in convolutional filters and FC weight matrices.

\subsection{DMC results}
\begin{figure*}[t!]
	\centering
	\begin{subfigure}{0.87\textwidth}
		\includegraphics[width=0.325\textwidth, trim={2pt 2pt 8pt 2pt}, clip]{plots/dmc_prop00_00.pdf}
		\includegraphics[width=0.325\textwidth, trim={2pt 2pt 8pt 2pt}, clip]{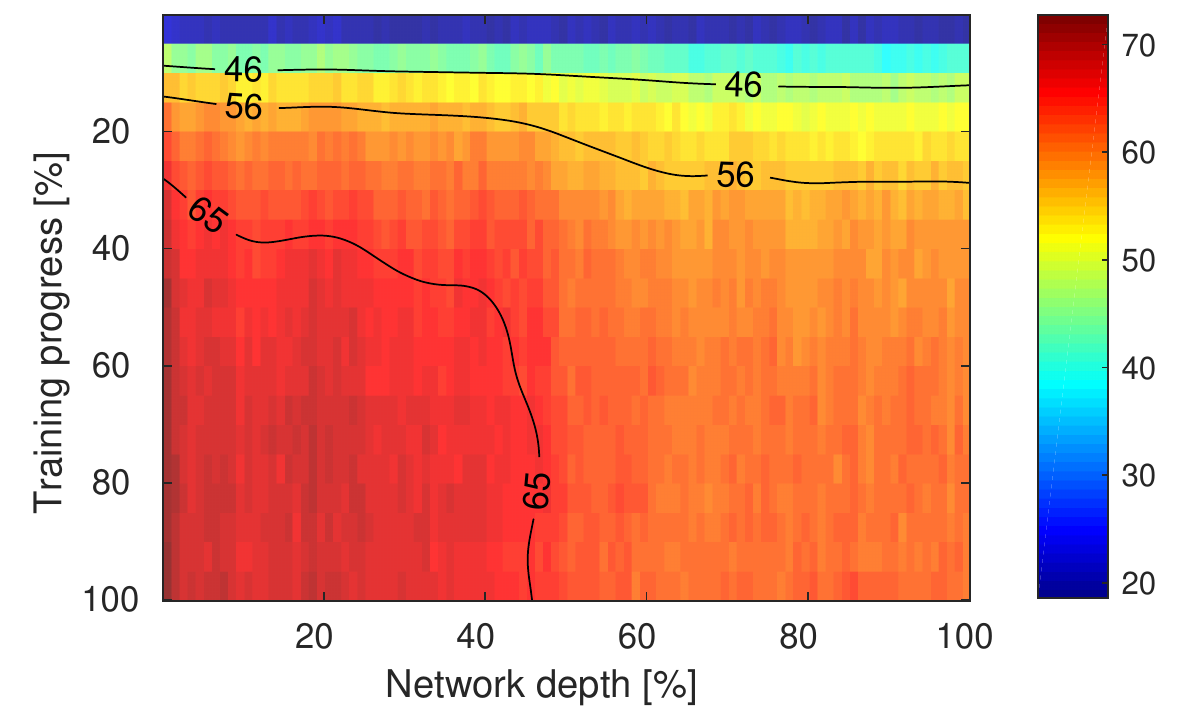}
		\includegraphics[width=0.325\textwidth, trim={2pt 2pt 8pt 2pt}, clip]{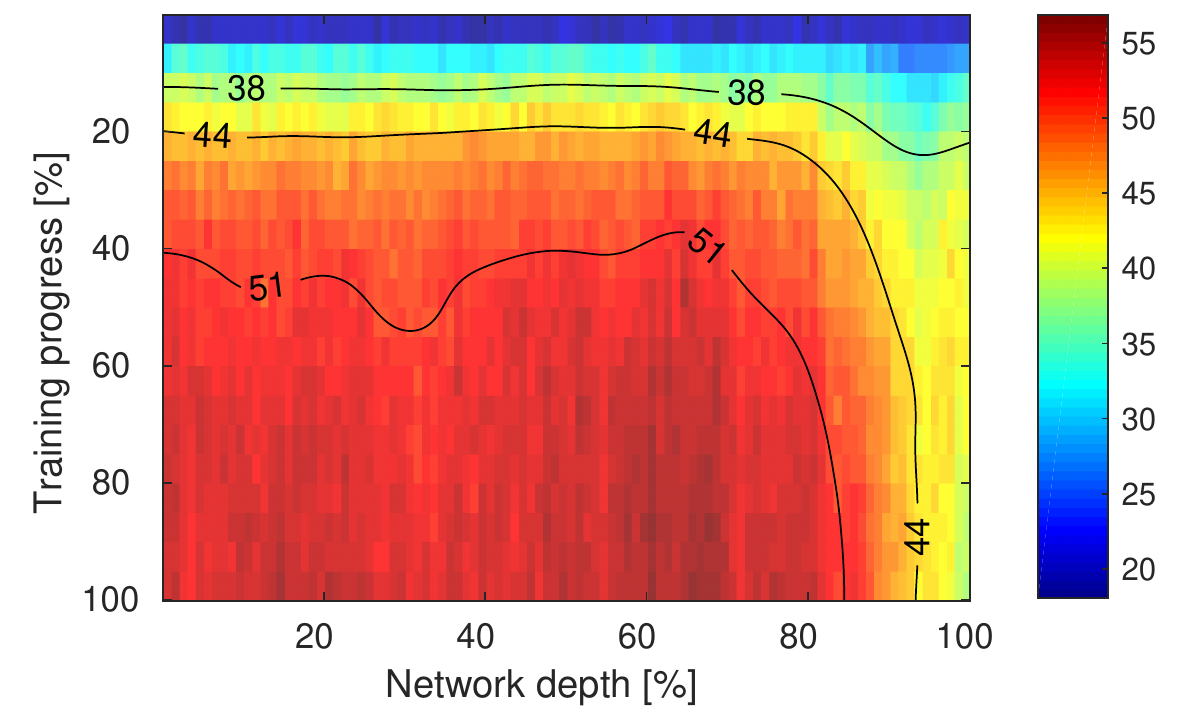}
		\caption{Dataset}
		\label{fig:dmc_dataset}
	\end{subfigure}\\
	\vspace{0.2cm}
	\begin{subfigure}{0.87\textwidth}
		\includegraphics[width=0.325\textwidth, trim={2pt 2pt 8pt 2pt}, clip]{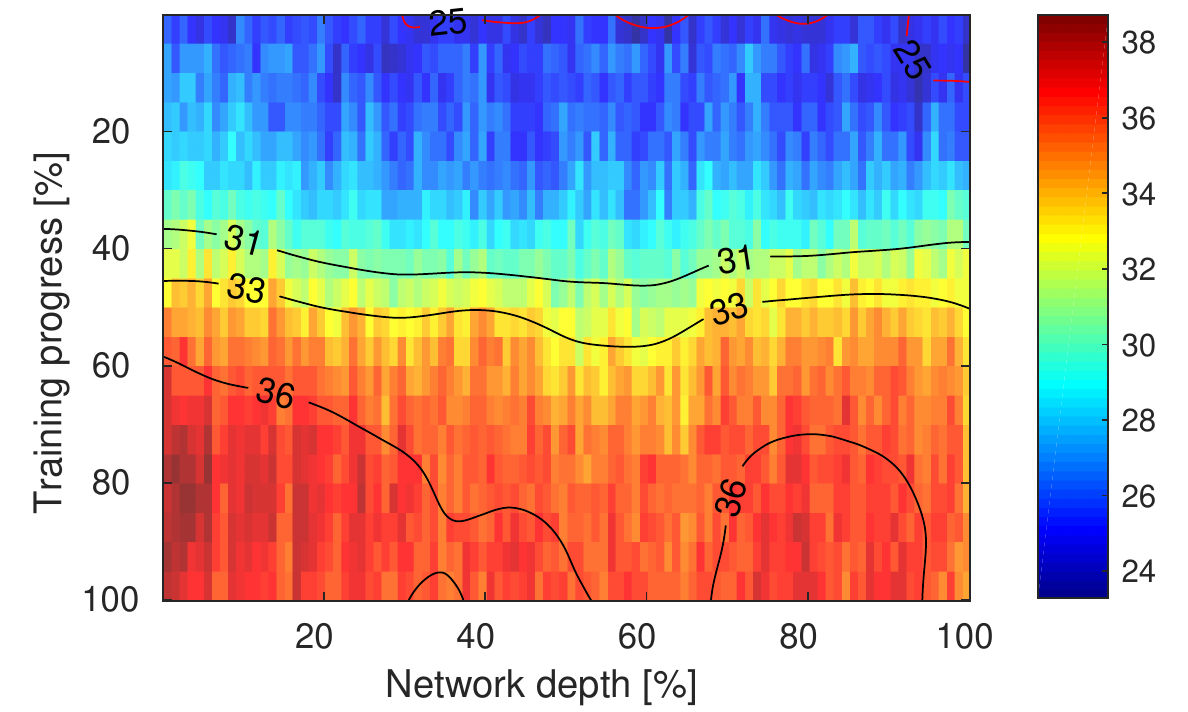}
		\includegraphics[width=0.325\textwidth, trim={2pt 2pt 8pt 2pt}, clip]{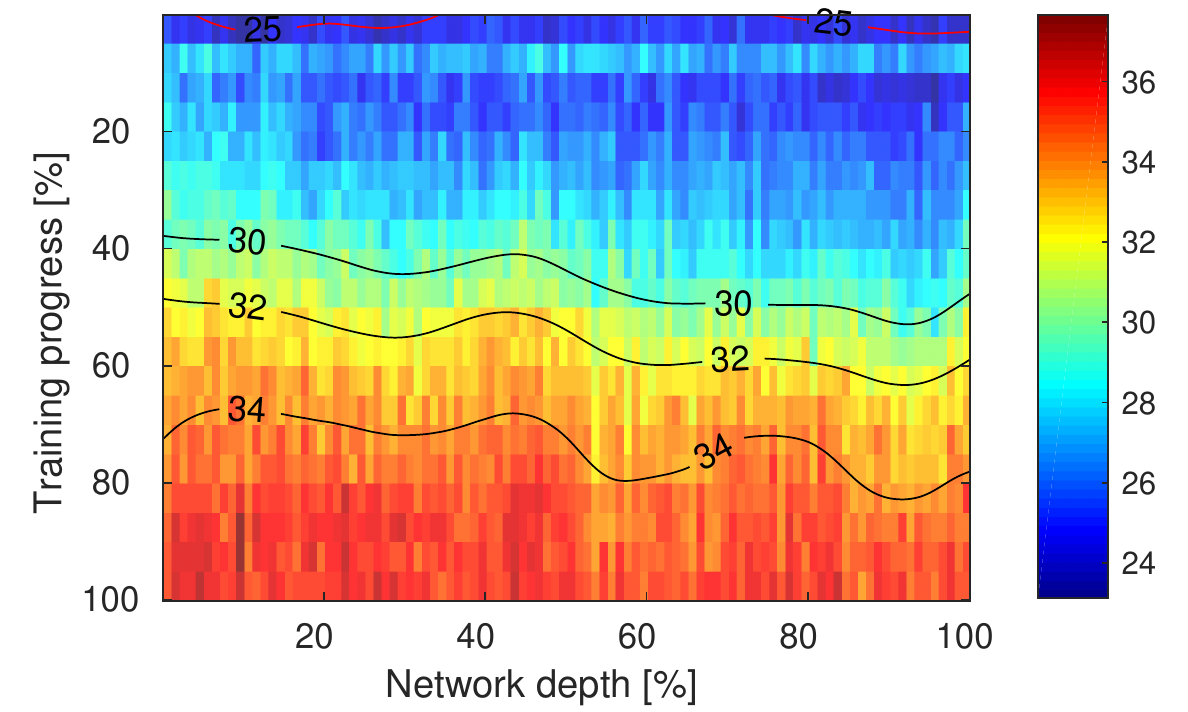}
		\includegraphics[width=0.325\textwidth, trim={2pt 2pt 8pt 2pt}, clip]{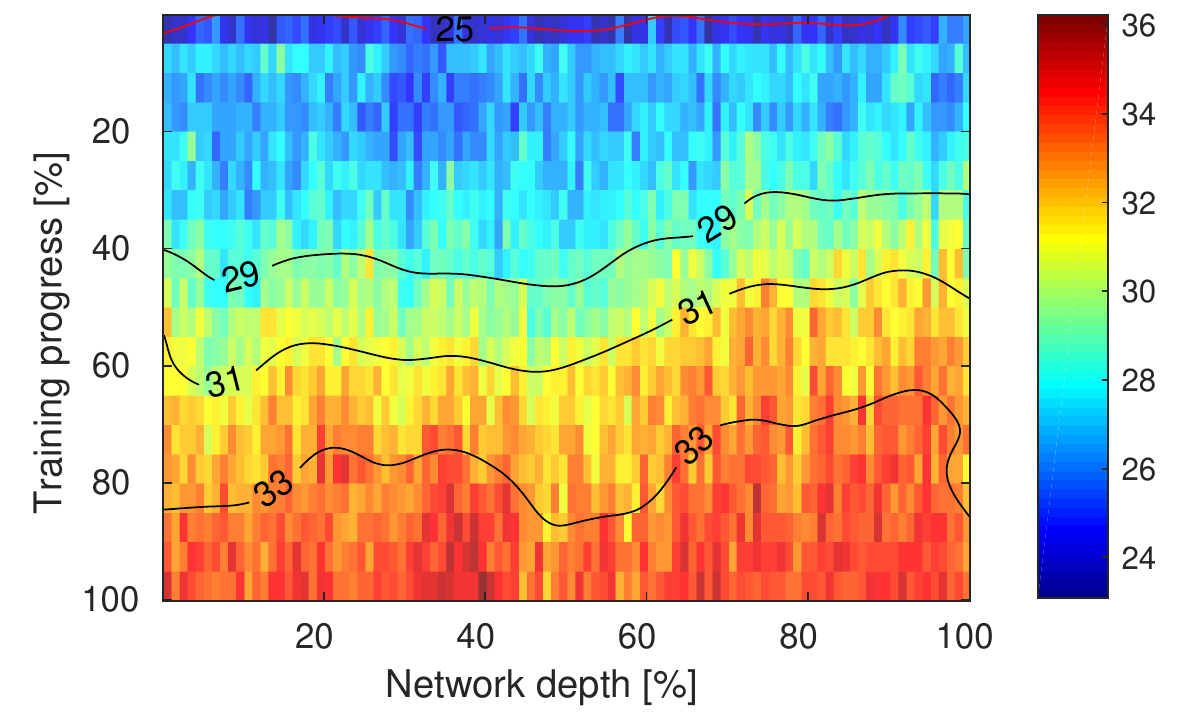}
		\caption{Batch size}
		\label{fig:dmc_bs}
	\end{subfigure}\\
	\vspace{0.2cm}
	\begin{subfigure}{0.87\textwidth}
		\includegraphics[width=0.325\textwidth, trim={2pt 2pt 8pt 2pt}, clip]{plots/dmc_prop04_00.pdf}
		\includegraphics[width=0.325\textwidth, trim={2pt 2pt 8pt 2pt}, clip]{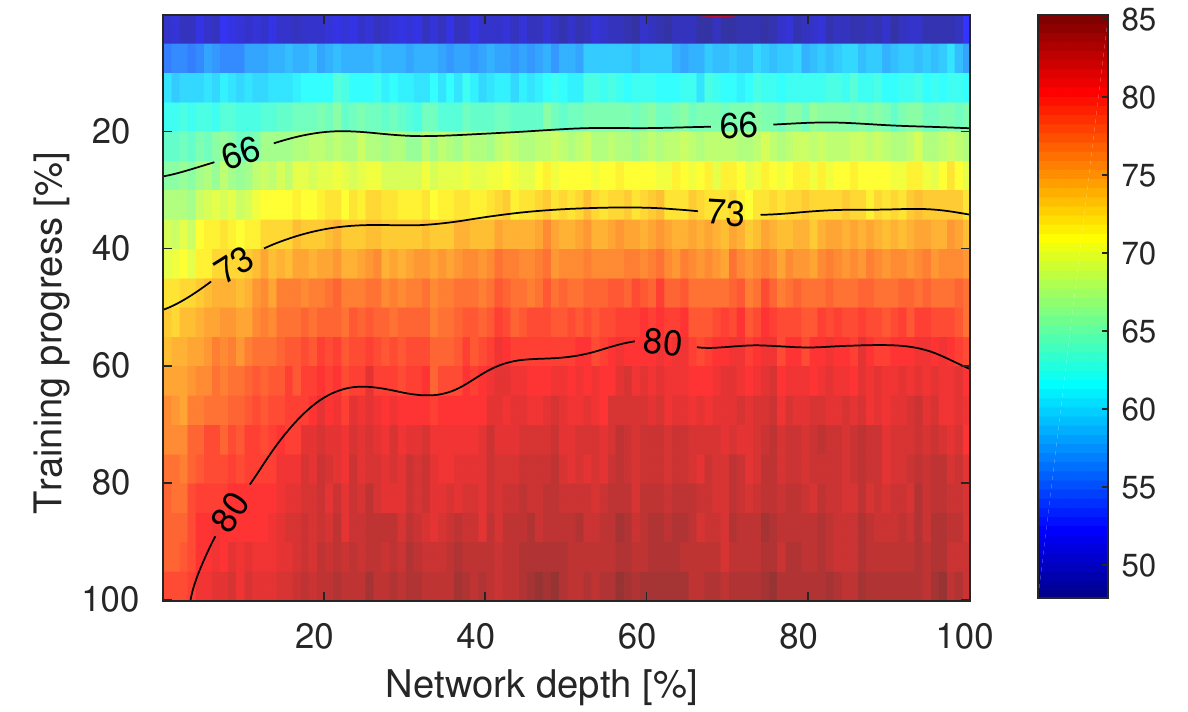}
		\includegraphics[width=0.325\textwidth, trim={2pt 2pt 8pt 2pt}, clip]{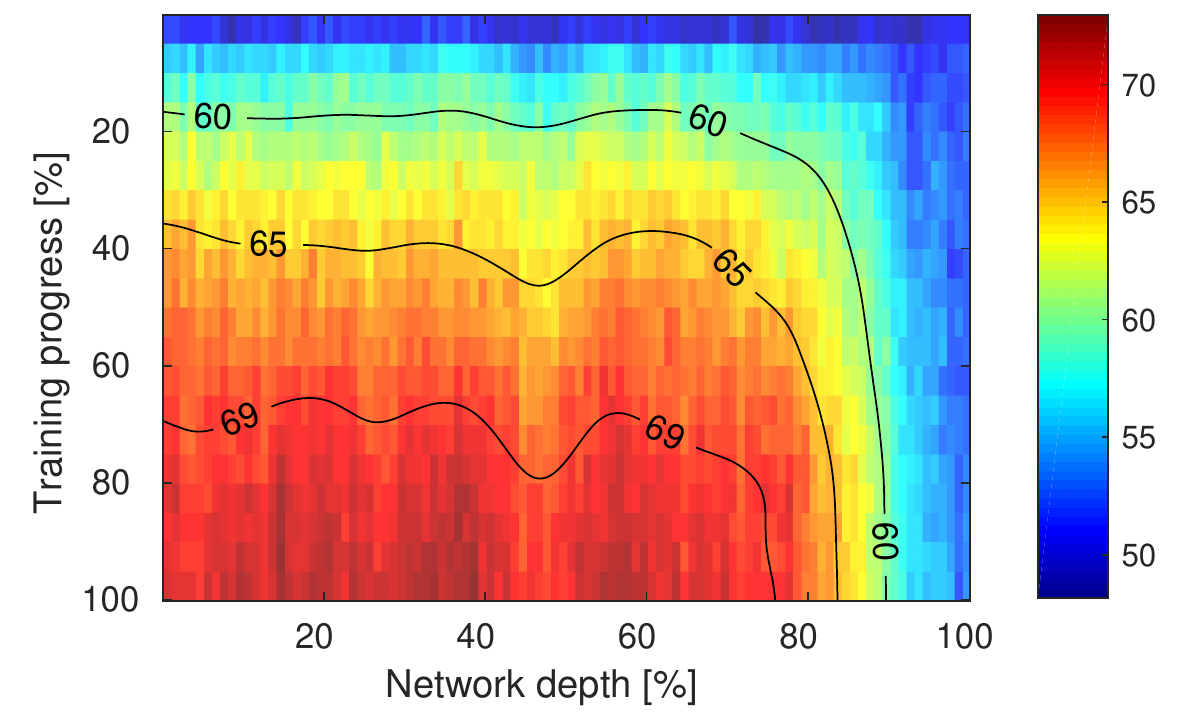}
		\caption{Augmentation}
		\label{fig:dmc_aug}
	\end{subfigure}\\
	\vspace{0.2cm}
	\begin{subfigure}{0.87\textwidth}
		\includegraphics[width=0.325\textwidth, trim={2pt 2pt 8pt 2pt}, clip]{plots/dmc_prop05_00.pdf}
		\includegraphics[width=0.325\textwidth, trim={2pt 2pt 8pt 2pt}, clip]{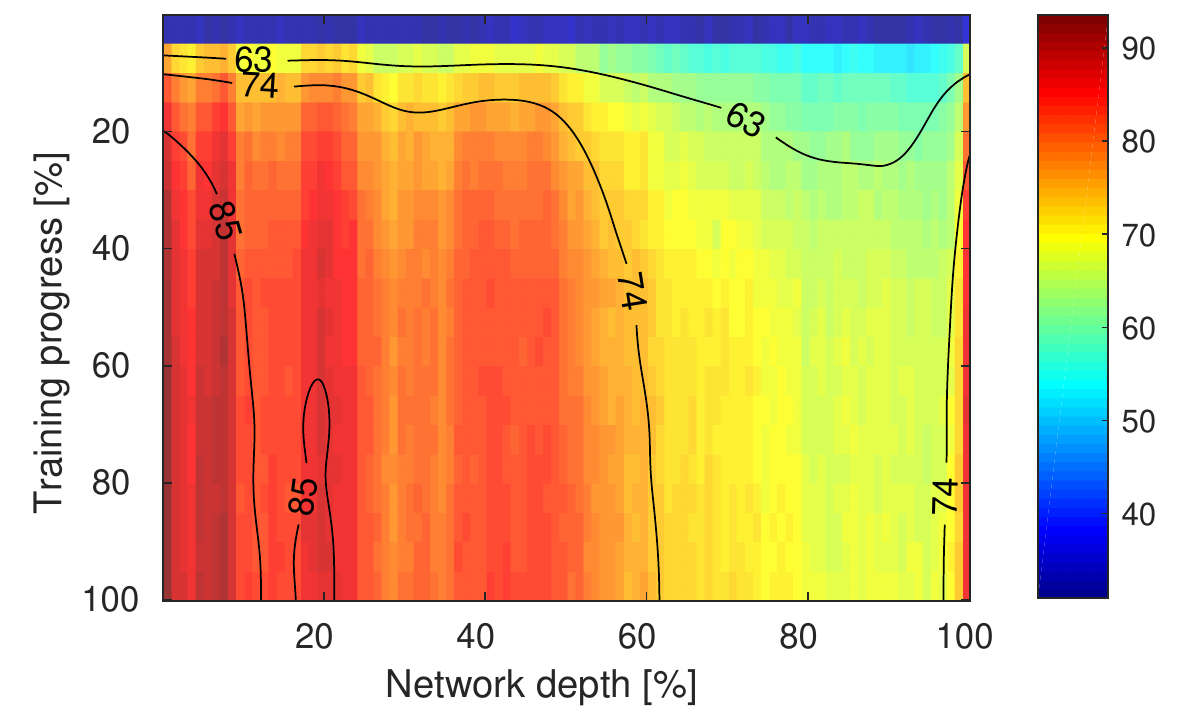}
		\includegraphics[width=0.325\textwidth, trim={2pt 2pt 8pt 2pt}, clip]{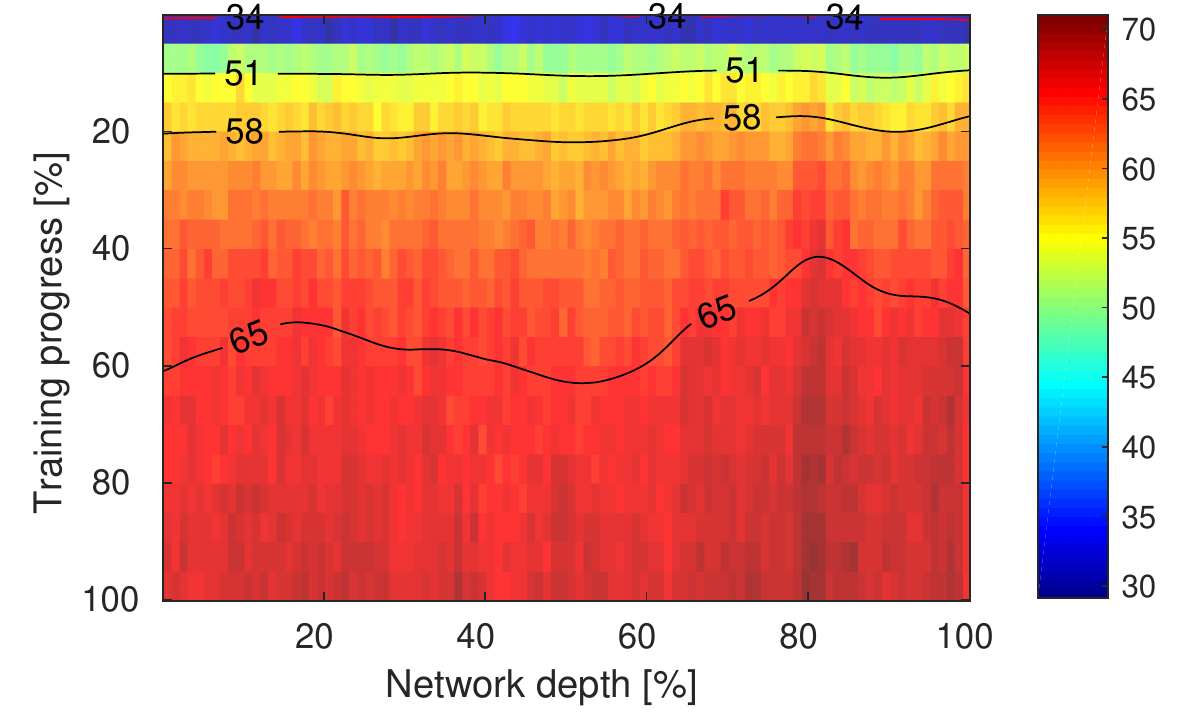}
		\caption{Optimizer}
		\label{fig:dmc_opt}
	\end{subfigure}\\
	\vspace{0.2cm}
	\begin{subfigure}{0.87\textwidth}
		\includegraphics[width=0.325\textwidth, trim={2pt 2pt 8pt 2pt}, clip]{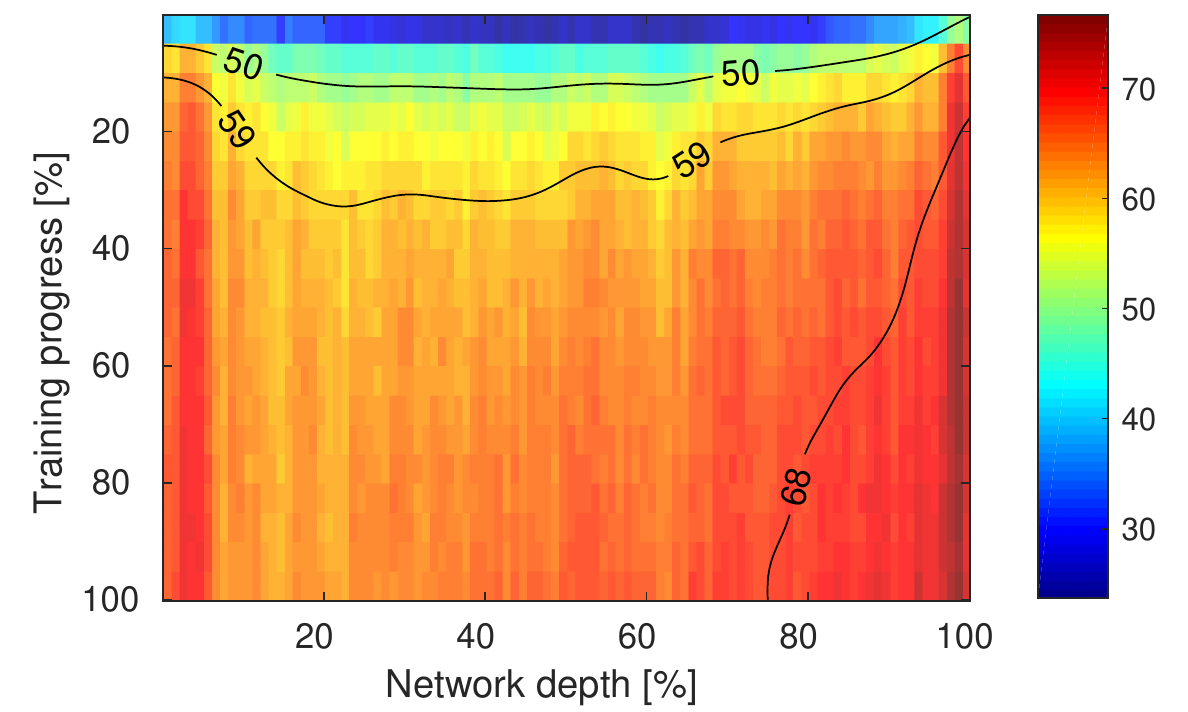}
		\includegraphics[width=0.325\textwidth, trim={2pt 2pt 8pt 2pt}, clip]{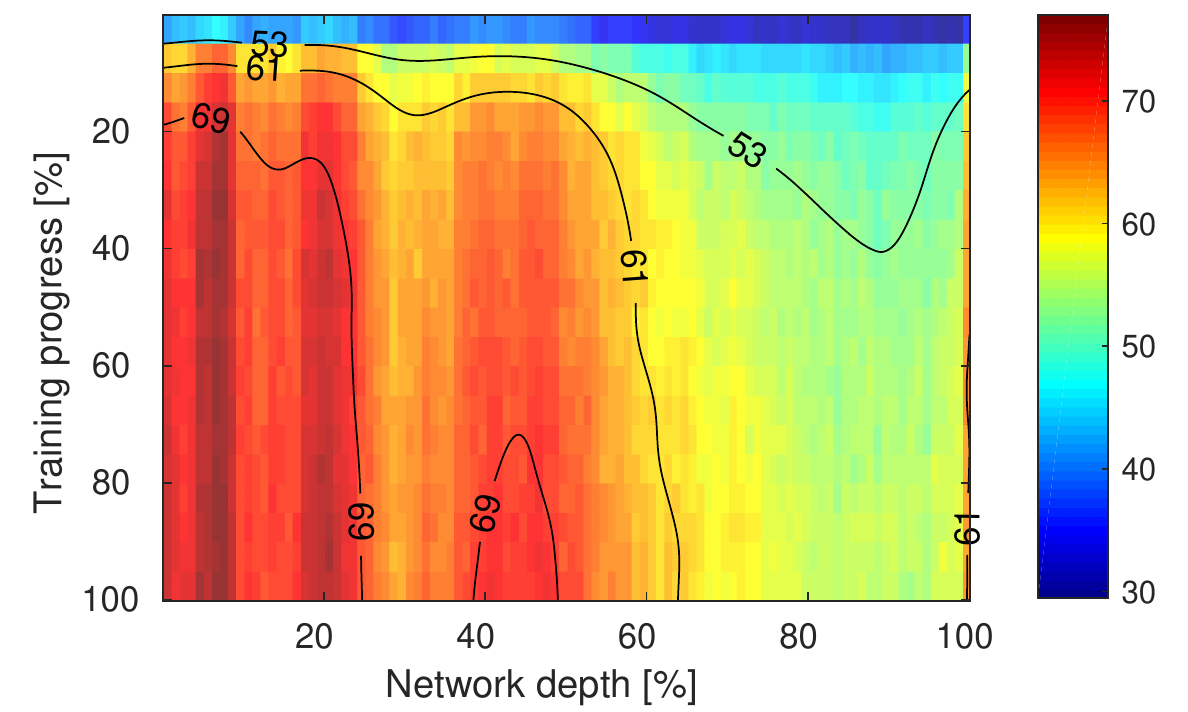}
		\includegraphics[width=0.325\textwidth, trim={2pt 2pt 8pt 2pt}, clip]{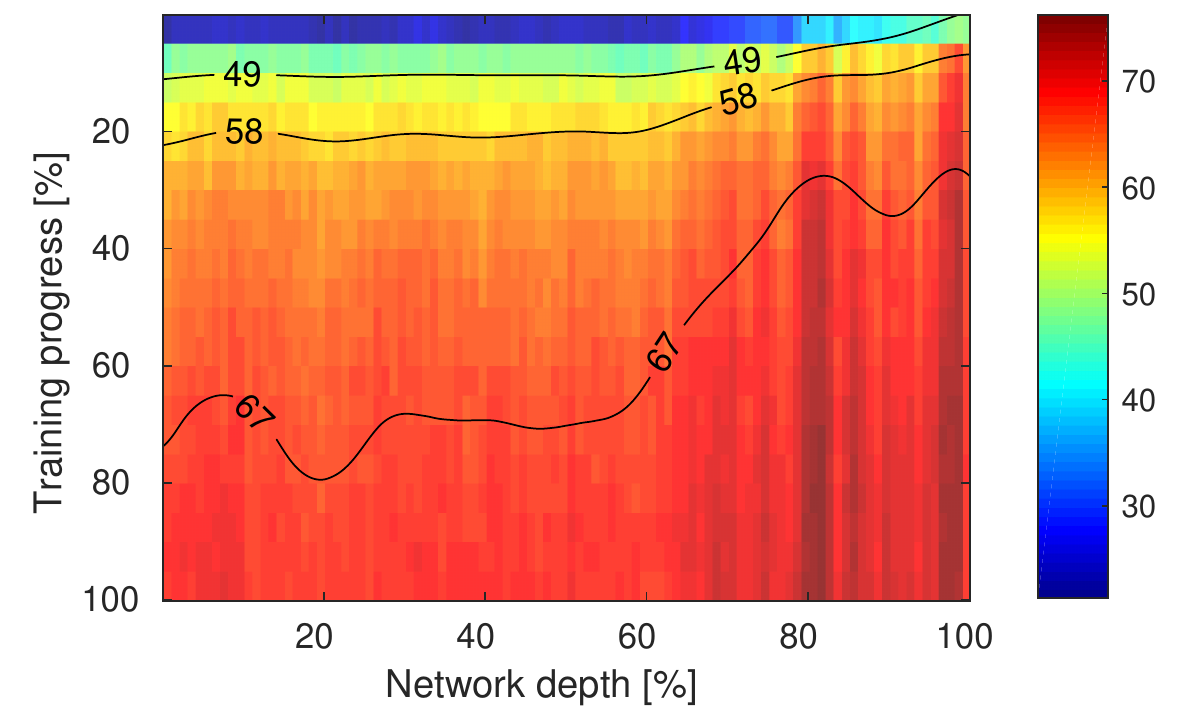}
		\caption{Activation function}
		\label{fig:dmc_act}
	\end{subfigure}\\
	\vspace{0.2cm}
	\begin{subfigure}{0.87\textwidth}
		\includegraphics[width=0.325\textwidth, trim={2pt 2pt 8pt 2pt}, clip]{plots/dmc_prop07_00.pdf}
		\includegraphics[width=0.325\textwidth, trim={2pt 2pt 8pt 2pt}, clip]{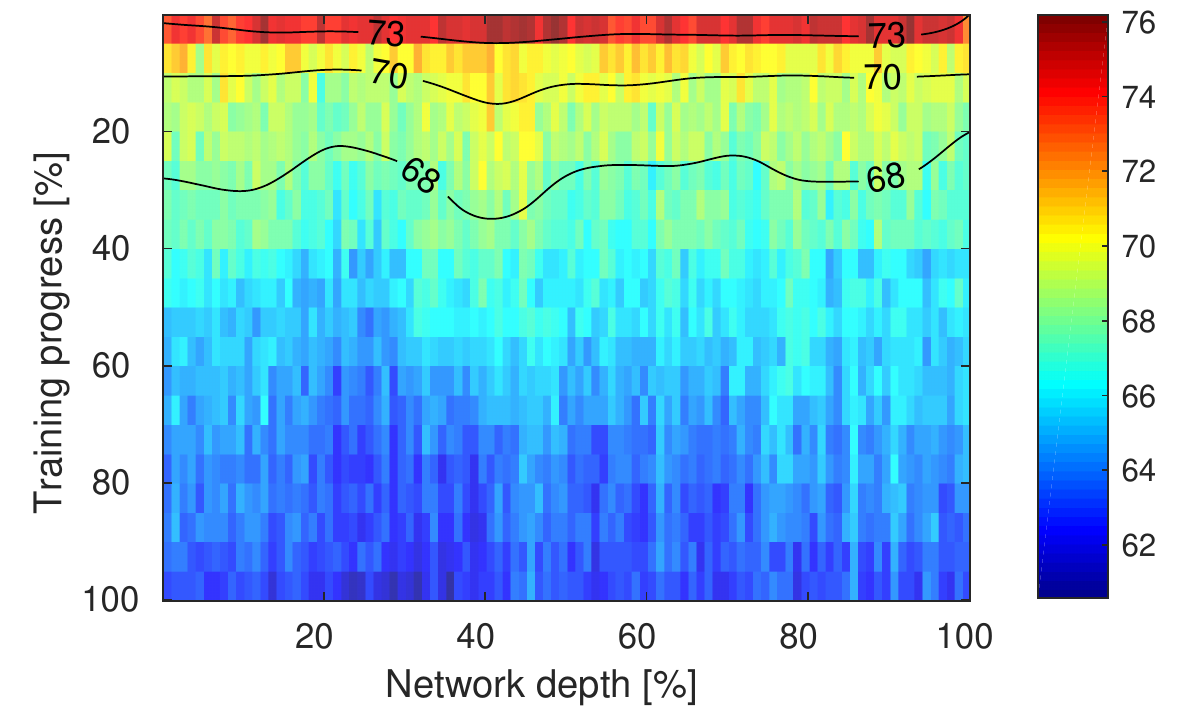}
		\includegraphics[width=0.325\textwidth, trim={2pt 2pt 8pt 2pt}, clip]{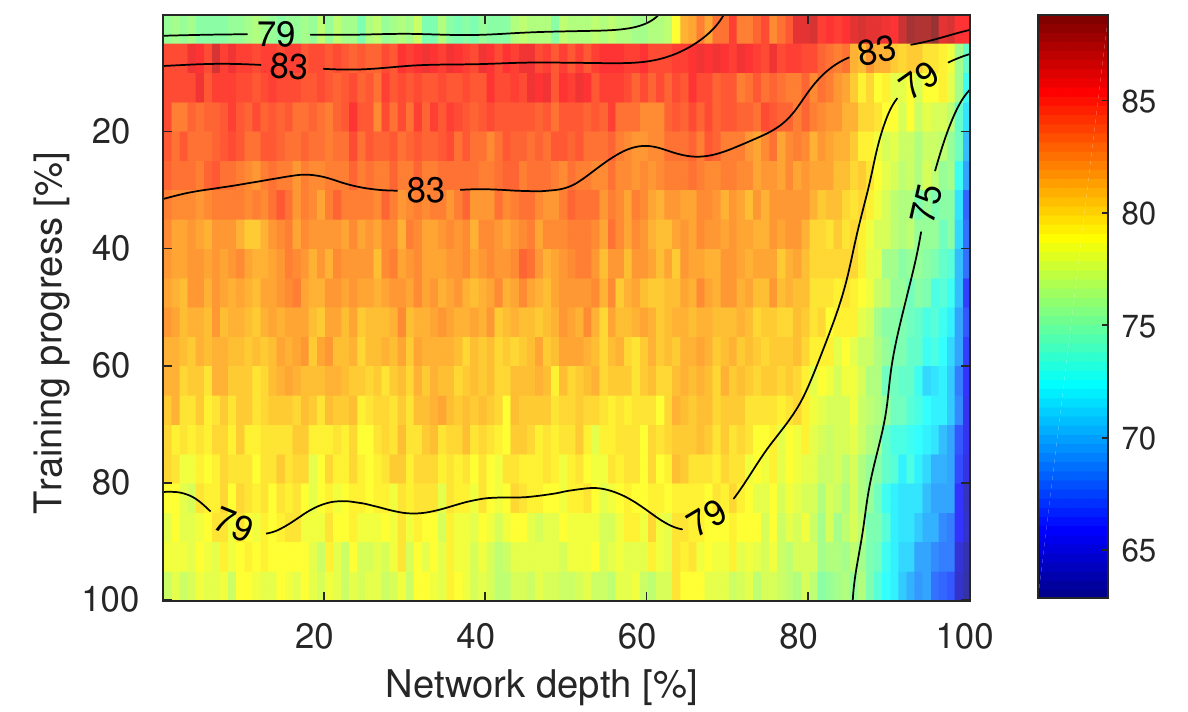}
		\caption{Initialization}
		\label{fig:dmc_init}
	\end{subfigure}
	\caption{\label{fig:dmc_1} DMC performance maps, demonstrating differences across training progress and network depth. The results in the left figures are from DMCs trained on subsets of all weights in a model, the middle uses only convolutional weights, and the right uses only FC weights. This means that the x-axis covers all weights, convolutional weights and FC weights, in the left, middle and right figures, respectively. Note that the ranges of the colormaps differ between plots.}
\end{figure*}

\begin{figure*}[t!]
	\centering
	\begin{subfigure}{0.87\textwidth}
		\includegraphics[width=0.325\textwidth, trim={2pt 2pt 8pt 2pt}, clip]{plots/dmc_prop08_00.pdf}
		\includegraphics[width=0.325\textwidth, trim={2pt 2pt 8pt 2pt}, clip]{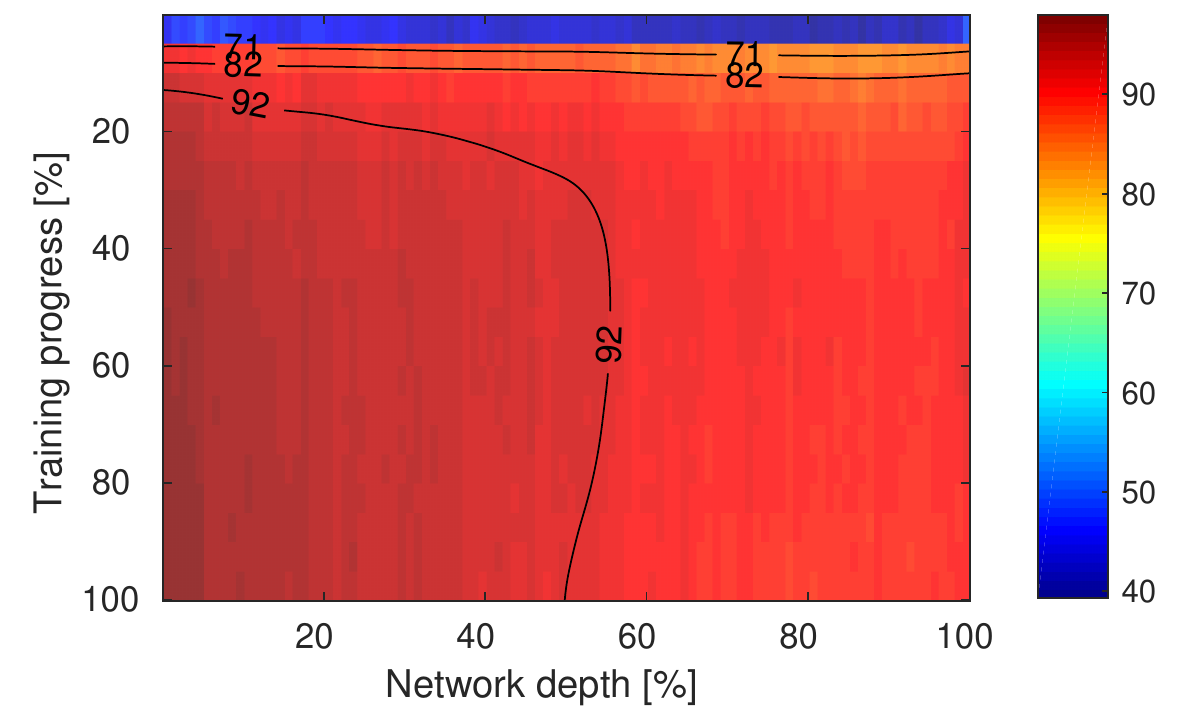}
		\includegraphics[width=0.325\textwidth, trim={2pt 2pt 8pt 2pt}, clip]{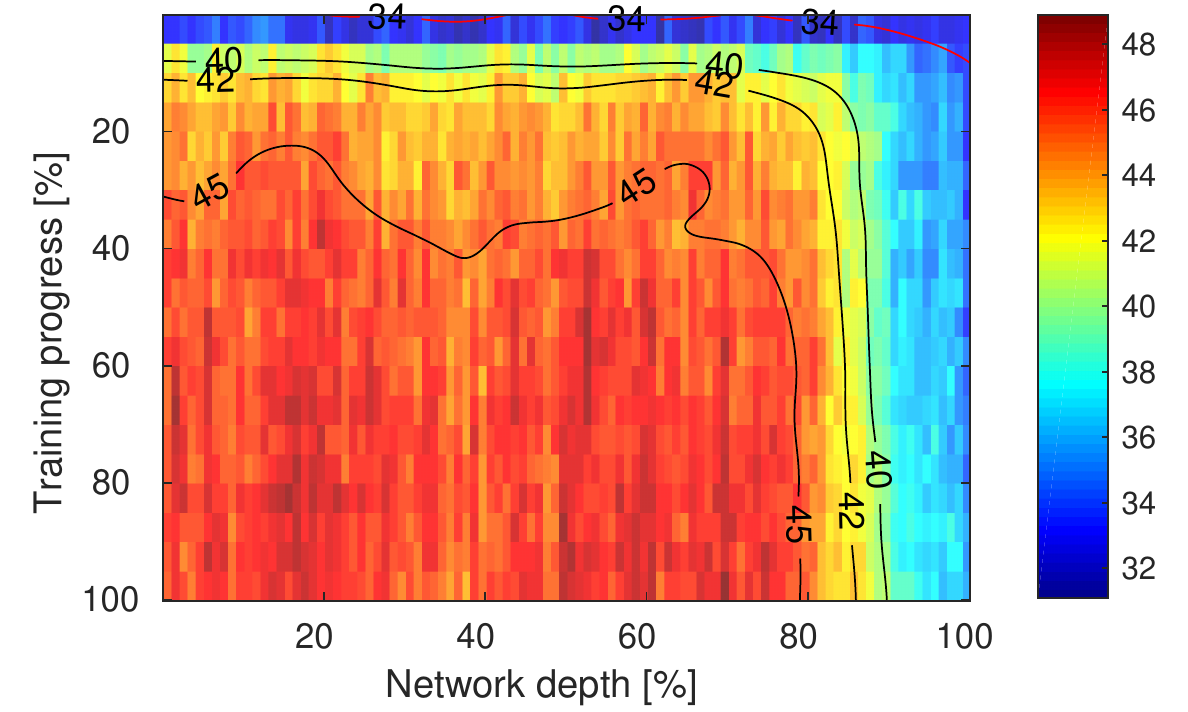}
		\caption{Filter size}
		\label{fig:dmc_filt}
	\end{subfigure}\\
	\vspace{0.2cm}
	\begin{subfigure}{0.87\textwidth}
		\includegraphics[width=0.325\textwidth, trim={2pt 2pt 8pt 2pt}, clip]{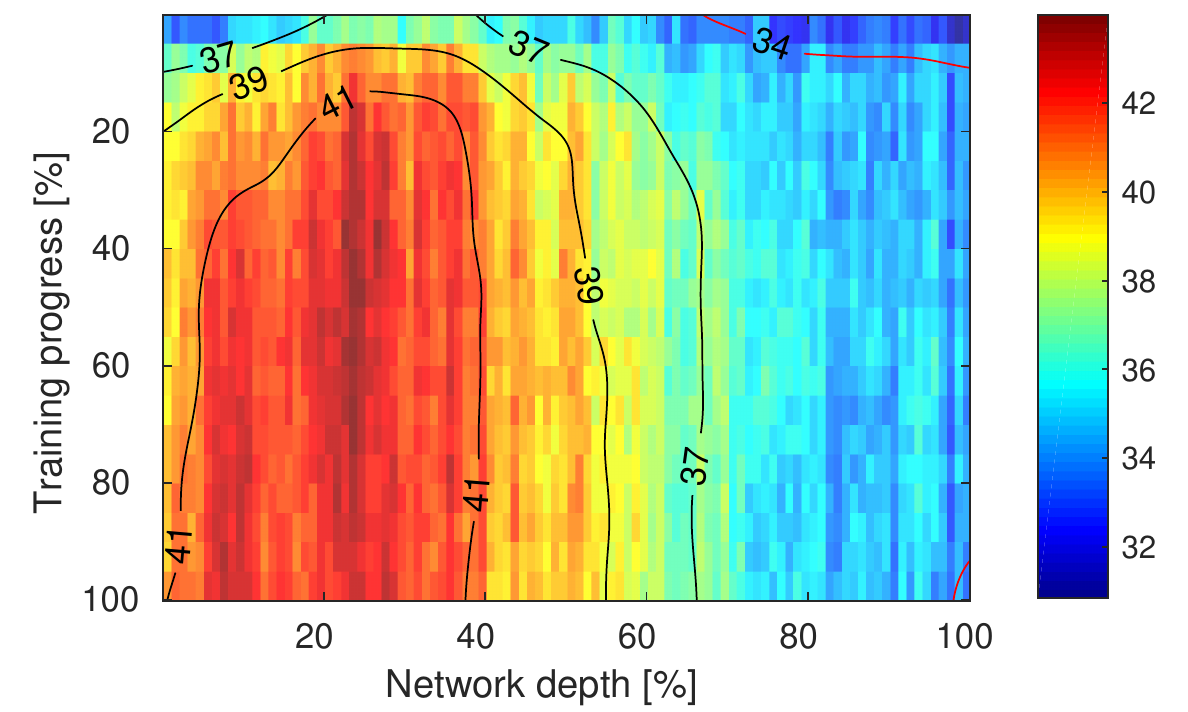}
		\includegraphics[width=0.325\textwidth, trim={2pt 2pt 8pt 2pt}, clip]{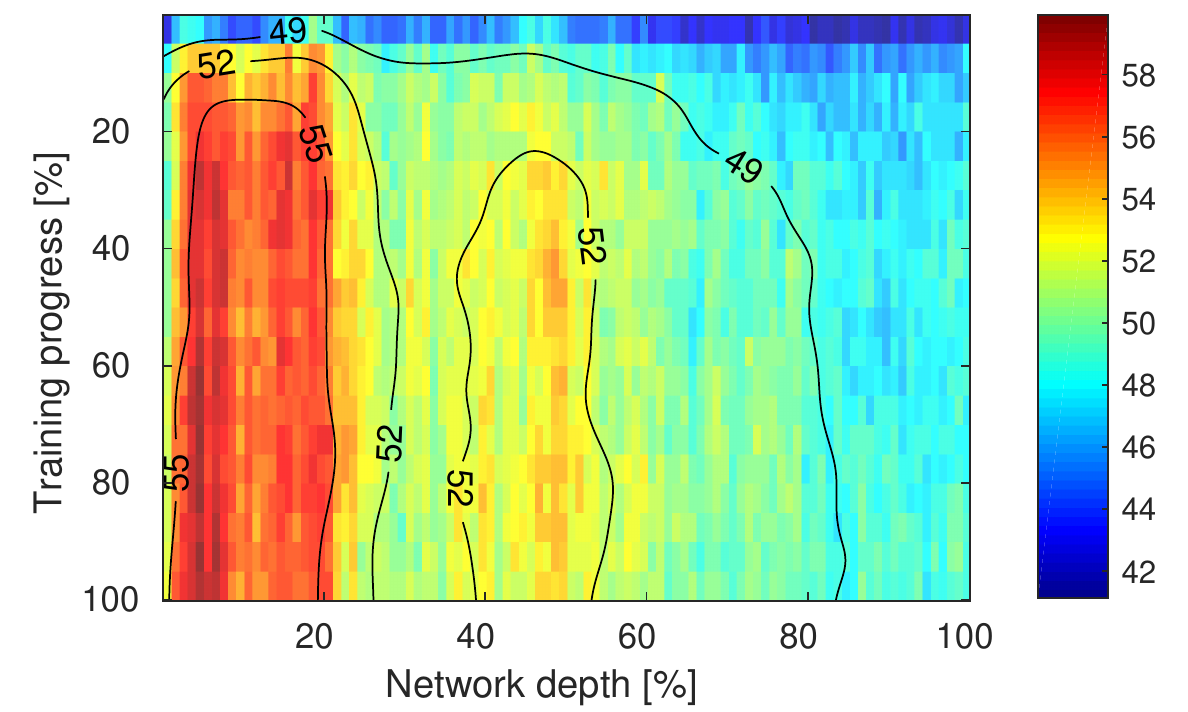}
		\includegraphics[width=0.325\textwidth, trim={2pt 2pt 8pt 2pt}, clip]{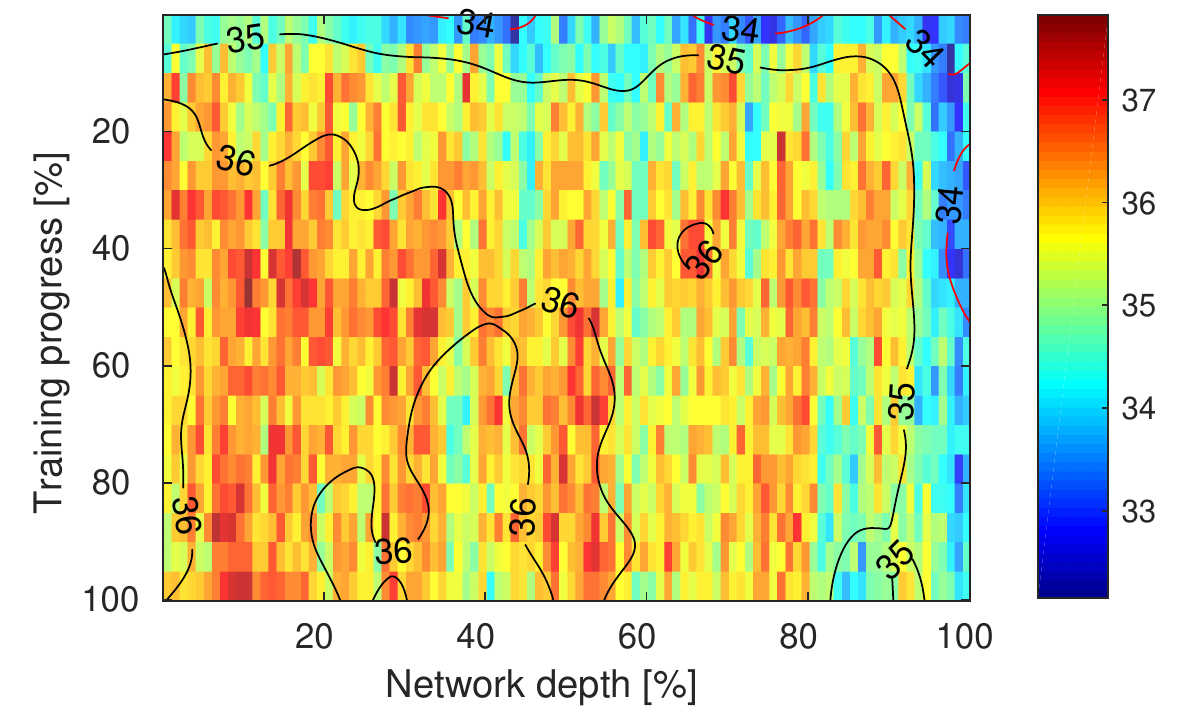}
		\caption{Depth conv.}
		\label{fig:dmc_depth_conv}
	\end{subfigure}
	\begin{subfigure}{0.87\textwidth}
		\includegraphics[width=0.325\textwidth, trim={2pt 2pt 8pt 2pt}, clip]{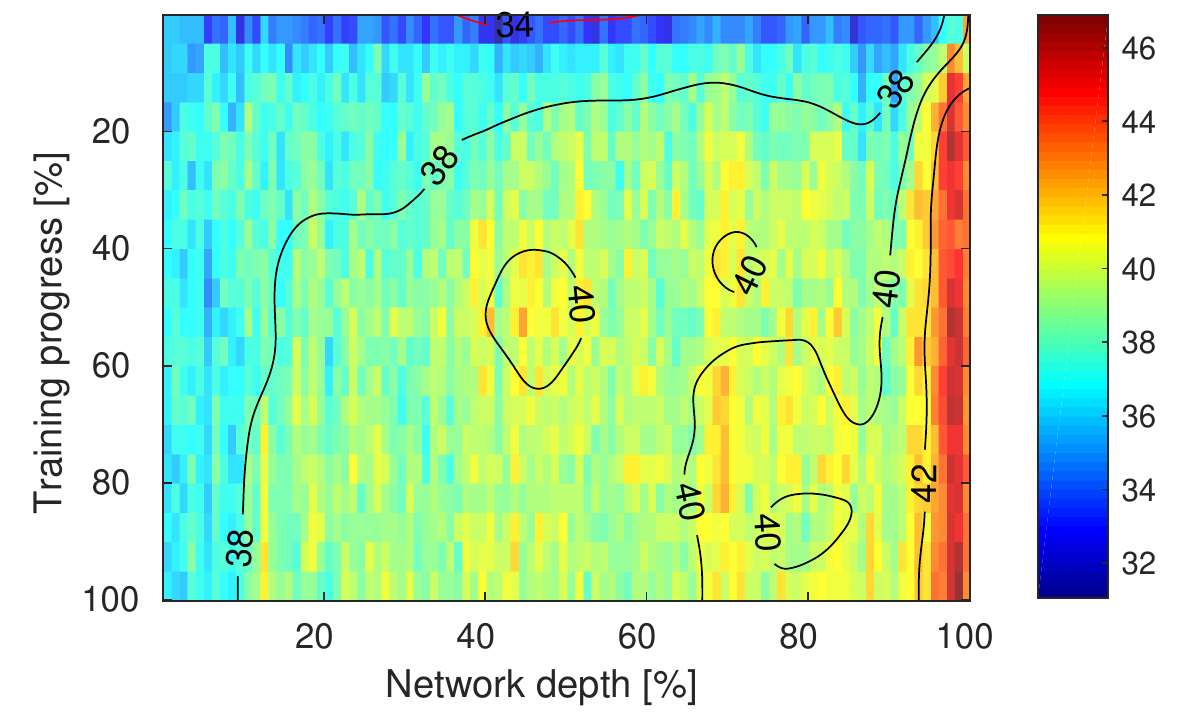}
		\includegraphics[width=0.325\textwidth, trim={2pt 2pt 8pt 2pt}, clip]{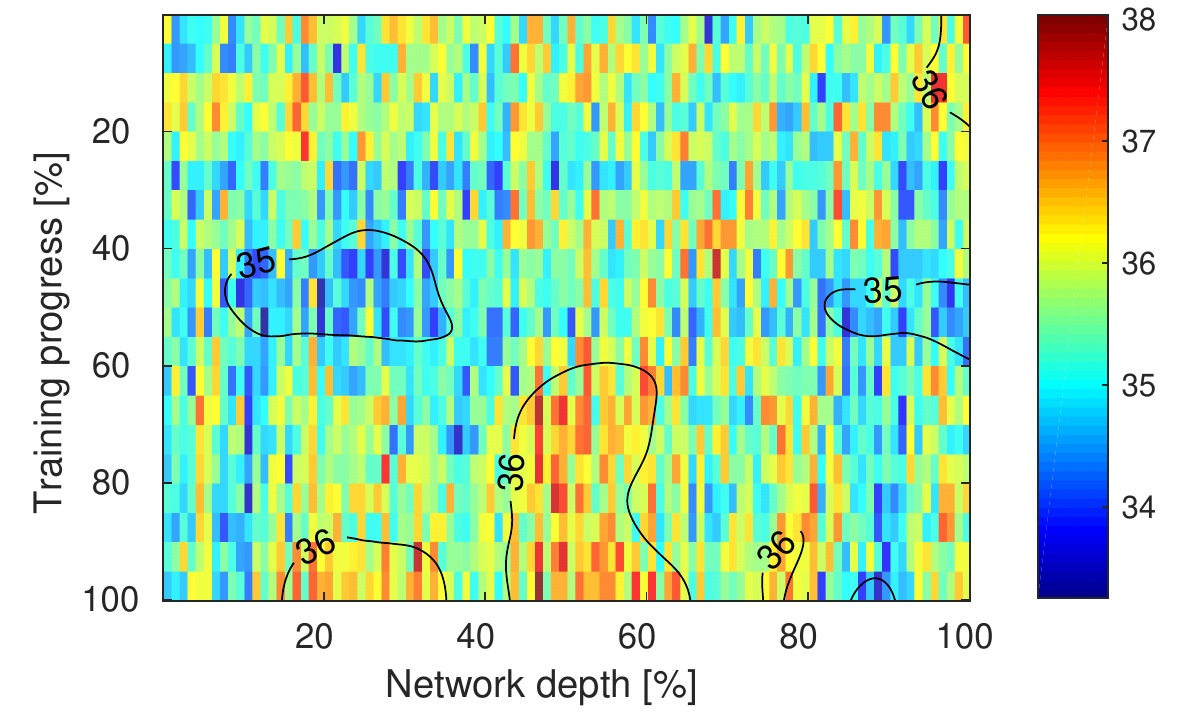}
		\includegraphics[width=0.325\textwidth, trim={2pt 2pt 8pt 2pt}, clip]{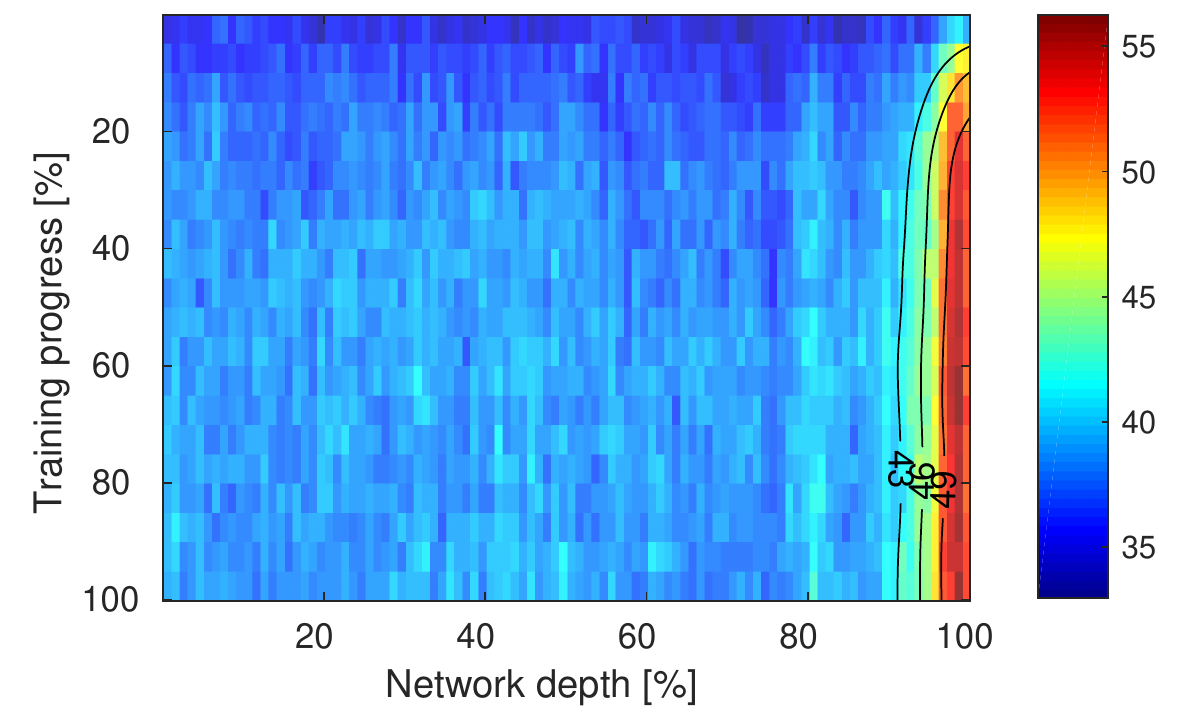}
		\caption{Depth FC}
		\label{fig:dmc_depth_fc}
	\end{subfigure}\\
	\vspace{0.2cm}
	\begin{subfigure}{0.87\textwidth}
		\includegraphics[width=0.325\textwidth, trim={2pt 2pt 8pt 2pt}, clip]{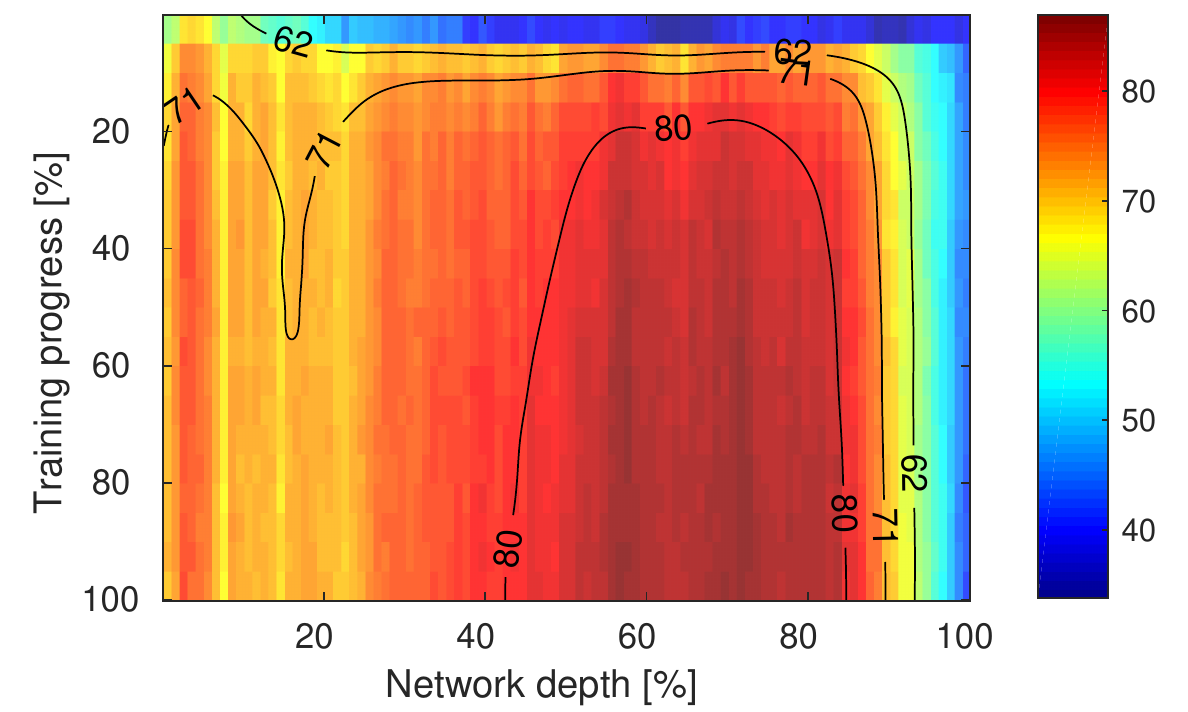}
		\includegraphics[width=0.325\textwidth, trim={2pt 2pt 8pt 2pt}, clip]{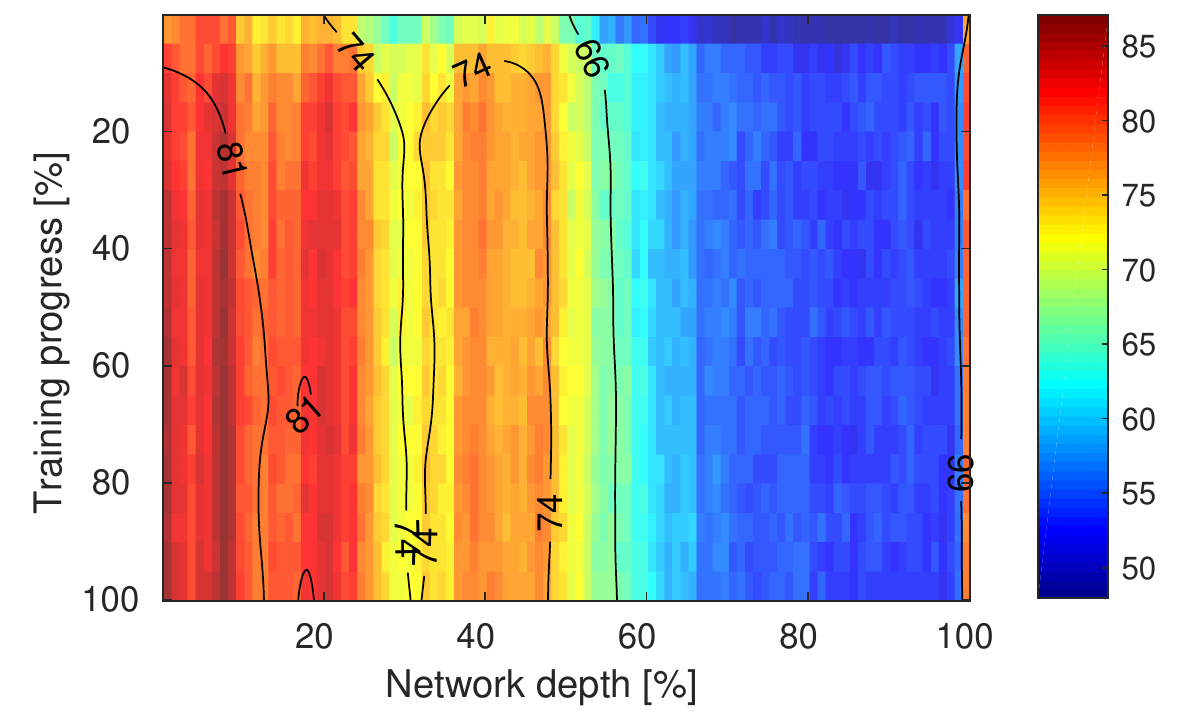}
		\includegraphics[width=0.325\textwidth, trim={2pt 2pt 8pt 2pt}, clip]{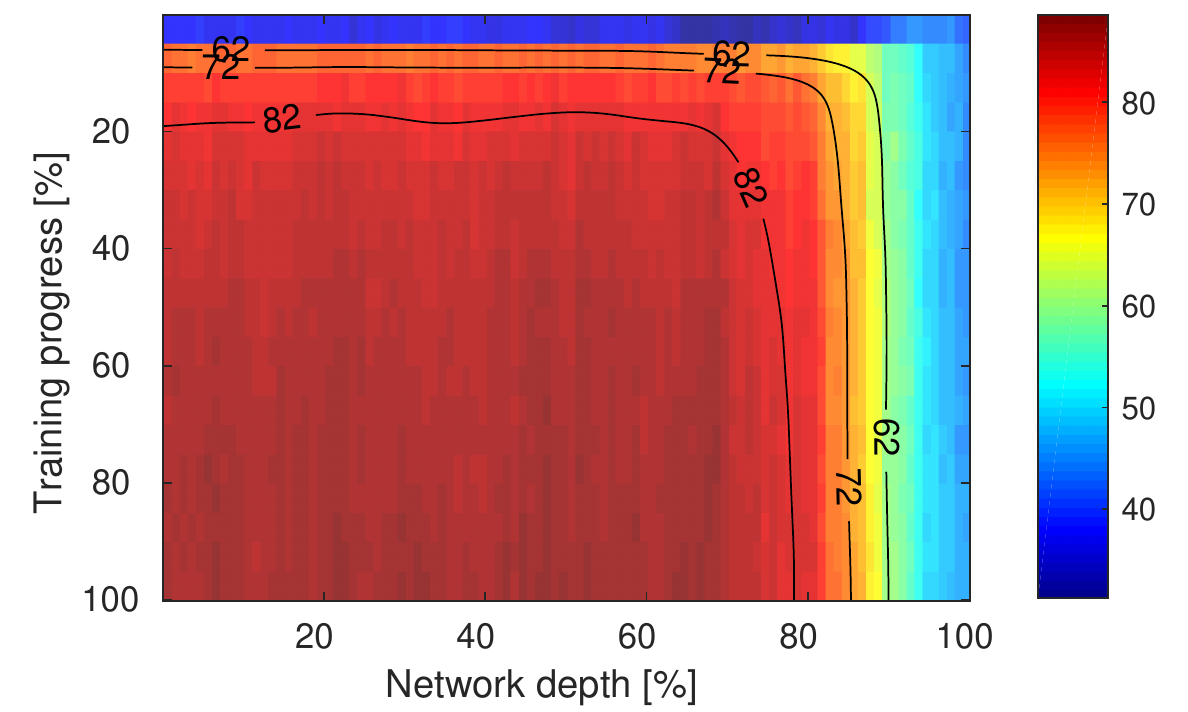}
		\caption{Width conv.}
		\label{fig:dmc_width_conv}
	\end{subfigure}\\
	\vspace{0.2cm}
	\begin{subfigure}{0.87\textwidth}
		\includegraphics[width=0.325\textwidth, trim={2pt 2pt 8pt 2pt}, clip]{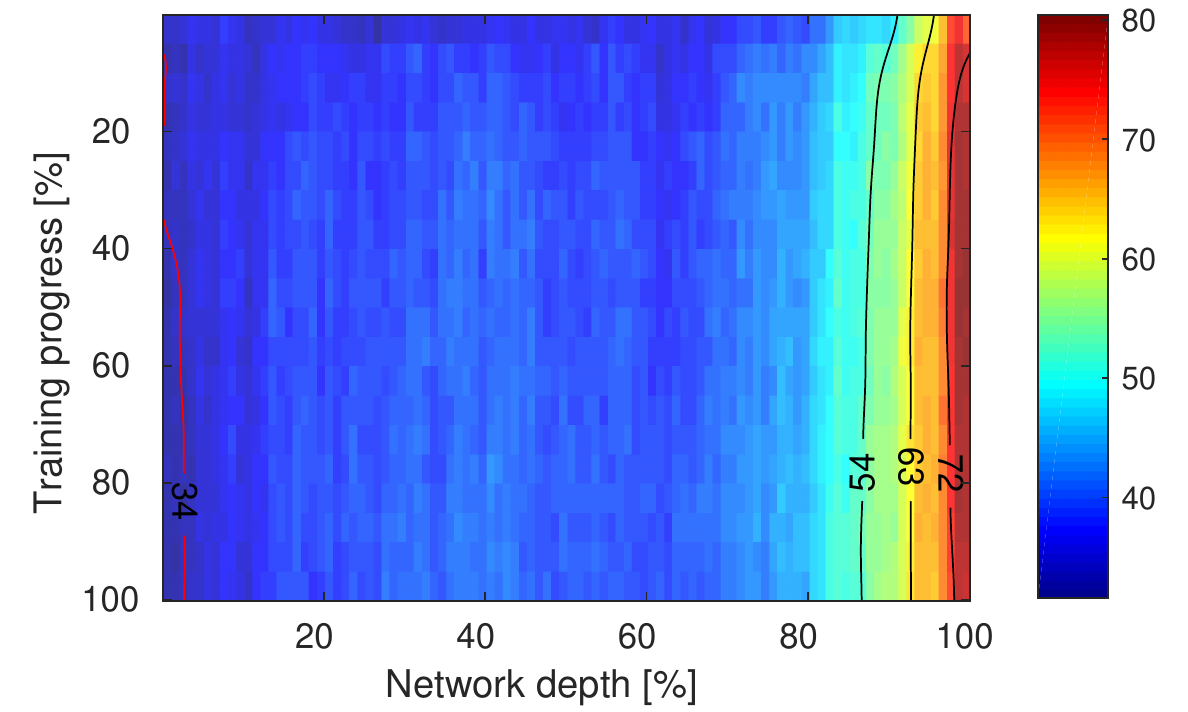}
		\includegraphics[width=0.325\textwidth, trim={2pt 2pt 8pt 2pt}, clip]{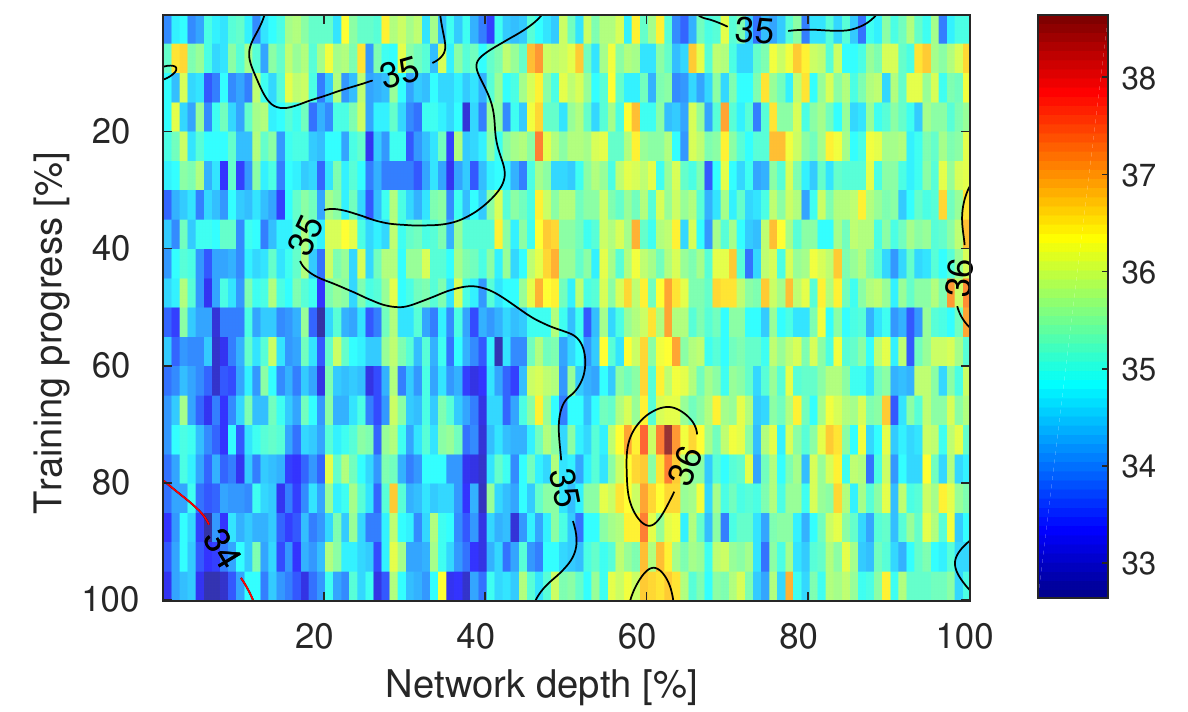}
		\includegraphics[width=0.325\textwidth, trim={2pt 2pt 8pt 2pt}, clip]{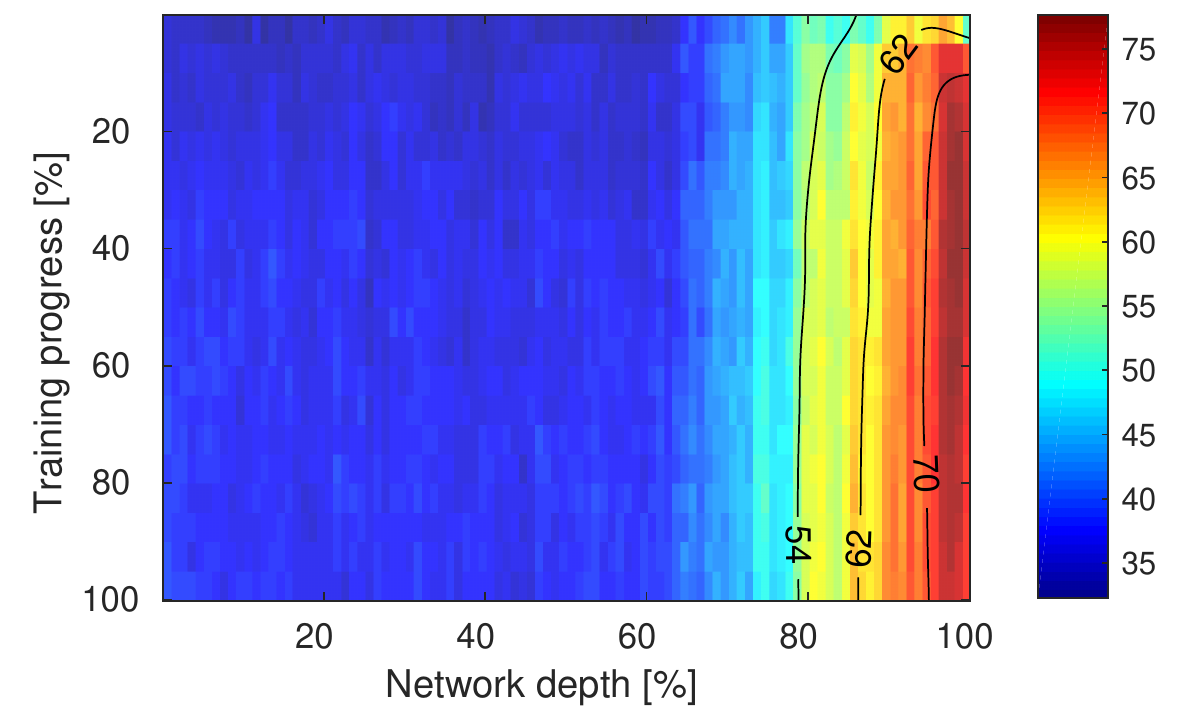}
		\caption{Width FC}
		\label{fig:dmc_width_fc}
	\end{subfigure}
	\caption{\label{fig:dmc_2} DMC performance maps, demonstrating differences across training progress and network depth. The results in the left figures are from DMCs trained on subsets of all weights in a model, the middle uses only convolutional weights, and the right uses only FC weights. This means that the x-axis covers all weights, convolutional weights and FC weights, in the left, middle and right figures, respectively. Note that the ranges of the colormaps differ between plots.}
\end{figure*}

\figref{dmc_1} and \ref{fig:dmc_2} show performance maps for all of the trained local DMCs. The columns correspond to DMCs trained on subsets picked from all positions of a weight vector $\theta$ (left), subsets only taken from the convolutional layers of $\theta$ (middle), and subsets only from the FC weights (right).

There are many things that can be seen in the results, and we only discuss around a few of these. For the DMCs trained to classify optimizer and activation function, the performance maps are very similar. Looking at the DMCs trained on weights from the whole network, we can only see that the performance increases in the beginning and end of the CNNs. However, looking at DMCs trained on only convolutional or FC weights, there are stripes of increasing performance. Inspecting the feature importances in \figref{coeff_global_lw}, we see how optimizer is best determined from bias weights and that activation function is easier to detect in the running mean weights. Thus, the striped patterns are most likely pinpointing the location of bias and running mean weights for optimizer and activation meta-classifiers, respectively. Although these types of weights occur at different locations due to the varying architectures of the dataset, on average they will end up in the indicated locations.

For the initialization DMC, there is a faster decrease in performance in the earlier layers. Looking at the performance of DMCs trained on convolutional and FC weights, it is clear how this decrease is contained in the convolutional layers, i.e. that these weight faster move away from the initialization. Also, in the last one or two FC layers the decrease is more rapid.

For the DMCs trained to detect filter size, the DMC trained on only convolutional layer weights is superior. This is expected, since the DMC only has to learn about the frequency information induced by vectorization of the convolutional filters. However, what is more interesting is how a large fraction of the FC layers also reflect an increase in accuracy, pointing to how the filter size affects the network on a more global level.

\end{document}